\documentclass{article}

\usepackage[main, final]{neurips_2025}
\usepackage{graphicx} 
\usepackage[utf8]{inputenc} %
\usepackage[T1]{fontenc}    %
\usepackage{hyperref}       %
\usepackage{url}            %
\usepackage{booktabs}       %
\usepackage{amsfonts}       %
\usepackage{nicefrac}       %
\usepackage{microtype}      %
\usepackage{amsmath}
\usepackage{changepage}
\usepackage{overpic}
\usepackage{wrapfig}
\usepackage{caption}
\usepackage{subcaption}
\usepackage{bm}

\usepackage{amssymb}
\usepackage{mathtools}
\usepackage{amsthm}
\usepackage{subcaption}

\usepackage{multirow}

\usepackage{amsthm}

\usepackage{overpic}
\theoremstyle{plain}
\newtheorem{theorem}{Theorem}[section]
\newtheorem{proposition}[theorem]{Proposition}

\theoremstyle{definition}

\theoremstyle{remark}
\usepackage{algorithm}
\usepackage{algpseudocode}

\usepackage{booktabs}
\usepackage[table]{xcolor}
\usepackage{wrapfig} %

\newcommand{\norm}[1]{\left\lVert#1\right\rVert}  %

\title{Connecting Neural Models Latent Geometries with \\ Relative Geodesic Representations}

\author{%
  Hanlin Yu$^1$ %
  \\
  University of Helsinki \\
  \And
  Berfin Inal \\
  University of Amsterdam \\
  \And
  Georgios Arvanitidis \\
  DTU \\
  \And
  Søren Hauberg \\
  DTU \\
  \And
  Francesco Locatello \\
  IST Austria \\
  \And
  Marco Fumero$^1$\\
  IST Austria \\
}

\usepackage{amsmath}
\usepackage{comment}
\usepackage{enumitem}

\DeclareMathOperator*{\argmax}{arg\,max}

\begin{document}

\maketitle

\vspace{-0.25cm}

\begin{abstract}Neural models learn representations of high-dimensional data on low-dimensional manifolds. Multiple factors, including stochasticities in the training process, model architectures, and additional inductive biases, may induce different representations, even when learning the same task on the same data. However, it has recently been shown that when a latent structure is shared between distinct latent spaces, relative distances between representations can be preserved, up to distortions. Building on this idea, we demonstrate that exploiting the differential-geometric structure of latent spaces of neural models, it is possible to capture \emph{precisely} the transformations between representational spaces trained on similar data distributions. Specifically, we assume that distinct neural models parametrize approximately the same underlying manifold, and introduce a representation based on the \emph{pullback metric} that captures the intrinsic structure of the latent space, while scaling efficiently to large models. We validate experimentally our method on model stitching and retrieval tasks, covering autoencoders and vision foundation discriminative models, across diverse architectures, datasets, pretraining schemes and modalities. Code is available at \url{https://github.com/marc0git/RelativeGeodesics}.
\end{abstract}

\footnotetext[1]{Corresponding emails: \texttt{marco.fumero@ist.ac.at}, \texttt{hanlin.yu@helsinki.fi}}

\vspace{-0.3cm}

\section{Introduction} 

\begin{wrapfigure}[17]{r}{0.49
\linewidth}
\vspace{-1.5cm}
    \centering    \includegraphics[width=\linewidth]{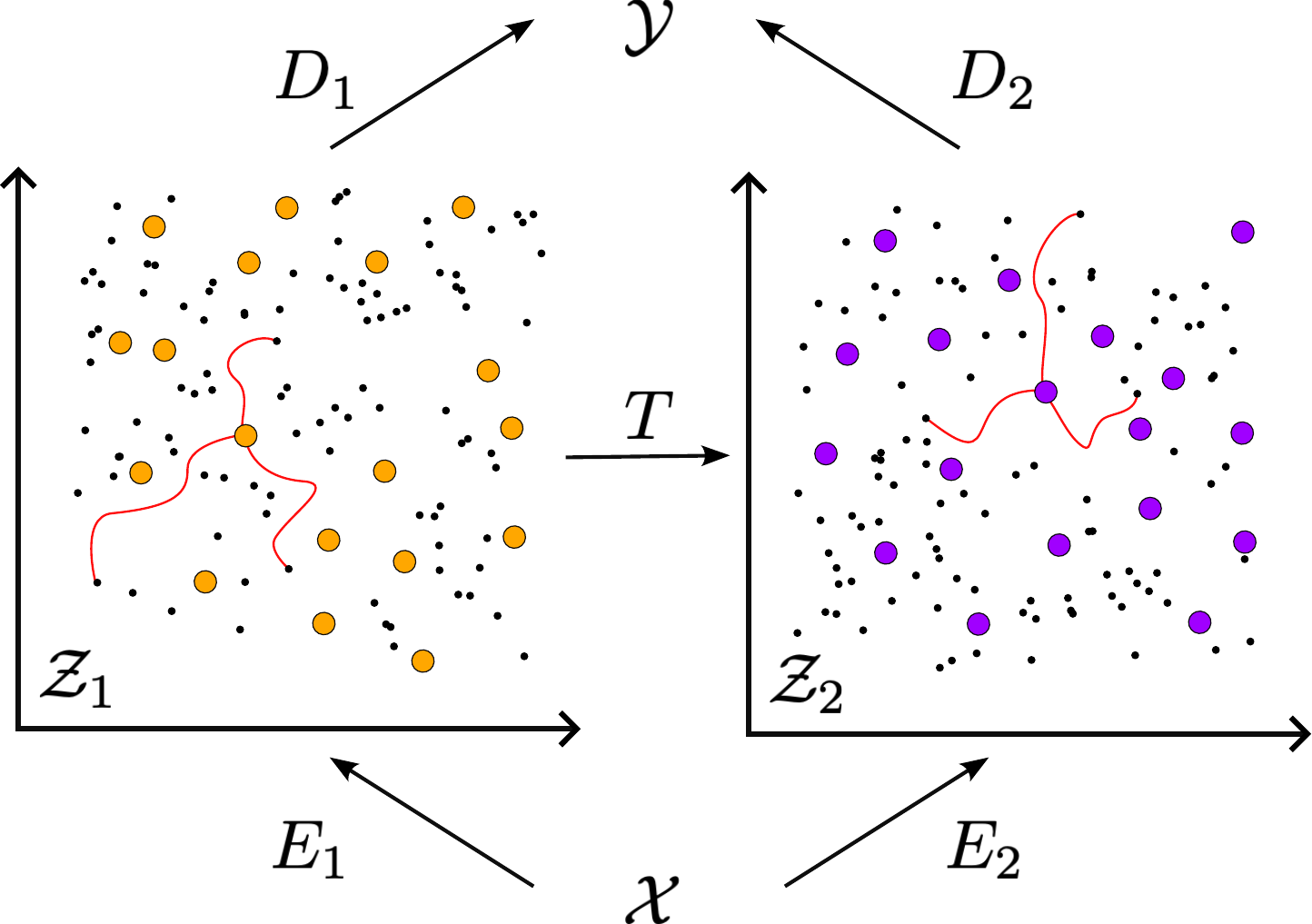}
    \caption{\emph{Neural models trained on similar data learn parametrizations of the same manifold.} NNs learn parametrizations ($D_1, D_2$) of the same underlying manifold $\mathcal{Y}$ up to isometries $T$. Pulling back the metric from $\mathcal{Y}$ makes relative geodesic representations invariant to transformations $T$ between latent spaces $\mathcal{Z}_1$ and $\mathcal{Z}_2$.}
    \label{fig:teaser}
\end{wrapfigure}

Neural models learn meaningful representations of high-dimensional data generalizing to many tasks, spanning different data modalities and domains. 
Recent research reveals that these models often develop similar internal representations given similar inputs \citep{li2015convergent,Moschella2023,fumero2024latentfunctionalmapsspectral,kornblith2019similarity}, a phenomenon that was observed in biological networks \citep{Laakso2000ContentAC,haxby2001distributed}.
Remarkably, even when models have different architectures, their internal representations can frequently be aligned through a simple, e.g. orthogonal, transformation \citep{maiorca2024latent,lahner2023direct,moayeri2023text}. This suggests a certain consistency in how neural nets encode information, emphasizing the importance of studying the internal representations and the transformations that relate them, to the extent to hypothesize whether neural nets are converging toward a unique representation of reality \citep{huh2024platonicrepresentationhypothesis}.

One strategy to understand how different models are related is to identify representations that are \emph{invariant} to transformations between distinct models' representational spaces.
A simple and effective recipe is that of \emph{relative representations} \citep{Moschella2023}, where samples are represented as a function of a fixed set of latent representations. The similarity function employed is cosine similarity, hinting at the fact that representations across distinct models are subject to \emph{angle preserving} transformations.
However,  the choice of similarity function should not be limited to only capturing invariances of one class of transformations. As shown in \citet{cannistraci2023bricks,fumero2021learning}, other choices can be good as well, and there is not a clear best choice among different transformations for capturing transformation across distinct latent spaces. %
We posit that when it is possible to relate distinct neural models' representational spaces, 
neural models are learning distinct parametrizations of the \emph{same} underlying manifold (see Figure \ref{fig:teaser}). 
In this paper, we employ geodesic distance in the latent space
for relative representations. This approach ensures that the relative space remains approximately invariant to the isometries and reparametrization of the data’s manifold, as characterized by a Riemannian structure. Our contributions can be summarized as follows: 
\begin{itemize}[leftmargin=*]
    \item We observe that distinct neural models learn parametrization of the same underlying manifold when trained on similar data. 
    \item We propose a new representation that captures the isometric transformation between data manifolds learned by distinct models, by leveraging the pullback metric.
    \item We propose to employ a scalable approximation of the geodesic energy to compute intrinsic distances that preserve the ranks of true distances.
    \item We show how to get meaningful pullback metrics from discriminative models, such as classifiers and self-supervised models.
    \item \looseness=-1 We test relative geodesics on retrieval and stitching tasks on autoencoders and vision foundation models, across different models, training schemes, and modalities, outperforming prior methods.
\end{itemize}

\section{Related Work}

\textbf{Representation alignment.~~}
Numerous studies have shown that neural networks trained under different initializations, architectures, or objectives learn highly similar internal representations \citep{Moschella2023,bonheme2022variationalautoencoderslearninsights, kornblith2019similarity, cannistraci2023bricks,li2015convergent,bengio2014representationlearningreviewnew,maiorca2024latent, huh2024platonicrepresentationhypothesis, guth2024rainbowdeepnetworkblack, chang2022geometrymultilinguallanguagemodel, conneau2018wordtranslationparalleldata, tsitsulin2020shapedataintrinsicdistance,nejatbakhsh2024comparing}. This correspondence becomes stronger in wide and large networks \citep{pmlr-v162-barannikov22a, morcos2018insightsrepresentationalsimilarityneural, somepalli2022neuralnetslearnmodel}. Leveraging these aligned embeddings, a simple linear transformation often suffices to map one network's latent space onto another's, enabling techniques such as model stitching, where components from different networks can be interchanged with minimal loss in performance \citep{fumero2024latentfunctionalmapsspectral, NEURIPS2021_01ded425, csiszárik2021similaritymatchingneuralnetwork}. 
In practice, aligning latent spaces using a linear transformation achieves comparable downstream task performance \citep{moayeri2023text, merullo2023linearlymappingimagetext, maiorca2024latent, lahner2023direct}.

\textbf{Latent space geometry.~~}
Early work on the geometry of deep latent representations focused on autoencoders, where the decoder's mapping from latent to data space induces a natural \emph{pullback metric} under the assumption that the ambient space is Euclidean \citep{Shao2018, Tosi2014,Arvanitidis2018}. The Riemannian viewpoint allows one to compute geodesic paths and meaningful distances that respect the manifold structure of the learned embedding. Subsequent research has introduced computationally efficient approximations, such as energy-based proxies, and extended these ideas to estimate local curvature for improved interpolation and sampling \citep{chen2019fast, chadebec2022geometricperspectivevariationalautoencoders, loaizaganem2024deepgenerativemodelslens, arvanitidis:aistats:2021,arvanitidis:aistats:2022a}. In the context of discriminative models, one can obtain a Riemannian metric primarily using two approaches \citep{Grosse2022}, either by pulling back the Fisher Information Matrix \citep{Amari2016,Arvanitidis2022} or by assuming a Euclidean geometry on the output space and pulling back the $L2$ metric. Interestingly, one can obtain some identifiability guarantees by taking geometry into consideration \citep{syrota2025identifying}.\looseness=-1

\section{Method}
\label{sec:method}
\subsection{Notation and background}
\label{sec:notation_and_background}

Neural networks (NNs) are parametric functions $F_\theta$, composed of an \emph{encoding} map and a \emph{decoding} map, represented as $F_\theta = D_{\theta_2} \circ E_{\theta_1}$. The encoder $E_{\theta_1}: \mathcal{X} \mapsto \mathcal{Z}$ generates a latent representation $\bm{z} = E_{\theta_1}(\bm{x})$, where $\bm{x} \in \mathcal{X}$ is mapped from the input domain $\mathcal{X}$ to the latent space $\mathcal{Z}$. %
The decoder $D_{\theta_2}$ is responsible for performing the task at hand, such as reconstruction or classification. For simplicity, we omit the parameter dependence ($\theta$) in our notation moving forward. For any single module $E$ (or equivalently $D$), we use $E_\mathcal{X}$ to denote that the module $E$ was trained on the domain $\mathcal{X}$. In the next sections, we will provide the necessary background to introduce our method.

\textbf{Latent space communication.~}
Given a pair of domains $(\mathcal{X},\mathcal{X}')$, a pair of neural models trained on them $(F_\mathcal{X}^1,F_\mathcal{X'}^2)$ and a partial correspondence between the domains $\Gamma: \mathcal{A}_\mathcal{X} \mapsto \mathcal{A}_\mathcal{X'}$ where $\mathcal{A}_\mathcal{X} \subset \mathcal{X}$ and $\mathcal{A}_\mathcal{X'} \subset \mathcal{X}'$, the problem of \emph{latent space communication} is the one of finding a full correspondence $\Lambda: E^1(\mathcal{X}) \mapsto E^2(\mathcal{X}')$ between the two domains, from $\Gamma$. In a simplified setting, for example two models trained with different initialization or architectures on the same data, $\mathcal{X}=\mathcal{X}'$ and the correspondence is the identity. When $\mathcal{X} \neq \mathcal{X}'$ the problem recovers the multimodal setting.

\textbf{Relative representations.~}
 The relative representations framework \citep{Moschella2023} provides a straightforward approach to represent each sample in the latent space according to its similarity to a set of fixed training samples, denoted as \textit{anchors}. Representing samples in the latent space as a function of the anchors corresponds to transitioning from an absolute coordinate frame into a \emph{relative} one defined by the anchors and the similarity function. 
Given a domain $\mathcal{X}$, an encoding function $E_\mathcal{X}: \mathcal{X} \to \mathcal{Z}$, a set of anchors $\mathcal{A}_\mathcal{X}\subset\mathcal{X}$, and a similarity or distance function $d: \mathcal{Z}\times \mathcal{Z} \to \mathbb{R}$, the \emph{relative representation} for a sample $\bm{x} \in \mathcal{X}$ is:
\begin{equation*}
    RR(\bm{z}; \mathcal{A}_\mathcal{X}, d) = \bigoplus_{\bm{a}_i\in \mathcal{A}_\mathcal{X}} d(\bm{z},E_\mathcal{X}(\bm{a}_i)),
\end{equation*}
where $\bm{z} = E_\mathcal{X}(\bm{x})$, and $\bigoplus$ denotes row-wise concatenation. 
In the original method \citep{Moschella2023}, $d$ corresponds to cosine similarity. This choice induces a representation invariant to \emph{angle-preserving transformations}. In this work, our focus is to \emph{leverage the intrinsic geometry of latent spaces to employ a metric that captures isometric transformations between data manifolds.}

\textbf{Latent space geometry.~}
For the latent space of a neural network, it is generally hard to reason about its Riemannian structure. However, it is often easier to assign a Riemannian structure to the output space. As such, one can define a \emph{pullback metric} from the output space to the latent space, which is a standard operation in Riemannian geometry (see Ch.2.4 of \citet{do1992riemannian}).

Formally, the decoder $D:\mathcal{Z} \mapsto \mathcal{Y}$ takes as input a latent representation $\bm{z} \in \mathcal{Z}$ and outputs $y$. Given a Riemannian metric defined on $y$ as $G_{\mathcal{Y}}(y)$, one can obtain the Riemannian metric at $\bm{z}$ as:
\begin{equation*}
G_{\mathcal{Z}}(\bm{z}) = \left(\frac{\partial \bm{y}}{\partial \bm{z}}\right)^{\!\!\top} G_{\mathcal{Y}}(\bm{y}) \left(\frac{\partial \bm{y}}{\partial \bm{z}}\right) = J_D(\bm{z})^\top G_{\mathcal{Y}}(\bm{y}) J_D(\bm{z}),
\end{equation*}
where $J_D(\bm{z})$ is the Jacobian of $D$ at $\bm{z}$.
The metric tensor $G_{\mathcal{Y}}$ is useful to compute quantities such as lengths, angles, and areas on $\mathcal{M}$.  Given a smooth curve $\bm{\gamma}:[a,b] \mapsto \mathcal{M}$, its arc length is defined as:
\begin{align} \label{eq:arc_length}
L(\bm{\gamma}) &= \int^{b}_{a} \sqrt{\bm{v}(t)^{\top} G_{\mathcal{Y}}(\bm{\gamma}(t)) \bm{v}(t)^{\top}} \mathrm{d} t,
\end{align}
where $\bm{v}(t)=\dot{\bm{\gamma}}(t)$.
A slight variation of the above functional gives the geodesic energy $\mathcal{E}$ of $\bm{\gamma}$ \citep{Arvanitidis2018, Shao2018}
\begin{align} \label{eq:energy}
\mathcal{E}(\bm{\gamma}) &= \frac{1}{2} \int^{b}_{a} \bm{v}(t)^{\top} G_{\mathcal{Y}}(\bm{\gamma}(t)) \bm{v}(t)^{\top} \mathrm{d} t.%
\end{align}
Both can be discretized and approximated in practice using finite difference approaches \citep{yang:arxiv:2018, Shao2018}. Geodesics minimize both the length and the energy, where for optimization the latter is usually preferred for numerical stability \citep{dggm}. These quantities have the property of being \emph{invariant} to certain reparametrizations, as formalized in the following proposition:

\begin{proposition}
\label{thm:inv_rep}
Let \(\bm{\gamma}:[0,1]\to \mathcal{M}\) be a smooth curve on a Riemannian manifold \((\mathcal{M},G)\), and let $\bm{\gamma}'$ be the same curve expressed using another coordinate chart of the same manifold with Riemannian metric $G'$. The Riemannian length and energy of \(\bm{\gamma}\) are invariant under such change of coordinates:
\begin{align*}
\mathcal{E}[\bm{\gamma}]
&=\frac{1}{2}\int_{0}^{1}\bigl\lVert\tfrac{d\bm{\gamma}}{d t}\bigr\rVert^{2}_{G}\,d t
=\frac{1}{2}\int_{0}^{1}\bigl\lVert\tfrac{d\bm{\gamma}'}{d t}\bigr\rVert^{2}_{G'}\,d t,\\
L[\bm{\gamma}]
&=\int_{0}^{1}\bigl\lVert\tfrac{d\bm{\gamma}}{d t}\bigr\rVert_{G}\,d t
=\int_{0}^{1}\bigl\lVert\tfrac{d\bm{\gamma}'}{d t}\bigr\rVert_{G'}\,d t.
\end{align*}
Furthermore, the Riemannian arc length of \(\bm{\gamma}\) is invariant under reparameterizations of the curve. Denote the original parameter of $\bm{\gamma}$ as $t \in [0,1]$, where $\tau \in [0,1]$ is a new parameter such that $t=\varphi(\tau)$, where $\varphi$ satisfies $\varphi(0)=0$, $\varphi(1)=1$ and $\varphi'(\tau) > 0$. Define \(\tilde{\bm{\gamma}}(\tau) = \gamma(\varphi(\tau))\). Then:
\[
L[\tilde{\bm{\gamma}}]
=\int_{0}^{1}\bigl\lVert\tfrac{d\tilde{\bm{\gamma}}}{d\tau}\bigr\rVert_{G(\tilde{\bm{\gamma}}(\tau))}\,d\tau
=\int_{0}^{1}\lVert\tfrac{d\bm{\gamma}}{d t}\rVert_{G(\bm{\gamma}(t))}\,dt
= L[\bm{\gamma}].
\]
\end{proposition}
We provide the proof in Appendix~\ref{app:proof_inv_rep}.

\subsection{Relative geodesics representations} %
\begin{algorithm}
\caption{Relative Geodesic Representations}
\begin{algorithmic}[1]
\Require Sample $\bm{x} \in \mathcal{X}$, anchors $\mathcal{A}_\mathcal{X}$, encoder $E$, decoder $D$, distance $d_{\mathcal{X}}$ induced by metric $G_{\mathcal{Y}}$, steps $N$, step size $\Delta t$, mode $\in \{\texttt{energy}, \texttt{distance}\}$
\Ensure $RR^{geo}(x; \mathcal{A}_\mathcal{X})$

\State $\bm{z} \gets E(\bm{x})$, \quad $RR^{geo} \gets [\ ]$
\For{$a \in \mathcal{A}_\mathcal{X}$}
  \State $\bm{z}_a \gets E(a)$, \quad $d \gets 0$
  \For{$j = 1$ to $N$}
    \State $\bm{\gamma}_j \gets (1 - \tfrac{j}{N}) \bm{z} + \tfrac{j}{N} \bm{z}_a$
    \State $\bm{\gamma}_{j-1} \gets (1 - \tfrac{j-1}{N}) \bm{z} + \tfrac{j-1}{N} \bm{z}_a$
    \State $\bm{v} \gets \frac{D(\bm{\gamma}_j) - D(\bm{\gamma}_{j-1})}{\Delta t}$
    \State $G \gets G_{\mathcal{Y}}(D(\bm{\gamma}_j))$
    \State $s \gets \bm{v}^\top G \bm{v}$
    \State $d \gets d + \Delta t \cdot (\texttt{energy} \Rightarrow \tfrac{1}{2}s,\ \texttt{distance} \Rightarrow \sqrt{s})$
  \EndFor
  \State Append $d$ to $RR^{geo}$
\EndFor
\State \Return $RR^{geo}$
\end{algorithmic}
\label{algorithm:relative_geodesic}
\end{algorithm}

From a differential geometry perspective, the problem of latent space communication can be interpreted as finding a transformation between the data manifolds $\mathcal{M}_1,\mathcal{M}_2$ approximated by two neural models $F_\mathcal{X}^1,F_\mathcal{X'}^2$. The relative representation framework captures this transformation implicitly if equipped with the right metric: we posit that a natural candidate for this metric is the geodesic distance defined on $\mathcal{M}_1,\mathcal{M}_2$, respectively. This choice makes the relative representations \emph{invariant} to isometric transformation $T$ of the manifolds $\mathcal{M}_1,\mathcal{M}_2$. 
However, for high-dimensional problems, the high cost of computing the geodesic (corresponding to minimizing Eq.~\ref{eq:energy}) makes this impractical \citep{Shao2018,chen2019fast}. Furthermore, one can argue against directly using the latent geometry induced by deterministic models from a theoretical perspective \citep{Hauberg2019}, as it may result in undesirable properties, for example the geodesics going outside of the data manifold.

\begin{figure}[t]
  \centering
  \begin{subfigure}[t]{0.48\textwidth}
    \centering
    \begin{overpic}[width=\linewidth]{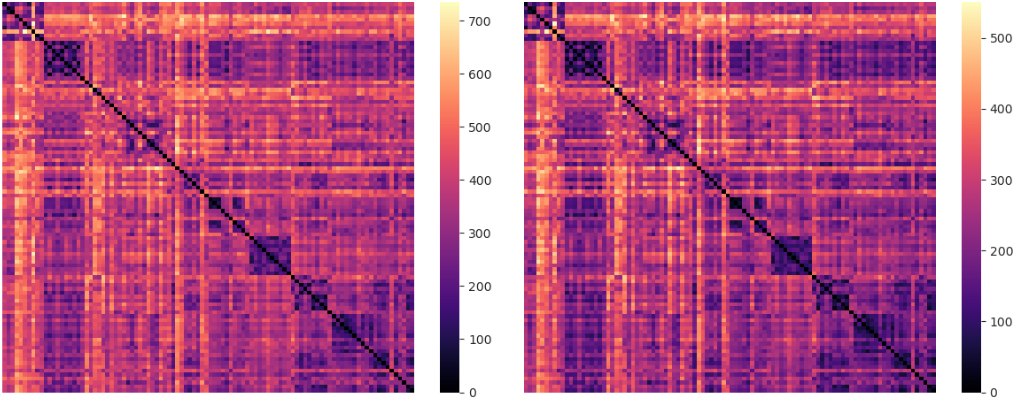}
    \put(1,42){\footnotesize{Approximate energies}}
    \put(54,42){\footnotesize{Geodesic energies}}
    \end{overpic}
    \caption{MNIST }
    \label{fig:mnist-10}
  \end{subfigure}
  \hfill
  \begin{subfigure}[t]{0.48\textwidth}
    \centering
    \begin{overpic}[width=\linewidth]{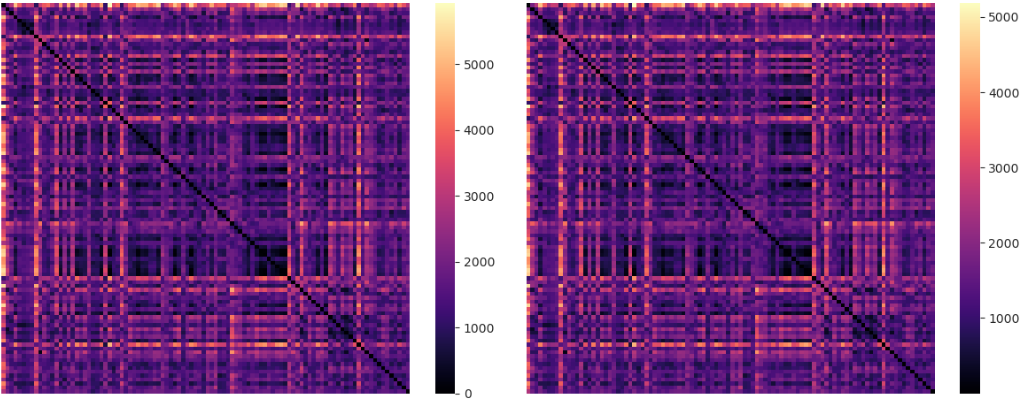}
    \put(1,41){\footnotesize{Approximate energies}}
    \put(54,41){\footnotesize{Geodesic energies}}
    \end{overpic}
    \caption{CIFAR-10 }
    \label{fig:cifar10-10}
  \end{subfigure}
  \caption{\looseness=-1 Pairwise latent‐space energy matrices for (a) MNIST and (b) CIFAR-10. In each subfigure, the left heatmap shows the straight-line energy approximation and the right shows the  geodesic energies of the ground truth geodesic curve. The Spearman rank correlations between the two measures are \(\rho=0.99\) for MNIST and \(\rho=1.00\) for CIFAR-10, demonstrating near-perfect agreements. }
  \label{fig:comparison-10}
\end{figure}

We therefore approximate the geodesic quantities by directly considering the energy (or the length) of the straight line (in the Euclidean sense) connecting representations in the latent space: %
\begin{equation*}
    RR^{geo}(\bm{z}; \mathcal{A}_\mathcal{X}) = \bigoplus_{\bm{a}_i\in \mathcal{A}_\mathcal{X}} \mathcal{E} (\tilde{\bm{\gamma}}_\alpha(\bm{z},E_\mathcal{X}(\bm{a}_i))),
\end{equation*}
where $\tilde{\bm{\gamma}}_\alpha(\bm{z}_1, \bm{z}_2)= (1- \alpha) \bm{z}_1 + \alpha \bm{z}_2 $ is the convex combination between the points $\bm{z}_1,\bm{z}_2$. 
The approximation gives a natural upper bound to the geodesic distance: for $\bar{\bm{\gamma}}$ it can be shown to relate to the arc length of a curve defined in Eq.~\ref{eq:arc_length} and the energy in Eq.~\ref{eq:energy} using the following bounds:

\begin{equation}
    d(\bm{z}_0, \bm{z}_1)^2 \leq L^2(\tilde{\bm{\gamma}}) \leq 2\mathcal{E}(\tilde{\bm{\gamma}}).
\label{eq:two_bounds}
\end{equation}
The proof is in Appendix~\ref{app:proof_bounds}. Moreover the approximation is far more \emph{efficient} to compute, without requiring minimization of Equation \ref{eq:energy}, and is \emph{accurate}, as empirically verified in Figure \ref{fig:comparison-10}.

\textbf{Discretization.~} When the step size is sufficiently small, the energy and arc length in the latent space, as defined in Equations \ref{eq:arc_length} and \ref{eq:energy}, can be approximated by their counterparts in the output space using discretized finite difference schemes \citep{Shao2018}:

\begin{align}
\mathcal{E}(\bm{\gamma}) &= \sum_{i=1}^{N}E_{i} = \frac{1}{2}\sum_{i=1}^{N}\bm{v}(t_{i})^{\top}G(t_{i})\bm{v}(t_{i}) \Delta t,  \\
L(\bm{\gamma}) &= \sum_{i=1}^{N}d_{i} = \sum_{i=1}^{N}\sqrt{\bm{v}(t_{i})^{\top}G(t_{i})\bm{v}(t_{i})} \Delta t,
\end{align}
where $\Delta t = \frac{1}{N}$, with $N$ being the number of discretization steps. For Euclidean geometry, the geodesic arc lengths are given in closed form as the geodesics are straight lines. Unlike the energy, the curve length is invariant under reparametrizations (proposition~\ref{thm:inv_rep}). As such we focus on the curve length in our experiments.
The resulting algorithm is summarized in Algorithm~\ref{algorithm:relative_geodesic}. In practice, with specific choice of $G_{\mathcal{Y}}$ one can avoid approximating the distance between $D(\bm{\gamma}_{j},\bm{\gamma}_{j-1})$ explicitly using $G_{\mathcal{Y}}$ by directly calculating the distance or energy between $\bm{\gamma}_{j}$ and $\bm{\gamma}_{j-1}$ on $\mathcal{Y}$.

\textbf{Approximate geodesic energies.~}
Our choice comes with three advantages: \emph{(i)} \emph{efficiency}: avoiding minimization of Eq.\ref{eq:energy} the computation for every sample reduces to a single forward pass for every discretization step $\bm{\gamma}$ and for each anchor, resulting in overall complexity of $O(T A)$ forward passes of the decoder,  %
where $A$ the number of anchors %
is the number of discretization steps. \emph{(ii)} Directly using the arc length ensures \emph{invariance} to reparametrizations of the manifold, matching our assumptions. \emph{(iii)} As we only need reasonably accurate estimates of the arc lengths rather than the geodesic trajectory, the approach is \emph{accurate}. Specifically, to assess how close the straight
line energy approximation \eqref{eq:energy} is to the true geodesic energies, we first encoded 100 samples (10 per class, sorted by label) from MNIST \citep{deng2012mnist} and CIFAR-10 using a simple convolutional autoencoder (architecture detailed in Appendix~\ref{app:arch}). We then computed pairwise geodesic energy matrices over these latent representations using both methods, and the results are displayed in Fig.~\ref{fig:comparison-10}. Visually, both energy matrices exhibit the same block-diagonal structure, mainly due to belonging to the same class, and clustering patterns. Numerically, their Spearman rank correlation exceeds 0.99 with only 8 discretization points (see Appendix~\ref{app:geodesic_approximation} for correlation results across different numbers of discretization steps and for implementation details).

\subsection{Choice of pullback metric}

The properties of the relative geodesic representations are determined by (i) the choice of the output space, (ii) the choice of the metric to pullback from the output space and (iii) the pretraining objective (e.g. reconstruction or classification) on which the decoder was trained.

\textbf{Generative models.~} For models trained on a reconstruction loss such as autoencoders, or on generative objectives, such as variational autoencoders \cite{kingma2013auto}, pulling back metrics such as $L2$ distance have been shown to effectively reflect the underlying geometry of the latent space \citep{Tosi2014,Arvanitidis2018,Hauberg2019}.

\textbf{Discriminative models.~}For discriminative models, such as classifiers or instance based discriminative models \citep{ibrahim2024occams}, it is not immediate how to assign a Riemannian structure to the space of latent representations. From the perspective of information geometry, perhaps the most natural choice is the Fisher information matrix \citep{Amari2016}, in which case the metric in the output space can be obtained as the one with categorical likelihood. However, neural networks typically experience Neural Collapse \citep{kothapalli2023neural}, possibly rendering the resulting geometry troublesome. We empirically inspect this approach in Appendix~\ref{app:relgeo_fisher}.
In this work we consider two principled approaches for discriminative models based on classification decoder heads and instance discrimination heads.

\textit{Pulling back from classifiers.~} 
Perhaps the most natural idea is, as discussed in Section~\ref{sec:notation_and_background}, to construct a pullback metric based on the model's outputs, by simply pulling back the Euclidean $L2$ metric from the logit space of the classifier. 
Given an arbitrary encoder model, we train a classification head upon the latent representations (extracted e.g. from the last layer) and pulling back the Euclidean $L2$ metric from the output logits of the head. The resulting relative geodesics representation will inherit properties of both the decoder head (up to class information) and the pretrained encoder.

\textit{Pulling back from instance discrimination decoders.~}
Diet \citep{ibrahim2024occams} is a self-supervised training method which has been shown to learn representations with strong generalization to downstream tasks, and yield identifiability guarantees \citep{reizinger2025cross-entropy}. Specifically, in the infinite data limit representations from the Diet objective align the cluster centers of von-Mises Fisher (vMF) distributions, which lie on a unit sphere. The loss based on simple instance discrimination is:

\begin{equation}
\mathcal{L}_{Diet} = \mathbb{E}_{\bm{x},i}\left[-\log\frac{\exp\left(\bm{w}_{i}^\top f(\bm{x})\right)}{\sum_{j}\exp\left( \bm{w}_{j}^\top f(\bm{x})\right)}\right],
\label{eq:diet_loss}
\end{equation}
where $\bm{W}$ is a linear projection and $f$ is a nonlinear map. Further, assigning the same instance label to data augmentations was shown beneficial to improve invariance. %
Originally proposed to train the entire neural network \citep{ibrahim2024occams}, 
we instead use it to learn a decoder $D$ on top of the pretrained neural network latent representations, by setting $D=f \circ \bm{W}$. To construct relative geodesic representations, we propose to pullback the spherical metric from the penultimate layer of the diet decoder, before $\bm{W}$. 
Further discussions on Diet can be found in Appendix~\ref{app:diet}. %

Both classifier and instance discriminator approaches discussed above use proper pullback metrics and fall under our proposed framework of relative geodesic representations: these representations inherit semantic information, up to class or instance level, from the decoder, while retaining structure of the pretrained encoder. Notably, when the encoder is pretrained, the relative geodesics representations are computationally efficient, since the decoder remains lightweight and the approximate geodesic energy computation is cheap. As we demonstrate in the following experimental sections, this approach yields meaningful and identifiable representations that are consistent across models.

\section{Experiments}
\label{sec:experiments}

In the following we will evaluate the performance of relative geodesic representations on latent communication problems across models with different initializations, architectures, sizes and modalities.

\textbf{Tasks description.~}
We evaluate our approach on two representative instantiations of the latent communication problem: \emph{retrieval} and \emph{neural stitching}.
In a retrieval setting we aim to solve the latent communication problem up to the \emph{instance} level.
Given pairs of model encoders $(E_\mathcal{X}^1, E_{\mathcal{X}'}^2)$ and access to their latent representations, we seek to recover a full correspondence $\Lambda$ starting from a partial one $\Gamma$.
For neural stitching the goal is to solve the latent communication problem up to the \emph{task-label} level.
Classical stitching approaches train an adapter $\Psi$ between intermediate components of distinct neural networks so that
$D_{\mathcal{X}'}^2 \circ \Psi \circ E_\mathcal{X}^1$
remains functional on a downstream task (e.g., classification).
In Section~\ref{sec:experiments_on_vision_foundation_models}, we operate in the \emph{zero-shot stitching} regime~\citep{Moschella2023}, where no adapter is trained explicitly.
Instead, we solve implicitly for the transformations between representations by mapping them into \emph{relative representation spaces}.
This enables stitching pairs of models without any fine-tuning or additional supervision.

In Section~\ref{sec:autoencoders}, we evaluate \emph{relative geodesic representations} on generative models, focusing on autoencoders. This analysis examines performance across networks trained with different initializations and datasets.
In Section~\ref{sec:experiments_on_vision_foundation_models}, we extend the evaluation to \emph{discriminative foundation models}, assessing performance at scale across diverse architectures, pretraining objectives (e.g., self-supervised and classification), datasets, and modalities.

\subsection{Experimental evaluation on autoencoders}
\label{sec:autoencoders}
In the following sections, we evaluate relative geodesic representations on the latent communication problem across autoencoder models trained with different initializations, architectures and datasets.

\subsubsection{Aligning independently trained neural representational spaces}
\begin{figure}[h]
 \begin{subfigure}[b]{0.32\linewidth}    
\includegraphics[width=\linewidth]{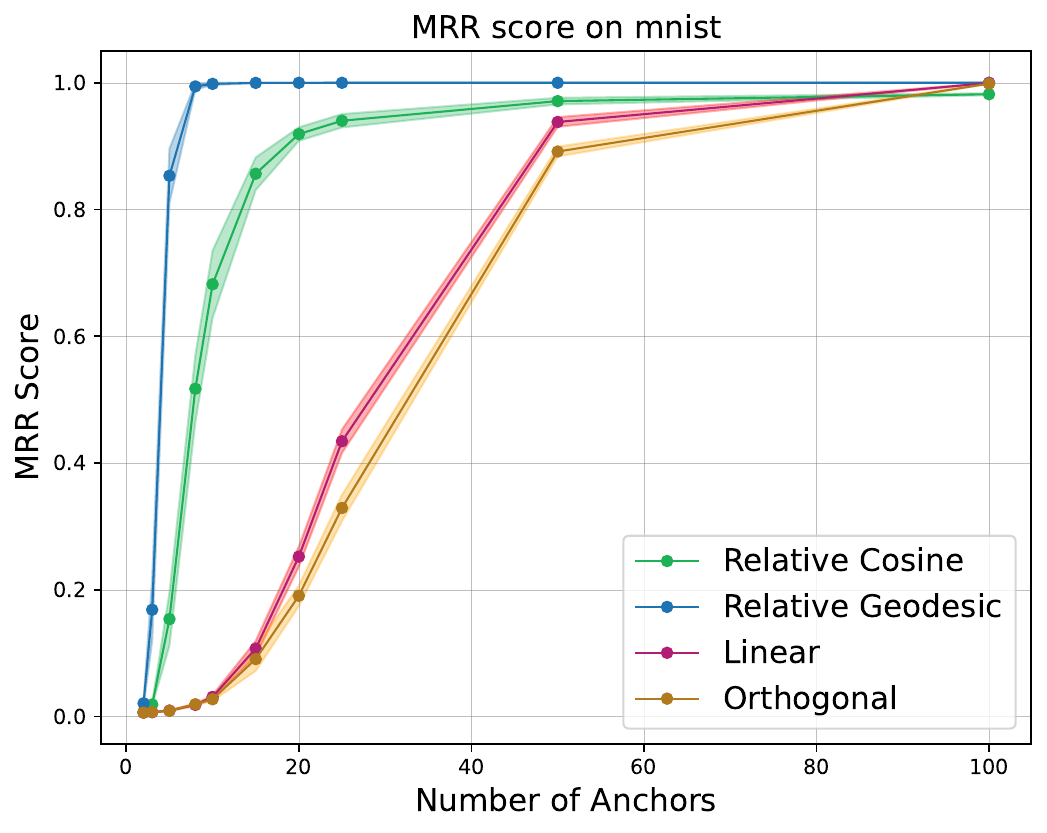}
 \end{subfigure}
  \begin{subfigure}[b]{0.32\linewidth}    
\includegraphics[width=\linewidth]{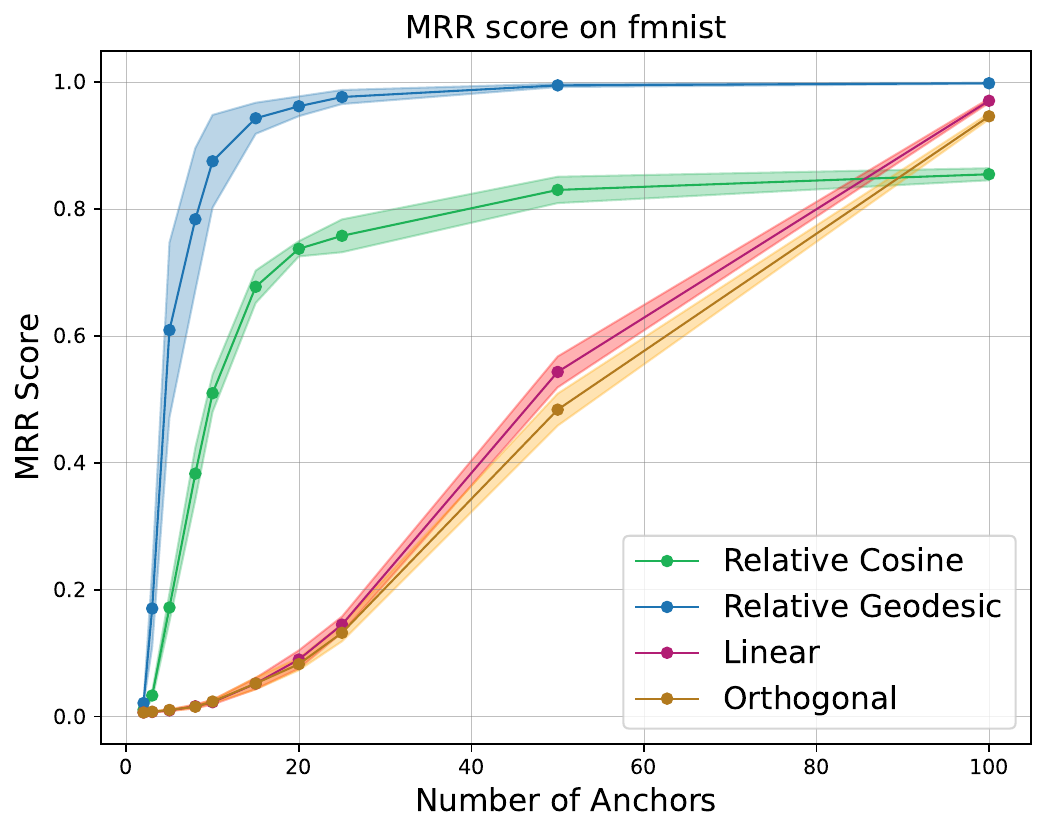}
 \end{subfigure}
  \begin{subfigure}[b]{0.32\linewidth}    
\includegraphics[width=\linewidth]{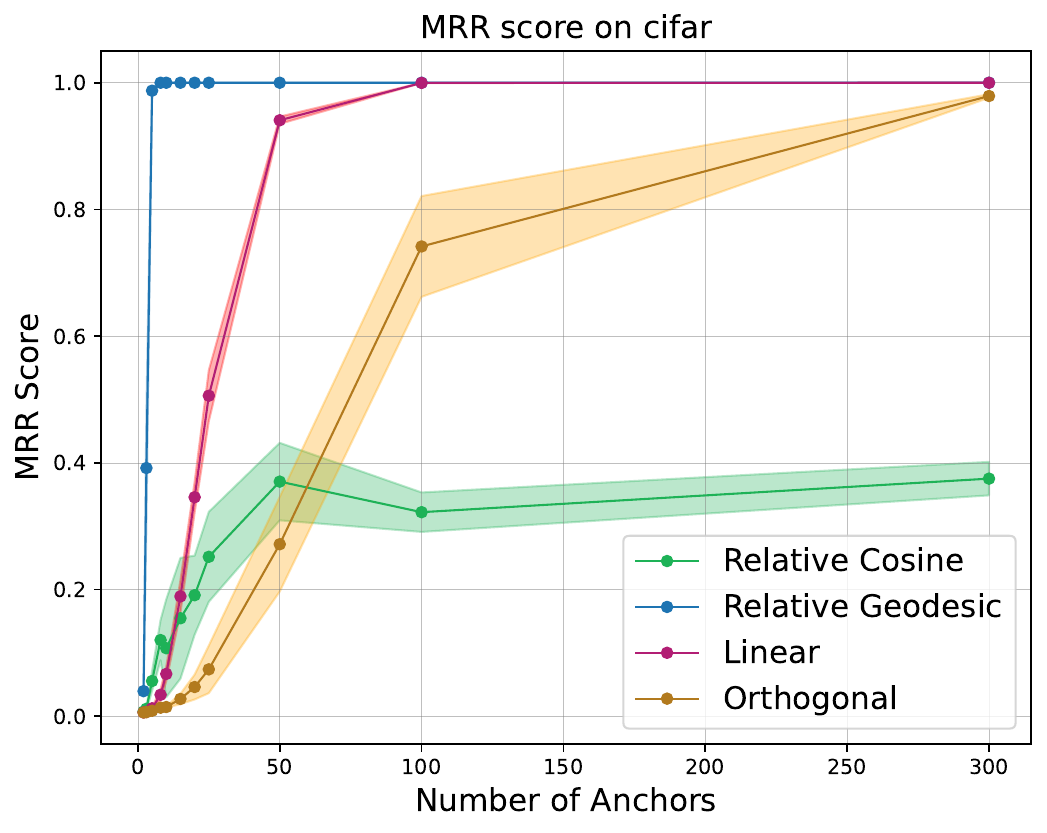}
 \end{subfigure}
 \caption{\emph{Aligning latent spaces of autoencoders}: MRR score as a function of the number of anchors on pairs of autoencoders trained with different initializations on the \texttt{MNIST} (left), \texttt{FashionMNIST} (center), \texttt{CIFAR10} (right) datasets, respectively. In green, we plot the performance of \citet{Moschella2023}; in red and orange the linear and orthogonal baselines respectively; in blue, our method. The shaded area indicates standard deviation across 5 different random sets of anchors. Relative geodesic consistently outperforms baselines, obtaining peak performance.}
 \label{fig:mrr_autoencoders}
\end{figure}

\textbf{Setting.~} \label{sec:exp_mrr_autoencoders}
For the following experiment, we trained pairs of convolutional autoencoders $(F_1,F_2)$ with different initializations on \texttt{MNIST} \citep{deng2012mnist}, \texttt{FashionMNIST} \citep{xiao2017online}, \texttt{CIFAR10} \citep{krizhevsky2009learning} datasets. The architecture of the convolutional autoencoder is detailed in Appendix~\ref{app:arch}. After training, we extracted 10k samples from the test set, and mapped them to the latent spaces of the two models, to representations $\mathbf{Z}_1=E_1(\mathbf{X}),\mathbf{Z}_2=E_2(\mathbf{X})$ respectively. Starting from a small set of anchors in correspondence $\Gamma: \mathcal{A}_\mathcal{X} \mapsto \mathcal{A}_\mathcal{Y}$, the objective is to evaluate how well it is possible to recover the full correspondence $\Lambda$ between the representations $\mathbf{Z}_1,\mathbf{Z}_2$ from the relative representations.
As a baseline, we compare with relative representations using cosine similarity \citep{Moschella2023}, and with fitting a linear or orthogonal mapping using $\Gamma$.

\textbf{Analysis of results.~}
Fig.~\ref{fig:mrr_autoencoders} plots the performance in terms of MRR on \texttt{MNIST}, \texttt{FashionMNIST} and \texttt{CIFAR10} datasets.  To obtain the score, we first compute similarity matrices between relative representations of the two spaces as $\mathbf{D}(\mathbf{Z}_1,\mathbf{Z}_2)$ where $\mathbf{D}_{i,j}=\frac{RR(\mathbf{Z}_1)_i^T RR(\mathbf{Z}_2)_{j}}{\|RR(\mathbf{Z}_1)_i\|_2 \|RR(\mathbf{Z}_2)_{j}\|_2}$. Then we compute the Mean Reciprocal Rank (MRR, see Appendix~\ref{app:MRR}) on top of the similarity matrix. In the figure, we plot MRR as a function of a random set of anchors, where the shaded areas indicate the standard deviations over 5 different sets of random anchors with the same cardinality. Our method consistently performs better than \citet{Moschella2023}, saturating the score with few anchors on all the domains, despite the different degrees of complexity of the latent spaces. In addition, our method shows significantly less variance, being more robust to the choice of the anchor set.

\textbf{Takeaway.} Relative geodesic representation near-perfectly captures transformations between representational spaces of models initialized differently, sample efficiency and robustness. %

\subsubsection{Stitching autoencoder models}

\begin{figure}[h!]
    \centering
\begin{overpic}[width=\linewidth,trim={9cm 5cm 7cm 4cm},clip]{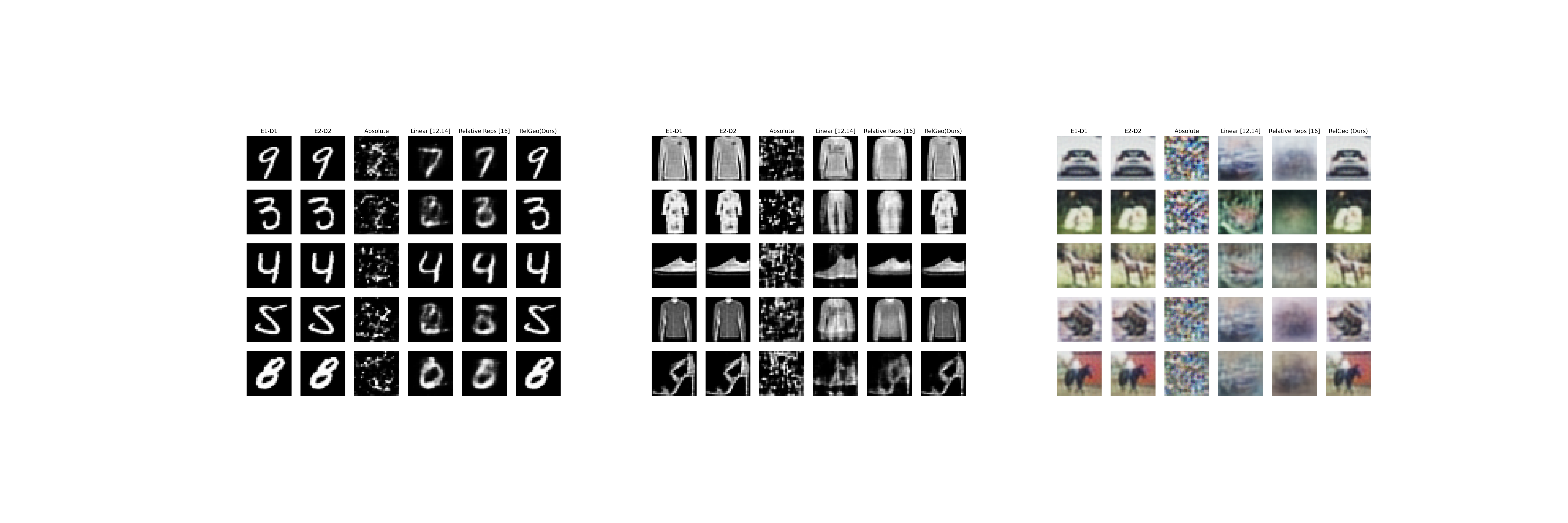}
\end{overpic}
    \caption{\emph{Stitching on Autoencoders}: We visualize qualitative reconstructions of samples, stitching autoencoders of models trained with different initializations on \texttt{MNIST} (left), \texttt{FashionMNIST} (center), \texttt{CIFAR10} (right). The first two columns show reconstructions from the original models; middle three columns represent baselines \citep{maiorca2024latent,Moschella2023}; the rightmost column is our method. Relative geodesic yields the best stitching results using just 5 anchors.}
    \label{fig:autoencoder_qualitative}
\end{figure}

\textbf{Setting.~}
We consider the same pairs of autoencoders trained on the \texttt{MNIST}, \texttt{FashionMNIST}, \texttt{CIFAR10} datasets of Section~\ref{sec:exp_mrr_autoencoders}. 
Starting from a set of five random anchors, we estimate a transformation $T$ between the model representational spaces $Z_1,Z_2$. In this experiment, to keep differently from \citet{Moschella2023}, in which zero-shot stitching was achieved by training once a decoder module with relative representations and then exchanging different encoder modules, here we achieve stitching without training any decoder. We compute relative representations with respect to the set of anchors, and compute a similarity matrix $\mathbf{D}(\mathbf{Z}_1,\mathbf{Z}_2)$. Then we compute the vector $\mathbf{c}= \argmax_i(\mathbf{D})$ representing a correspondence between the two representation matrices $\mathbf{Z}_1$, $\mathbf{Z}_2$, and use $c$ to fit a linear transformation $T$ to approximate the transformation between the two domains. We perform stitching by performing the following operation for a sample $x \in \mathcal{X}$: $ \tilde{x}= D_2 \circ T \circ E_1(x)$.  %

\textbf{Analysis of results.~}
We visualize the results of reconstructions of random samples in Fig.~\ref{fig:autoencoder_qualitative}, comparing against \citet{Moschella2023,lahner2023direct,maiorca2024latent}.
For each dataset, each column represents respectively: (i) the original autoencoding mapping for a sample $x$ of model $F_1$, $D_1( E_1(x))$, (ii) $D_2( E_2(x))$, (iii)  the mapping $D_2( E_1(x))$, (iv) the mapping $D_2( T_{anchors} E_1(x))$ where  $T_{anchors}$ is estimated on the five available anchors, (v) the mapping $D_2( T_{cosine} E_1(x))$ where $T_{cosine}$ is estimated among all 10k samples with the correspondence $c$ obtaining in the relative space of \citet{Moschella2023}, (vi) Our result $D_2( T_{relgeo} E_1(x))$, where $T_{relgeo}$ is estimated from the correspondence obtained in the relative geodesic space. As shown in Fig.~\ref{fig:autoencoder_qualitative}, while the baselines do not reach a good enough reconstruction quality, reconstructions with our method are almost perfect in accordance with the results in Fig.~\ref{fig:mrr_autoencoders}.

\textbf{Takeaway.} Relative geodesics enable stitching of neural modules trained with different initializations.

\subsection{Experiments on vision foundation models}
In this section, we evaluate relative geodesic representations' performances on retrieval and model stitching tasks on vision foundation discriminative models across models pretrained with different objectives, architectures, sizes and modalities.
\label{sec:experiments_on_vision_foundation_models}

\subsubsection{Matching representational spaces of discriminative foundation models}\label{sec:retrievals}

\begin{table}[t!]
  \centering
    \caption{Average MRR cosine results for different methods across different datasets. Relative representations pulling back from diet decoder (\texttt{RelGeo(Diet)}) consistently provides better retrievals.}
  \resizebox{\linewidth}{!}{%
    \begin{tabular}{lccccc}
      \rowcolor{gray!20}\toprule
      \textbf{Method} & \textbf{CIFAR-10} & \textbf{CIFAR-100} & \textbf{ImageNet-1k} & \textbf{CUB} & \textbf{SVHN} \\
      \midrule
      \texttt{Rel(Cosine)} \citep{Moschella2023}& $0.129 \pm 0.135$ & $0.166 \pm 0.162$ & $0.221 \pm 0.178$ & $0.135 \pm 0.148$ & $0.068 \pm 0.08$ \\
      \texttt{RelGeo($L2$)} & $0.047 \pm 0.013$ & $0.112 \pm 0.031$ & $0.412 \pm 0.09$ & $0.28 \pm 0.129$ & $0.025 \pm 0.012$ \\
      \texttt{RelGeo(Diet)} & $\mathbf{0.387} \pm 0.145$ & $\mathbf{0.445} \pm 0.142$ & $\mathbf{0.566} \pm 0.111$ & $\mathbf{0.523} \pm 0.177$ & $\mathbf{0.314} \pm 0.188$ \\
      \bottomrule
    \end{tabular}
  }
\label{tbl:main_mrr_cosine_sym}
\end{table}

In this section, we test the compatibilities of representations of vision foundation models with different architectures, such as residual networks \citep{he2016deep} and vision transformers \citep{Dosovitskiy2021}, and with different pretraining objectives including classification and self-supervised learning.

\textbf{Setting.~}
We perform experiments on retrieval tasks on pretrained vision foundation models, investigating how well we can match representations together with different backbones subject to the decoding tasks, on 5 datasets, varying in complexity and size: \texttt{CIFAR10}, \texttt{CIFAR100} \citep{krizhevsky2009learning}, \texttt{SVHN} \citep{yuval2011svhn}, \texttt{CUB} \citep{wah2023cub}, and \texttt{ImageNet-1k} \citep{russakovsky2015imagenet}. For ImageNet-1k, we used $1000$ anchors, while for other datasets we used $500$. As backbones we consider ResNet-50 \citep{he2016deep}, Vision Transformers (ViT) \citep{Dosovitskiy2021} with both patch 16-224 and patch 32-384, and DINOv2 \citep{Oquab2024}. 
We compare the original formulation of relative representations with cosine similarity \citep{Moschella2023} denoted as \texttt{Rel(Cosine)}, relative geodesic representations pulling back from Euclidean logits denoted as \texttt{RelGeo($L2$)}, and pulling back the spherical metric using a Diet decoder denoted as \texttt{RelGeo(Diet)}.

\begin{figure}[h]
  \centering
  \begin{subfigure}[t]{\textwidth}
    \centering
    \includegraphics[width=0.8\linewidth]{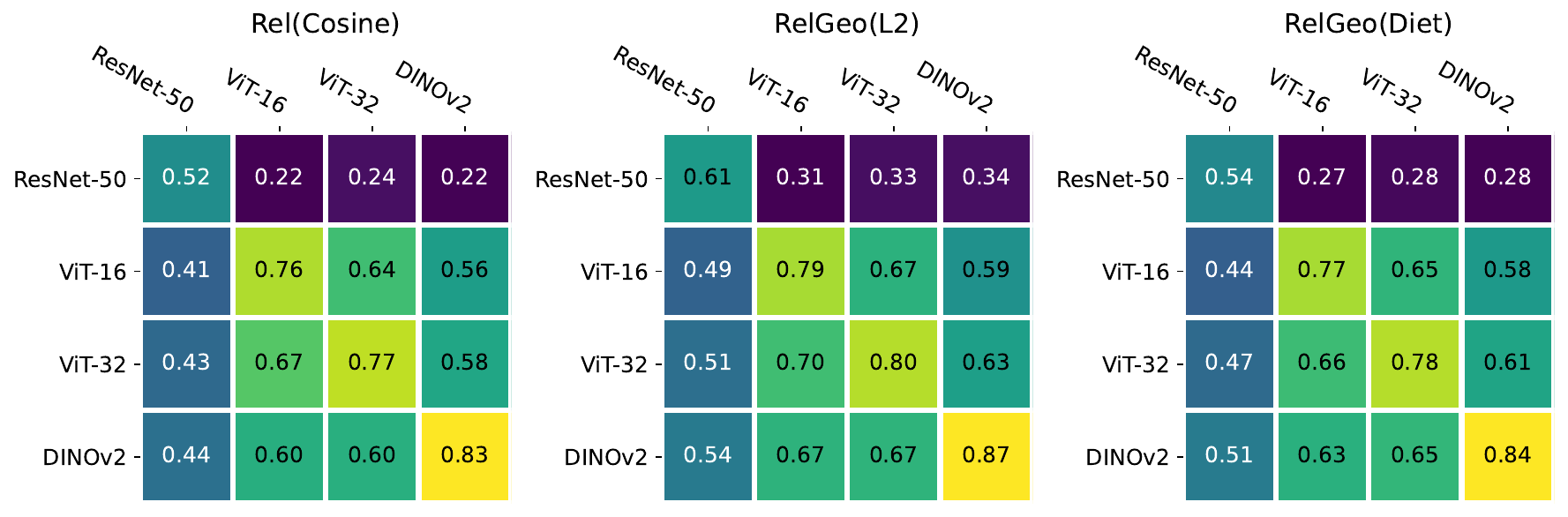}
  \end{subfigure}
  \hfill
  \begin{subfigure}[t]{\textwidth}
    \centering
    \includegraphics[width=0.8\linewidth]{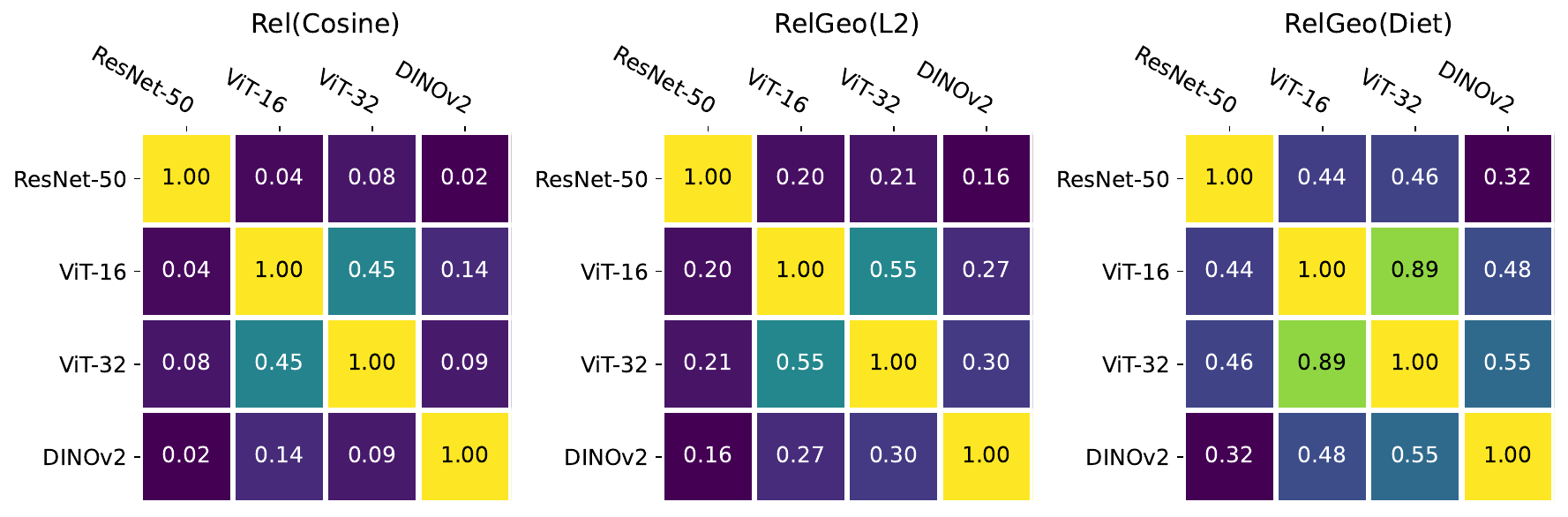}
  \end{subfigure}
  \caption{CUB Accuracies (top) and symmetricized MRR cosine (bottom). \texttt{RelGeo(Diet)} and especially \texttt{RelGeo($L2$)} provide strong stitching accuracies, while \texttt{RelGeo(Diet)} maintains strong instance identifiability.}
  \label{fig:main_cub_results}
\end{figure}

\textbf{Analysis of results.~} 
Table~\ref{tbl:main_mrr_cosine_sym} shows results from different methods averaged across all possible pairs of models  on the considered datasets. Additionally, Fig.~\ref{fig:main_cub_results} shows the results on CUB. While \texttt{RelGeo($L2$)} may result in worse MRR numbers, \texttt{RelGeo(Diet)} provides consistently improved retrieval performance. In Appendix~\ref{app:full_results} we report full results for the datasets.

\textbf{Takeaway.} Relative geodesic representations pulling back from instance discrimination decoders are identifiable across vision foundation models, improving retrieval performances.

\subsubsection{Zero-shot stitching of vision foundation models}
\begin{table*}[h!]
  \centering
  \caption{Average stitching performances across different settings. \texttt{RelGeo($L2$)} often outperforms \texttt{Rel(Cosine)}, while \texttt{RelGeo(Diet)} remains competitive.}
   \resizebox{\linewidth}{!}{%
  \begin{tabular}{lccccc}
    \rowcolor{gray!20}\toprule
    \textbf{Method} & \textbf{CIFAR-10} & \textbf{CIFAR-100} & \textbf{ImageNet-1k} & \textbf{CUB} & \textbf{SVHN} \\
    \midrule
    \texttt{Rel(Cosine)} \citep{Moschella2023}       & $0.907 \pm 0.09$ & $0.775 \pm 0.132$ & $\mathbf{0.549} \pm 0.152$ & $0.531 \pm 0.188$ & $0.384 \pm 0.115$ \\
    \texttt{RelGeo($L2$)}  & $\mathbf{0.955} \pm 0.03$ & $\mathbf{0.874} \pm 0.055$ & $0.501 \pm 0.159$ & $\mathbf{0.595} \pm 0.163$ & $\mathbf{0.59} \pm 0.054$ \\
    \texttt{RelGeo(Diet)}      & $0.915 \pm 0.074$ & $0.775 \pm 0.115$ & $0.479 \pm 0.17$ & $0.559 \pm 0.171$ & $0.416 \pm 0.079$ \\
    \bottomrule
  \end{tabular}
  }
  \label{tbl:main_stitching}
\end{table*}

Model stitching was introduced in \citet{lenc2015understandingimagerepresentationsmeasuring} to analyze neural network representational spaces, by training a linear layer to connect different layers and evaluating performance. Here we sidestep the need for trainable stitching layers and consider the zero-shot model stitching task defined in \citet{Moschella2023} to effectively test how components of vision foundation models can be reused. To do this, we leverage the space of relative geodesic representations as a shared compatible space. For the $i$th model $E_i$, we train one decoder $D_i$ on the relative representations induced by it, then evaluate the performance of using $D_i$ to decode the representations of model $E_j$, where $E_j$ may be a different model. This assesses how much two representation spaces can be merged with respect to the task defined by the decoder $D$, e.g., a classification head.

\textbf{Setting.~}
We perform experiments on pretrained vision foundation models from Hugging Face Transformers \citep{wolf2020transformers}, investigating how well we can match representations together for classification with different backbones with classification heads, on the same datasets and models as considered in Section~\ref{sec:retrievals},
similarly comparing \texttt{Rel(Cosine)}, \texttt{RelGeo($L2$)} and \texttt{RelGeo(Diet)}.

\textbf{Analysis of results.~}
The results of the different methods across the different datasets are shown in Table~\ref{tbl:main_stitching}, where we average over all possible model pairs. We further show the accuracies of the models on the CUB dataset in Fig.~\ref{fig:main_cub_results}. Both \texttt{RelGeo($L2$)} and \texttt{RelGeo(Diet)} provide strong stitching accuracies, with \texttt{RelGeo($L2$)} reflecting the benefits of pulling back class specific information. \texttt{RelGeo(Diet)} still results in good accuracies while having very strong MRR metrics, as shown in \ref{tbl:main_mrr_cosine_sym}.

\textbf{Takeaway.} Relative geodesic representations yield good accuracies and good MRRs, avoiding downgrading of performance when performing model stitching while retaining sample identifiability. %

\subsubsection{Matching different modalities}

In this section we evaluate relative geodesic representations in the multimodal setting.

\begin{wrapfigure}[13]{r}{0.49
\linewidth}
  \centering
  \begin{subfigure}[t]{\linewidth}
    \centering
    \includegraphics[width=\linewidth]{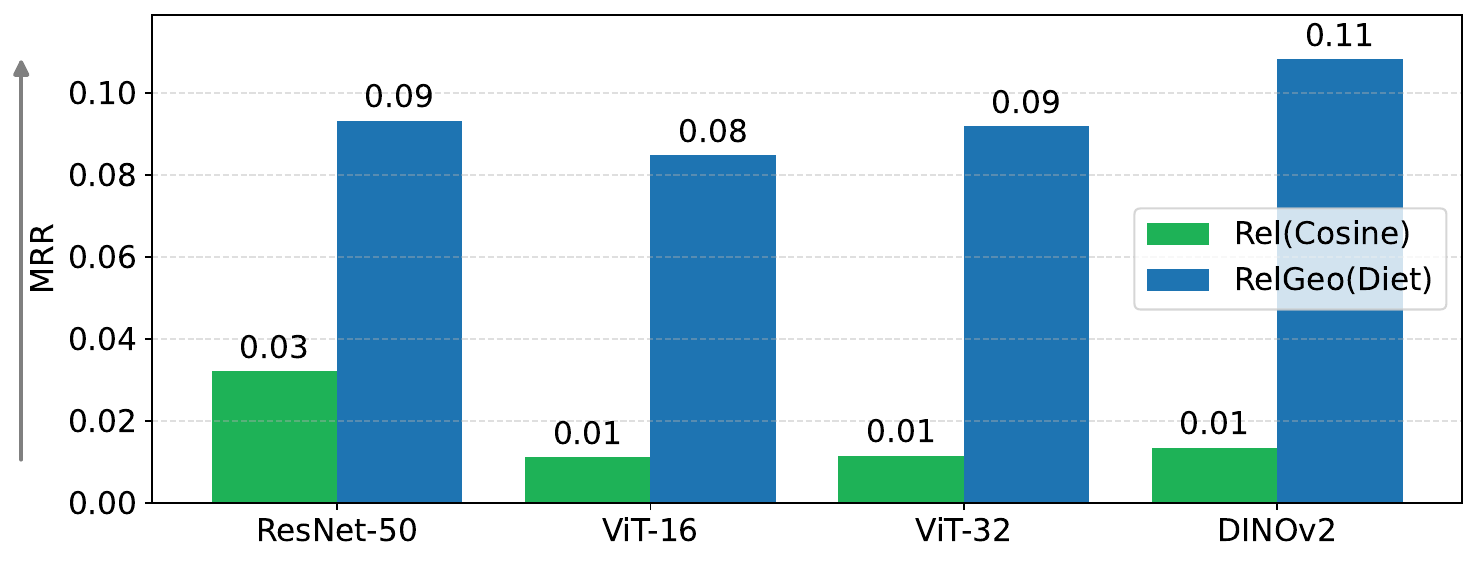}
  \end{subfigure}
  \caption{\emph{Matching multimodal models.} Symmetricized MRR cosine on Flickr30k. RelGeo(Diet) substantially improves upon Rel(Cosine) in aligning multimodal models.}
  \label{fig:main_multimodal}
\end{wrapfigure}
\textbf{Setting.~}
We study the retrieval task in terms of vision foundation models and the text encoders of CLIP \citep{radford2021learning} with both patch 16 and patch 32, using Flickr30k dataset \citep{young2014flickr30k}.  Keeping the text encoder of CLIP fixed, we swap the vision encoder with the ones of ResNet-50, DINOv2 and ViT, including different patch and model sizes. Due to the lack of class labels, \texttt{RelGeo($L2$)} is not applicable, and we compare \texttt{RelGeo(Diet)} with \texttt{Rel(Cosine)}. While we observed that using data augmentations is beneficial for \texttt{RelGeo(Diet)}, due to the lack of a principled approach to construct data augmentations on texts corresponding to image augmentations, we do not employ augmentations.

\textbf{Analysis of results.~} The results in terms of symmetricized MRR cosine metric with CLIP with patch 16 are shown in Figure~\ref{fig:main_multimodal}. We observe that RelGeo(Diet) yields significantly improved stitching performances upon Rel(Cosine). In Appendix~\ref{app:full_multimodal_results} we show the full pairwise matrices of MRR, comprising of the unimodal performances, inter vision models, and text models. 

\textbf{Takeaway.} Relative geodesic representations show promising results for obtaining identifiable representations in multimodal scenarios.

\section{Conclusions and discussion }\label{sec:conclusion_and_discussion}

We have introduced the framework of relative geodesic representation starting from the assumption that distinct neural models trained on similar data distributions learn to approximate the same underlying latent manifold. As a result, geodesic distances based on their representations are invariant to transformations between different representational spaces. We show that the geodesic energy and arc length of straight lines provide an efficient, low-cost metric for bridging these spaces, allowing us to measure similarity and align representations across different architectures, training objectives, and training procedures, while outperforming previous methods.

\textbf{Limitations and future work.~}
The accuracy of approximating geodesics using straight-line arc length (or energy) can deteriorate in regions of high curvature in the latent space, typically corresponding to areas far from the support of the training data. Moreover, this could require increasingly smaller step sizes, hurting the efficiency performance of the method. This suggests exploring nonlinear paths, and adaptive step sizes, e.g., by estimating the support of the data building KNN graphs in the latent space and forcing the path to not deviate too much from them. 
By employing the pullback metric from a given output space, the relative geodesic representation has the interesting property of restricting the alignment problem to the information relevant to the decoding task. This could be useful to \emph{(i)} further explore no training multi-modal alignment \citep{norelli2023asifcoupleddataturns}, where it is of interest to capture not only the shared information across modalities, but also the modality-specific information; \emph{(ii)} to better understand the relation between the representation similarity and decodability \citep{harvey2024representational} and the interaction between tasks and learned representations \citep{fumero2023leveraging}.

\begin{ack}
We thank Gregor Krzmanc, German Magai, Vital Fernandez for insightful discussions in the early stages of the project. HY was supported by the Research Council of Finland Flagship programme: Finnish Center for Artificial Intelligence FCAI. HY wishes to acknowledge CSC - IT Center for Science, Finland, for computational resources.
GA was supported by the DFF Sapere Aude Starting Grant ``GADL''.
SH was supported by a research grant (42062) from VILLUM FONDEN and partly funded by the Novo Nordisk Foundation through the Center for Basic Research in Life Science (NNF20OC0062606). SH received funding from the European Research Council (ERC) under the European Union’s Horizon Programme (grant agreement 101125003).
MF is supported by the MSCA IST-Bridge fellowship which has received funding from the European Union’s Horizon 2020 research and innovation program under the Marie Skłodowska-Curie grant agreement No 101034413.
\end{ack}

\bibliography{main}
\bibliographystyle{abbrvnat}

\newpage
\appendix
\section{Appendix}

\subsection{Proof of theoretical results}

\subsubsection{Proof of Proposition~\ref{thm:inv_rep}}
\label{app:proof_inv_rep}

\begin{proof}

We first prove the first half, i.e. the invariance of Riemannian curve length and energy across change of coordinates of the manifold. This can be proven by observing that the inner product at a point along the curve is invariant under such change, where we refer to $\bm{\gamma}$ and $\bm{\gamma}'$ as $x$ and $x'$:

\begin{align*}
\bigl\lVert\bm{\dot{x}}\bigr\rVert_{G(x)}^2 &= \bm{\dot{x}}^{\top} G\bigl(\bm{x}\bigr) \bm{\dot{x}} = \bigl(\tfrac{\partial \bm{x}}{\partial \bm{x'}}\bm{\dot{x'}}\bigr)^{\top} \bigl(\tfrac{\partial \bm{x'}}{\partial \bm{x}}\bigr)^{\top} G'\bigl(\bm{x'}\bigr) \tfrac{\partial \bm{x'}}{\partial \bm{x}} \tfrac{\partial \bm{x}}{\partial \bm{x'}}\bm{\dot{x'}} \\
&= \bm{\dot{x'}}^{\top} G'\bigl(\bm{x'}\bigr) \bm{\dot{x'}} = \bigl\lVert\bm{\dot{x'}}\bigr\rVert_{G'(x')}.
\end{align*}

As such, the length and the energy of the same curves on the manifold expressed in different coordinates are integrals of the same quantities, hence are equal.

We then prove the second half, i.e. the invariance of Riemannian curve length across reparameterizations of the curve. Based on Equation 4.7 from \citet{dggm}, we have

\begin{align*}
L[\bm{\gamma}] &= \int_{0}^{1}\bigl\lVert\tfrac{d\bm{\gamma}}{d t}\bigr\rVert_{G(\bm{\gamma}(t))}\,d t = \int_{0}^{1}\bigl\lVert\tfrac{d \bm{\gamma}}{d \varphi}\bigr\rVert_{G(\tilde{\bm{\gamma}}(\tau))} \tfrac{d \varphi}{d \tau}\,d \tau \\
&= \int_{0}^{1}\bigl\lVert\tfrac{d \bm{\gamma}}{d \varphi}\tfrac{d \varphi}{d \tau}\bigr\rVert_{G(\tilde{\bm{\gamma}}(\tau))} \,d \tau = \int_{0}^{1}\bigl\lVert\tfrac{d\tilde{\bm{\gamma}}}{d \tau}\bigr\rVert_{G(\tilde{\bm{\gamma}}(\tau))} \,d \tau \\
&= L[\tilde{\bm{\gamma}}].
\end{align*}

\end{proof}

\subsubsection{Proof of Equation~\ref{eq:two_bounds}}
\label{app:proof_bounds}

\begin{proof}

We first prove $d(\bm{z}_0, \bm{z}_1)^2 \leq L^2(\tilde{\bm{\gamma}})$, then prove $L^2(\tilde{\bm{\gamma}}) \leq 2\mathcal{E}(\tilde{\bm{\gamma}})$.

For the first part, according to the definition of geodesic distance, we have $d(\bm{z}_0, \bm{z}_1) \leq L(\tilde{\bm{\gamma}})$ and, as such, $d(\bm{z}_0, \bm{z}_1)^2 \leq L^2(\tilde{\bm{\gamma}})$.

The second part involves $L^2(\tilde{\bm{\gamma}}) \leq 2\mathcal{E}(\tilde{\bm{\gamma}})$, which can be proven using the Cauchy-Schwarz inequality. See Equation 7.14 in \citep{dggm}, where we denote $\bm{u}_{t} = \tfrac{d\tilde{\bm{\gamma}}}{d t} $ and $\bm{v}_{t} = 1 $.
\begin{align*}
L^2(\tilde{\bm{\gamma}}) & = \int_{0}^{1}\bigl\lVert\tfrac{d\tilde{\bm{\gamma}}}{d t}\bigr\rVert\,d t = \bigl\langle \bm{u},\bm{v} \bigr\rangle \leq \bigl\lVert \bm{u} \bigr\rVert \bigl\lVert \bm{v} \bigr\rVert 
&= \sqrt{\int_{0}^{1}\bigl\lVert \tfrac{d\tilde{\bm{\gamma}}}{d t} \bigr\rVert^{2}\, d t }\sqrt{\int_{0}^{1}1^{2}\, d t} = \sqrt{\int_{0}^{1}\bigl\lVert \tfrac{d\tilde{\bm{\gamma}}}{d t} \bigr\rVert^{2}\, d t } \\
&= 2\mathcal{E}(\tilde{\bm{\gamma}}).
\end{align*}

\end{proof}

\subsection{Additional explanations}

\subsubsection{Details on Diet}
\label{app:diet}

Proposed as a self-supervised learning method \citep{ibrahim2024occams}, Diet was also shown to yield interesting identifiable guarantees \citep{reizinger2025cross-entropy}, laying the theoretical foundation for \texttt{RelGeo(Diet)}, where we employ the resulting geometry.

One can consider such a scenario \citep{reizinger2025cross-entropy}: some latent variables $\bm{z}$ are drawn from a vMF distribution, and pushed forward through a continuous and injective generator function $g$ to obtain the data $\bm{x}$. Remarkably, given only $\bm{x}$ without the knowledge of $g$, it is possible to (to some degree) recover the latent variables $\bm{z}$ through parameterizing a model and optimizing the instance discrimination loss as given in Equation~\ref{eq:diet_loss}. Specifically, suppose there is a finite set of vectors $\bm{v}_{c}$ on a unit sphere, each representing a class, and a finite set of instances. One instance belongs to exactly one class, and every class is employed by some instance. Additionally, the instance labels are chosen uniformly, and the latent variables $\bm{z}$ are drawn from a vMF distribution centered around the corresponding cluster vector $\bm{v}_{c}$ with concentration parameter $\kappa$.

Then, after the model is trained using the loss function as in Equation~\ref{eq:diet_loss}, when both $f$ and $w$ are not unit-normalized, $f \circ g$ is linear. This can be proven rigorously by expanding upon the theoretical framework of non-linear ICA \citep{hyvarinen2023nonlinear_ica}. As such, we propose to utilize $f \circ g$ to form the representations. For further technical details on the assumptions and additional results, we refer interested readers to \citet{reizinger2025cross-entropy}.

Assuming spherical geometry, the distance between two points $\bm{x}$ and $\bm{y}$ can be computed as
\begin{equation*}
d(\bm{x},\bm{y}) = \arccos\left(\frac{\bm{x}^{\top}\bm{y}}{\norm{\bm{x}}\norm{\bm{y}}}\right).
\end{equation*}
In the above formula, points that do not precisely lie on the unit sphere are effectively projected onto it. Interestingly this bears a strong resemblance to the cosine distances as used in the original paper on relative representations \citep{Moschella2023}.

\subsubsection{Why it works}

Prior work on representational alignment has shown that representations from different models can often be approximately aligned using simple transformations, e.g. linear, orthogonal or locally linear maps. Even when models are trained independently -- with differrent architectures, modalities, or datasets that nonetheless share an underlying structure -- they tend to learn similar representations, suggesting convergence towards a shared encoding of entities \citep{huh2024platonicrepresentationhypothesis}. From a theoretical standpoint, identifiability results \citep{roeder2021linear} imply that if two discriminative models learn the same likelihood function, their internal representations must be equivalent up to a linear transformation. However, this ideal scenario rarely holds exactly in practice: training dynamics, nuisance factors, and unmodeled variability can all introduce distortion. In our case, it may be too strong to assume that two models learn different parameterizations of an identical manifold. Instead, we adopt a weaker assumption, that they do so up to some bounded distortion. Recent theoretical work has begun to explore relaxations of strict identifiability to account for such bounded distortions \citep{nielsen2025when}. Integrating these relaxations into our Riemannian framework presents a promising direction for future work.

In general, the few theoretical results available, e.g. \citep{roeder2021linear}, often rely on unrealistic assumptions, e.g. proofs in axiomatic settings, infinite-data regimes, or the requirement that two models learn exactly the same likelihood function. In our view, a meaningful first step toward bringing theory and practice is to relax these assumptions, as initiated in \citep{nielsen2025when}, and begin to model more realistic scenarios, e.g. including the dynamics introduced by model training.

We believe that Riemannian methods can play a key role in this direction to try to capture local alignments beyond linear global transformations of the space as considered in \citet{roeder2021linear} and possibly accounting for distortions measured in the linear space in practice. Nevertheless, we remark that neural networks could find qualitatively different solutions \citep{pascanu2025optimizers} and that the union of manifolds hypothesis might be more appropriate for modeling image data \citep{brown2023verifying}.

\subsection{Additional details}
\label{sec:additional_details}

\subsubsection{Mean Reciprocal Rank} \label{app:MRR}
Mean Reciprocal Rank (MRR) is a commonly used metric to evaluate the performance of retrieval systems, and has been used to evaluate the capabilities of representations for instance discrimination \citep{Moschella2023}. It measures the effectiveness of a system by calculating the rank of the first relevant item in the search results for each query.

To compute MRR, we consider the following steps:
\begin{enumerate}
    \item For each query, rank the list of retrieved items based on their relevance to the query.
    \item Determine the rank position of the first relevant item in the list. If the first relevant item for query $i$ is found at rank position $r_i$, then the reciprocal rank for that query is $\frac{1}{r_i}$.
    \item Calculate the mean of the reciprocal ranks over all queries. If there are $Q$ queries, the MRR is given by:
    \begin{equation*}
    \text{MRR} = \frac{1}{Q} \sum_{i=1}^{Q} \frac{1}{r_i}.
    \end{equation*}
    
    Here, $r_i$ is the rank position of the first relevant item for the $i$-th query. If a query has no relevant items in the retrieved list, its reciprocal rank is considered to be zero.
\end{enumerate}
MRR provides a single metric that reflects the average performance of the retrieval system, with higher MRR values indicating better performance.

Similar to stitching accuracies, MRR is generally asymmetric. However, it can also be made symmetric. Specifically, as MRR is calculated based on a distance matrix $D$, one can make the distance matrix symmetric by setting $D=\frac{1}{2}\left(D^{\top}+D\right)$. In Section~\ref{sec:retrievals} we reported the symmetric version. Otherwise we report both the original version and the symmetric version, and discriminate between these two by explicitly indicating it when it is symmetric.

\subsubsection{Architectural details}
\label{app:arch}
We provide in Table~\ref{tab:tiny_arch} the architectural details of the convolutional autoencoders employed in experiments in Figures \ref{fig:mrr_autoencoders} and \ref{fig:autoencoder_qualitative}.

\begin{table}[ht]
    \centering
    \begin{center}
    \caption{Architecture of the convolutional autoencoders.}
    \label{tab:tiny_arch}
    \begin{small}
    \begin{tabular}{p{5cm}}
        \toprule
        \multicolumn{1}{c}{Encoder}\\
        \midrule
        $3\times 3$ conv. 32 stride 2-ReLu \\
        $3\times 3$ conv. 64 stride 2-ReLu\\
        Flatten\\
        $(64*k*k) \times h$  Linear\\
        Latents\\
        \toprule
        \multicolumn{1}{c}{Decoder}\\
        \midrule
        $h \times  (64*k*k)$ Linear\\
        Unflatten\\
        $3\times 3$ conv. 64 stride 2-ReLu \\
        $3\times 3$ conv. 32 stride 2-ReLu\\
        Sigmoid \\
        \bottomrule
    \end{tabular}
    \end{small}
    \end{center}
\end{table}

For the classifier experiment, in order to obtain geometric representations we need a decoder. The architecture is shown in Table~\ref{tab:simple-cla}. For RelGeo(Diet), the last linear layer is configured with \textit{bias=False} in accordance with the original algorithm.

For evaluating the performances of the representations, we train a classification head with the same architecture as used by \citet{Moschella2023} as given in Table~\ref{tab:final-cla}.

\begin{table}[ht]
\centering
\begin{minipage}{0.45\textwidth}
    \centering
    \caption{Architecture of the simple decoders.}
    \label{tab:simple-cla}
    \begin{small}
    \begin{tabular}{p{5cm}}
        \toprule
        \multicolumn{1}{c}{Classification head}\\
        \midrule
        $input\_dim$ LayerNorm \\
        $input\_dim \times 500$ Linear-Tanh \\
        $500 \times num\_classes$ Linear \\
        \bottomrule
    \end{tabular}
    \end{small}
\end{minipage}
\hfill
\begin{minipage}{0.45\textwidth}
    \centering
    \caption{Architecture of the decoders for evaluations.}
    \label{tab:final-cla}
    \begin{small}
        \begin{tabular}{p{5cm}}
        \toprule
        \multicolumn{1}{c}{Final classification head}\\
        \midrule
        $input\_dim$ LayerNorm \\
        $input\_dim \times input\_dim$ Linear-Tanh \\
        InstanceNorm1d \\
        $input\_dim \times num\_classes$ Linear \\
        \bottomrule
    \end{tabular}
    \end{small}
\end{minipage}
\end{table}

\subsubsection{RelGeo(Diet) augmentations}

As noted by \citet{ibrahim2024occams}, it is beneficial to employ data augmentations when using Diet to perform self-supervised training of neural networks. We largely follow their approach, and considered different levels of data augmentations. Following \citet{ibrahim2024occams}, we consider different levels of data augmentations indexed by a scalar strength, which are summarized below using PyTorch pseudocode; strengths of a higher level employs the augmentations of lower levels as well.

0: No augmentations;

1: RandomResizedCrop((height, width)), RandomHorizontalFlip();

2: RandomApply(ColorJitter(0.4, 0.4, 0.4, 0.2)), p=0.3); RandomGrayscale(0.2);

3: RandomApply(GaussianBlur((3, 3), (1.0, 2.0)), p=0.2), RandomErasing(0.25).

\subsubsection{Compute resources}
\label{app:compute_resources}

Experiments regarding the geodesic approximation are conducted using NVIDIA A100 GPU and 12 CPU cores. Run time varies depending on the discretization steps, number of anchors and the used dataset.

The autoencoder stitching and retrieval experiments were conducted on a single NVIDIA RTX 3080TI GPU. Experiments involving vision foundation models were run on a compute cluster, each job using a single NVIDIA A100 GPU and 10 CPU cores, with runtimes of several hours. Preliminary experiments required additional resources, and in total we estimate having used several hundred GPU hours.

Further ablation studies on the running times can be found in Section~\ref{sec:running_times}.

\subsubsection{Geodesic approximation}
\label{app:geodesic_approximation}
\begin{figure}[ht]
  \centering
  \begin{subfigure}[t]{0.49\textwidth}
    \centering
    \begin{overpic}[width=\linewidth]{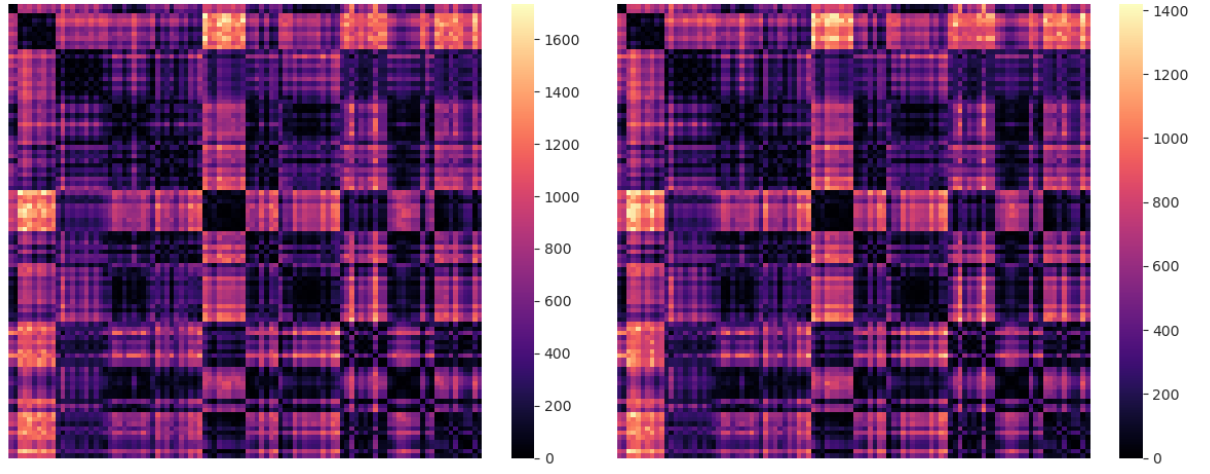}
    \put(1,42){\small{Approximate energies}}
    \put(54,42){\small{Geodesic energies}}
    \end{overpic}
    \caption{MNIST }
    \label{fig:mnist-10-dim2}
  \end{subfigure}
  \hfill
  \begin{subfigure}[t]{0.49\textwidth}
    \centering
    \begin{overpic}[width=\linewidth]{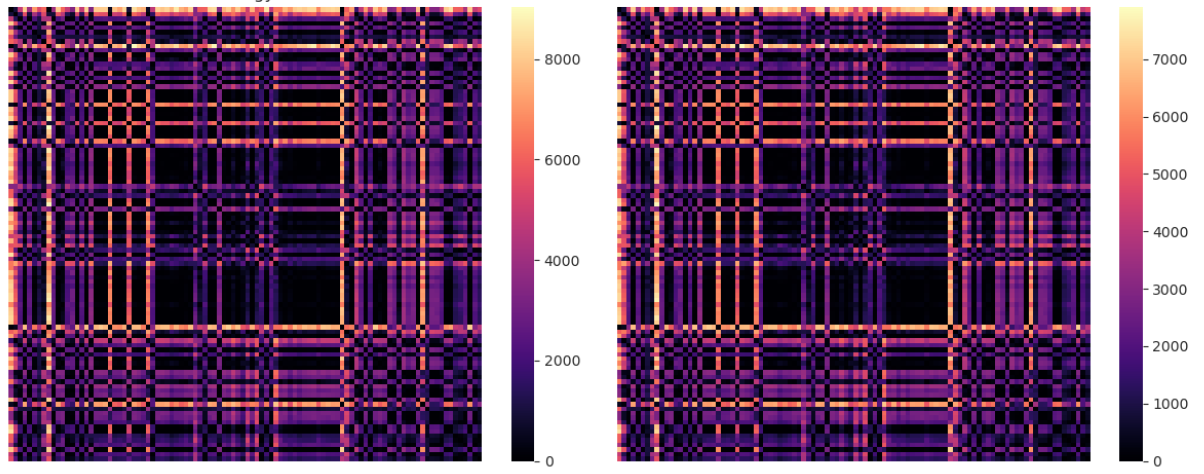}
    \put(1,42){\small{Approximate energies}}
    \put(54,42){\small{Geodesic energies}}
    \end{overpic}
    \caption{CIFAR-10 }
    \label{fig:cifar10-10-dim2}
  \end{subfigure}
  \caption{Pairwise latent‐space energy matrices for (a) MNIST and (b) CIFAR-10, with latent dimensionality 2. In each subfigure, the left heatmap shows the straight-line energy proxy and the right shows the full Riemannian geodesic energies. The Spearman rank correlation between the two measures is 0.99 for MNIST and \(\rho=1.00\) for CIFAR-10, demonstrating near-perfect agreement. }
  \label{fig:comparison-10dim2}
\end{figure}

Here, we provide the experimental details of the results presented in Fig. \ref{fig:comparison-10} and Fig. \ref{fig:comparison-10dim2}. To assess the geodesic energies, we used a small autoencoder, whose architecture is presented in Table \ref{tab:conv_autoencoder}.
\vspace{-.25cm}
\paragraph{Autoencoder training}
We trained a lightweight convolutional autoencoder (see Table \ref{tab:conv_autoencoder}) on both MNIST and CIFAR-10 to obtain the latent representations used in our experiments. For MNIST, the first convolutional layer was adjusted to accept a single input channel; for CIFAR-10 it used three channels. Each model was trained for 30 epochs using the Adam optimizer \citep{kingma2017adammethodstochasticoptimization} with a batch size of 64. We set the learning rate to 0.001, and fixed a random seed of 42 to ensure reproducibility.

\paragraph{Energy computation} After training, we selected 10 samples per class (100 total) in label order from each dataset and encoded them to produce their latent encodings. True geodesics are computed using Stochman library \citep{software:stochman}, which has Apache-2.0 license, which wraps the decoder into a pullback manifold, intializes a parameterized spline path between codes, and then optimizes its parameters to minimize the Riemannian energy. 
Geodesic energies are computed as in Eq. \ref{eq:energy}. 
Pairwise energies are computed and visualized in Figures \ref{fig:comparison-10} and \ref{fig:comparison-10dim2}, demonstrating the close agreement between the two measures under identical encoding and discretization settings. In Fig. \ref{fig:comparison-10}, latent dimensions for MNIST and CIFAR are 64 and 128 respectively, while in Fig. \ref{fig:comparison-10dim2}, latent dimension is 2 for both datasets.

\begin{table}[htbp]
  \centering
  \caption{ConvAutoencoder architecture (latent dim $d$).}
  \label{tab:conv_autoencoder}
  \begin{tabular}{ll}
    \toprule
    \textbf{Encoder} & Activation \\
    \midrule
    Conv2d(1, 32, kernel = 3, stride=2, pad=1)  & ReLU \\
    Conv2d(32, 64, kernel = 3, stride=2, pad=1) & ReLU \\
    Flatten                            & —    \\
    Linear(64*7*7, $d$)                  & —    \\[1ex]
    \midrule
    \textbf{Decoder} & Activation \\
    \midrule
    Linear($d$, 64*7*7)                  & ReLU \\
    Unflatten(64,7,7)                  & —    \\
    ConvTranspose2d(64, 32, kernel = 3, stride=2, pad=1, out\_pad=1) & ReLU \\
    ConvTranspose2d(32, 1, kernel = 3, stride=2, pad=1, out\_pad=1)  & Sigmoid \\
    \bottomrule
  \end{tabular}
\end{table}

\begin{figure}[ht]
  \centering
  \begin{subfigure}[t]{0.9\textwidth}
    \centering
    \includegraphics[width=\linewidth]{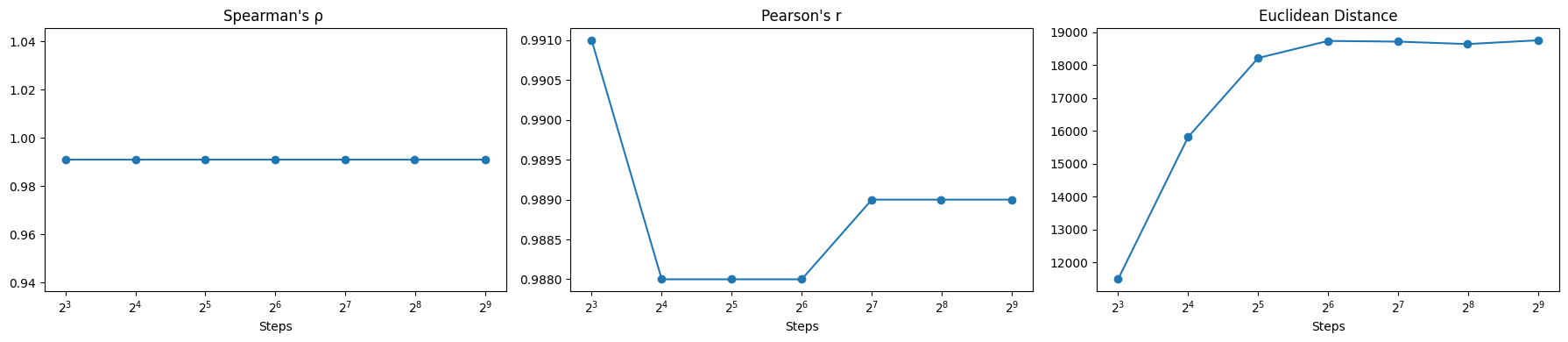}
    \caption{MNIST }
    \label{fig:mnist_discretization}
  \end{subfigure}
  \hfill
  \begin{subfigure}[t]{0.9\textwidth}
    \centering
    \includegraphics[width=\linewidth]{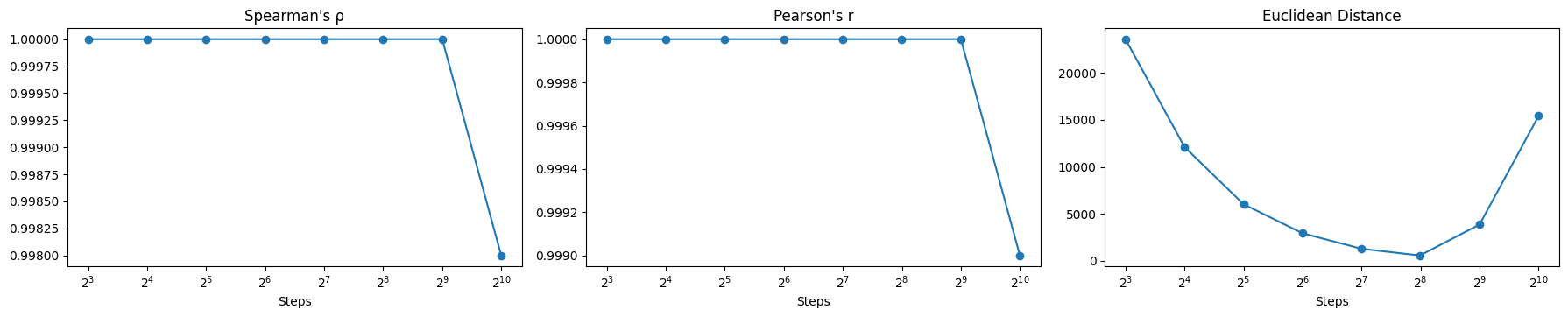}
    \caption{CIFAR-10 }
    \label{fig:cifar_discretization}
  \end{subfigure}
  \caption{Impact of varying discretization levels on similarity and energy metrics for (a) MNIST and (b) CIFAR‑10 datasets. Each subplot shows how Spearman’s $\rho$, Pearson’s $r$, and Euclidean distance change as the number of discretization levels increases.}
  \label{fig:comparison-10disc}
\end{figure}

\subsubsection{Autoencoder stitching and retrieval}

We provide the experimental details of the results presented in Figure \ref{fig:mrr_autoencoders} and Figure \ref{fig:autoencoder_qualitative}.
All models employed followed the architecture depicted in Table \ref{tab:conv_autoencoder}, with a latent dimensionality of 128.

We trained the lightweight convolutional autoencoder (see Table \ref{tab:conv_autoencoder}) on MNIST, CIFAR-10, FashionMNIST with 5 different seeds, to obtain the latent representations used in our experiments. For MNIST and FashionMNIST the first convolutional layer was adjusted to accept a single input channel; for CIFAR-10 it used three channels. Each model was trained for 50 epochs, reaching convergence, using the Adam optimizer \citep{kingma2017adammethodstochasticoptimization} with a batch size of 64. We set the learning rate to 0.001. 

\subsubsection{Vision foundation models}

We use the pretrained models as provided by Huggingface Transformers \citep{wolf2020transformers}, which has Apache-2.0 license, and the datasets as provided by HuggingFace Datasets \citep{lhoest2021datasets}, which also has Apache-2.0 license. The license information of the datasets are: CIFAR-10: unknown; CIFAR-100: unknown; CUB: unknown; ImageNet-1k: ImageNet agreement; SVHN: non-commercial use only.

Unless otherwise stated, we directly use the original test set of the dataset as the test set, while using $0.9$ of the original train set as the train set and the remaining as the validation set. Both the anchors and the Diet data points are selected from the validation set.

For CIFAR-100, we use the coarse labels. For SVHN, the objective is to predict the cropped digits. For CUB dataset, we use the version available at \url{https://huggingface.co/datasets/birder-project/CUB_200_2011-WDS}. Given the relatively small training set, we select $2000$ points as the validation set. When reporting aggregated MRR metrics in the tables, we always exclude the diagonal entries as these are generally (close to) $1$. For ImageNet-1k, we use the validation set and split it into the final train, val and test sets. Further details can be found in the provided code.

For all cases where we need to train classification heads, apart from the ones with Diet the heads are trained for $10$ epochs, while the ones with Diet are trained for $50$ epochs. The heads used to obtain the gometric information are trained using learning rate $5e-4$ and batch size $64$, while the heads used for stitching was trained using learning rate $1e-4$ and batch size $32$. We always use the Adam optimizer \citep{kingma2017adammethodstochasticoptimization}.

When reporting stitching results, we train three classification heads and average the accuracies as the final results.

\subsection{Additional results on Vision Foundation models}
\label{app:additional_results}

We provide additional results on vision foundation models. For ablation studies, we focus on the performances of the models on CUB dataset. We refer to accuracy as \textit{Accuracy}, symmetricized MRR based on cosine as \textit{MRR Cosine Sym}, symmetricized MRR based on cdist as \textit{MRR CDist Sym}, MRR based on cosine as \textit{MRR Cosine} and MRR based on cdist as \textit{MRR CDist}.

\subsubsection{Full results}
\label{app:full_results}

\begin{figure}[htbp]
  \centering
  \begin{subfigure}[t]{\textwidth}
    \centering
    \includegraphics[width=0.8\linewidth]{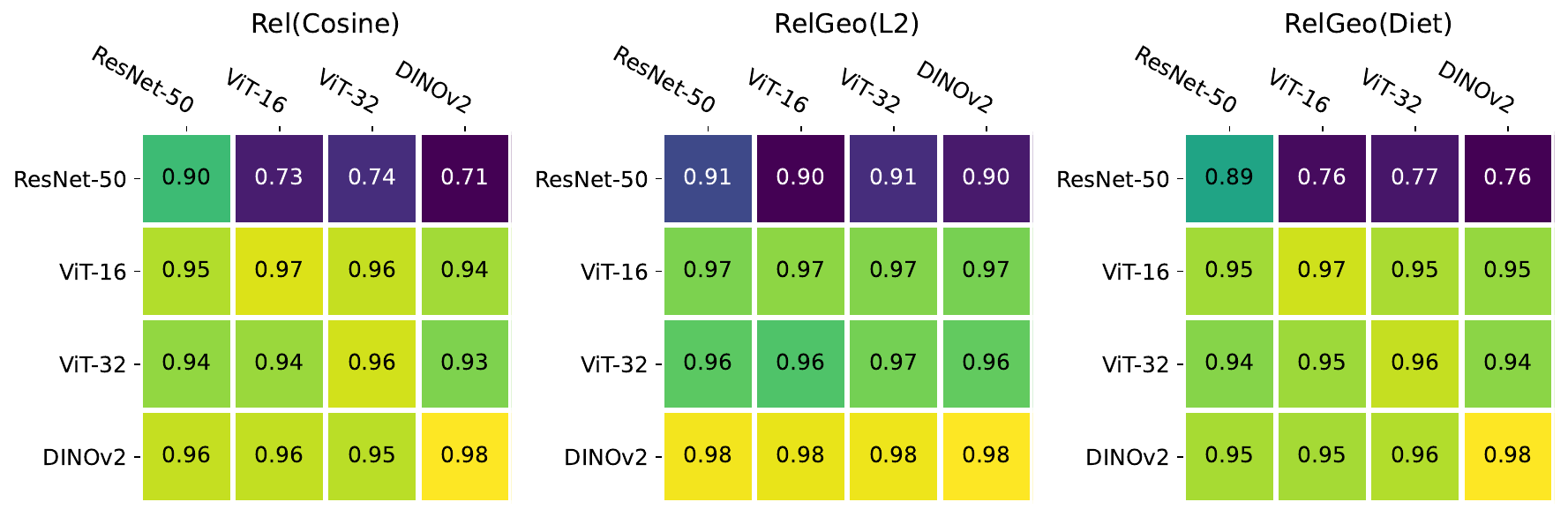}
  \end{subfigure}
  \hfill
  \begin{subfigure}[t]{\textwidth}
    \centering
    \includegraphics[width=0.8\linewidth]{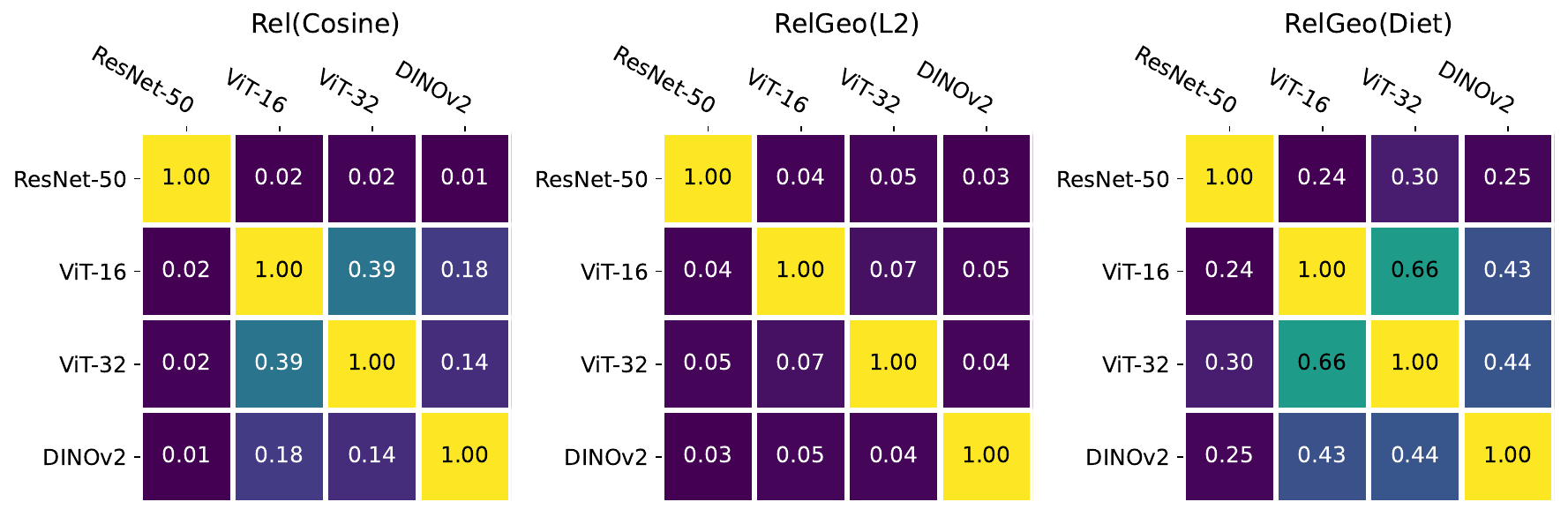}
  \end{subfigure}
  \hfill
  \begin{subfigure}[t]{\textwidth}
    \centering
    \includegraphics[width=0.8\linewidth]{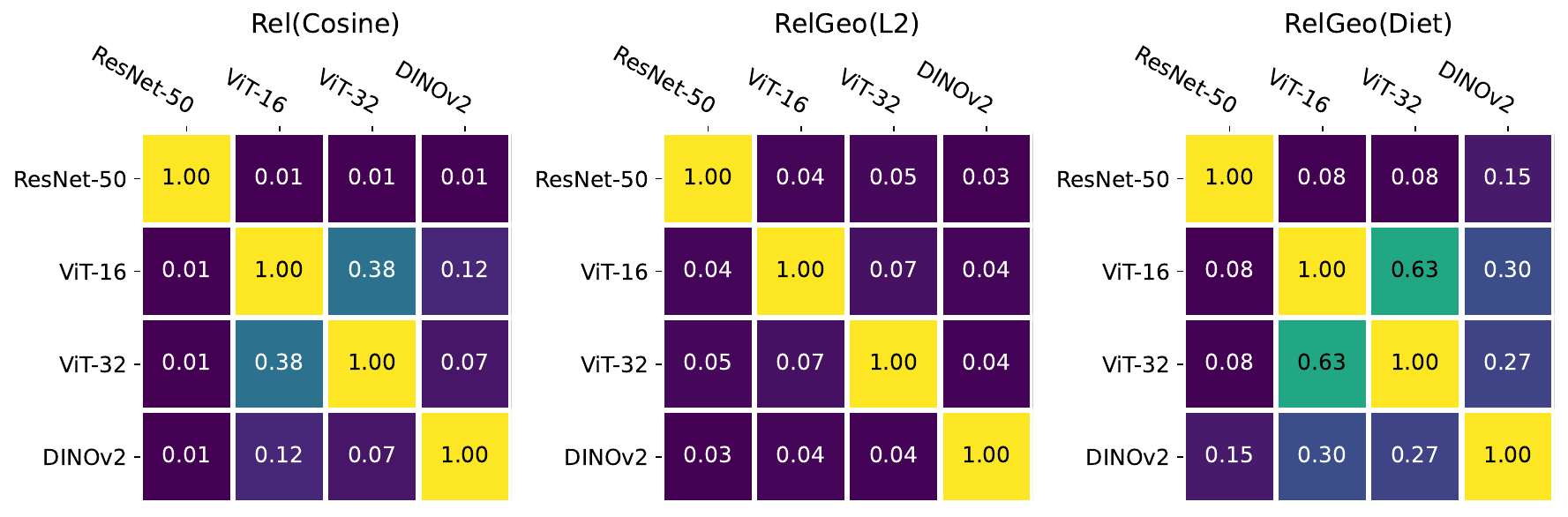}
  \end{subfigure}
  \hfill
  \begin{subfigure}[t]{\textwidth}
    \centering
    \includegraphics[width=0.8\linewidth]{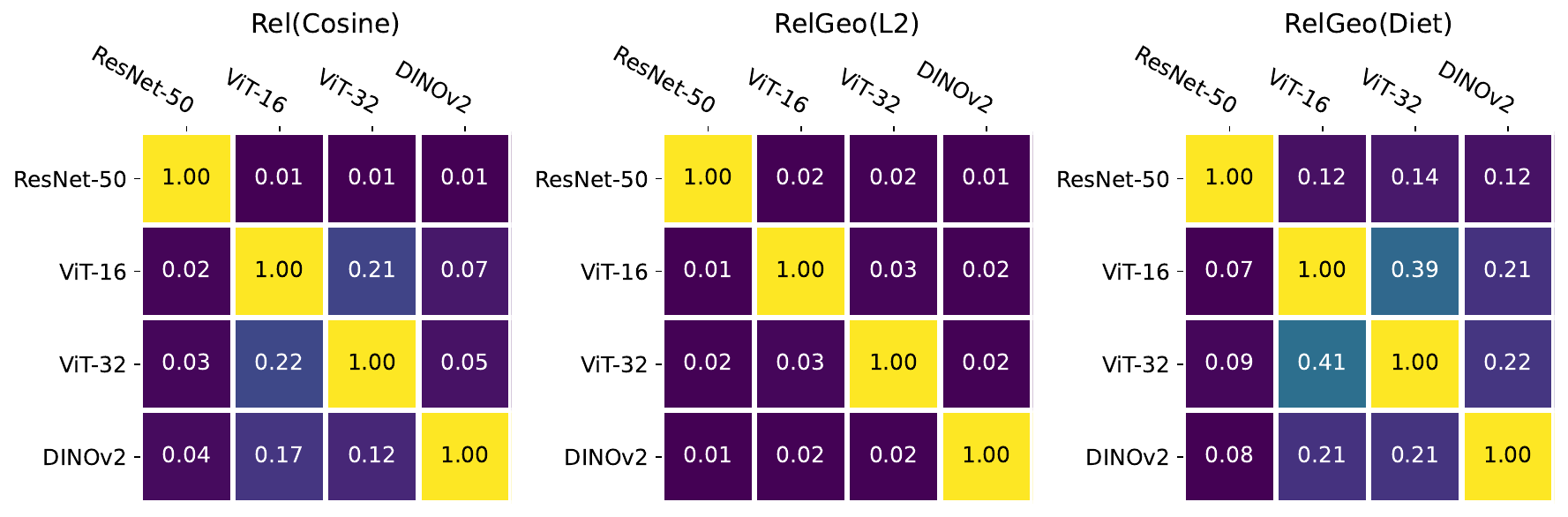}
  \end{subfigure}
  \hfill
  \begin{subfigure}[t]{\textwidth}
    \centering
    \includegraphics[width=0.8\linewidth]{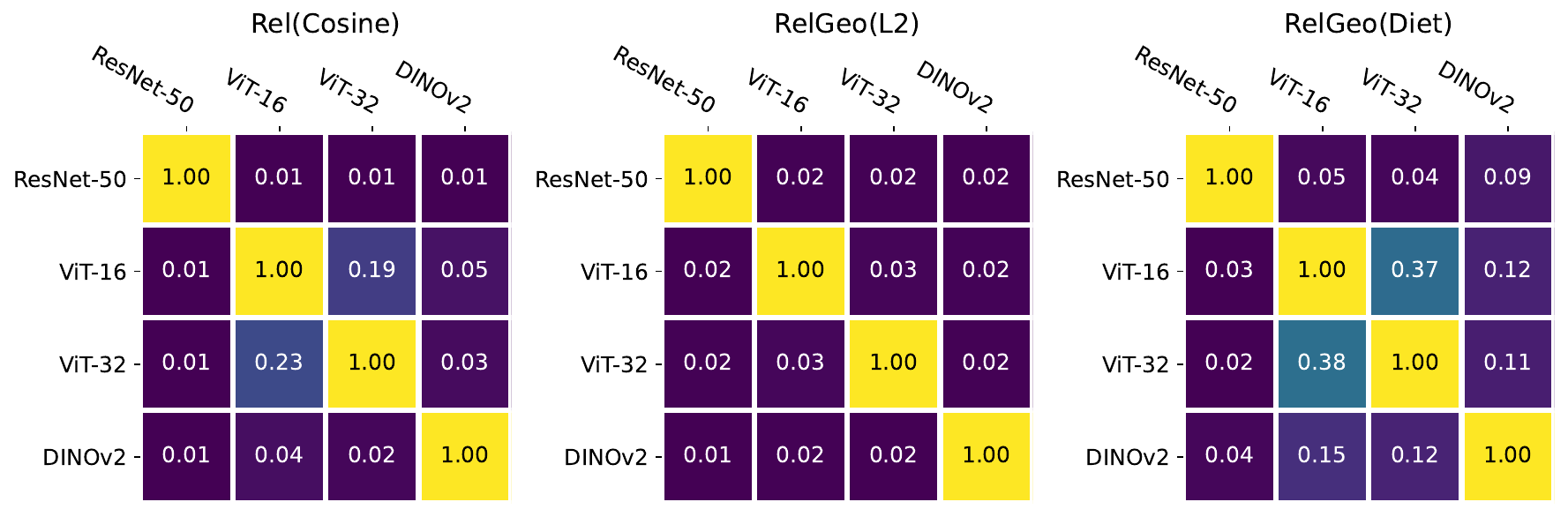}
  \end{subfigure}
  \caption{Results on CIFAR-10. From top to bottom: Accuracy, MRR Cosine Sym, MRR CDist Sym, MRR Cosine, MRR CDist}
  \label{fig:cifar10}
\end{figure}

\begin{figure}[htbp]
  \centering
  \begin{subfigure}[t]{\textwidth}
    \centering
    \includegraphics[width=0.8\linewidth]{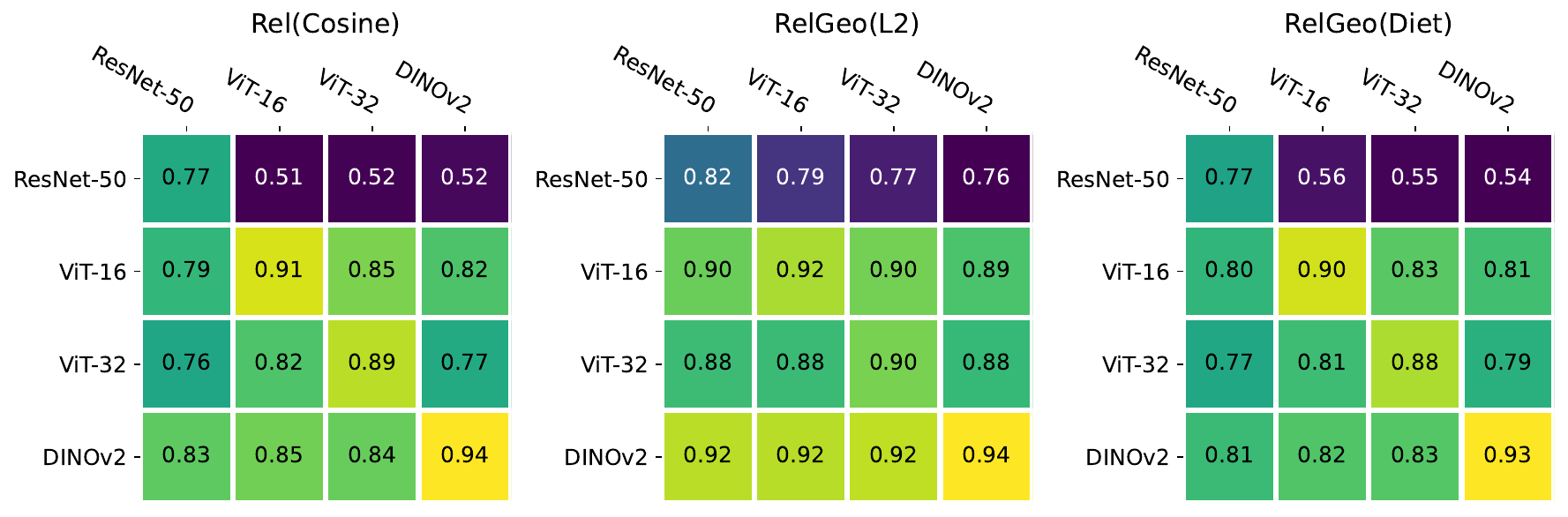}
  \end{subfigure}
  \hfill
  \begin{subfigure}[t]{\textwidth}
    \centering
    \includegraphics[width=0.8\linewidth]{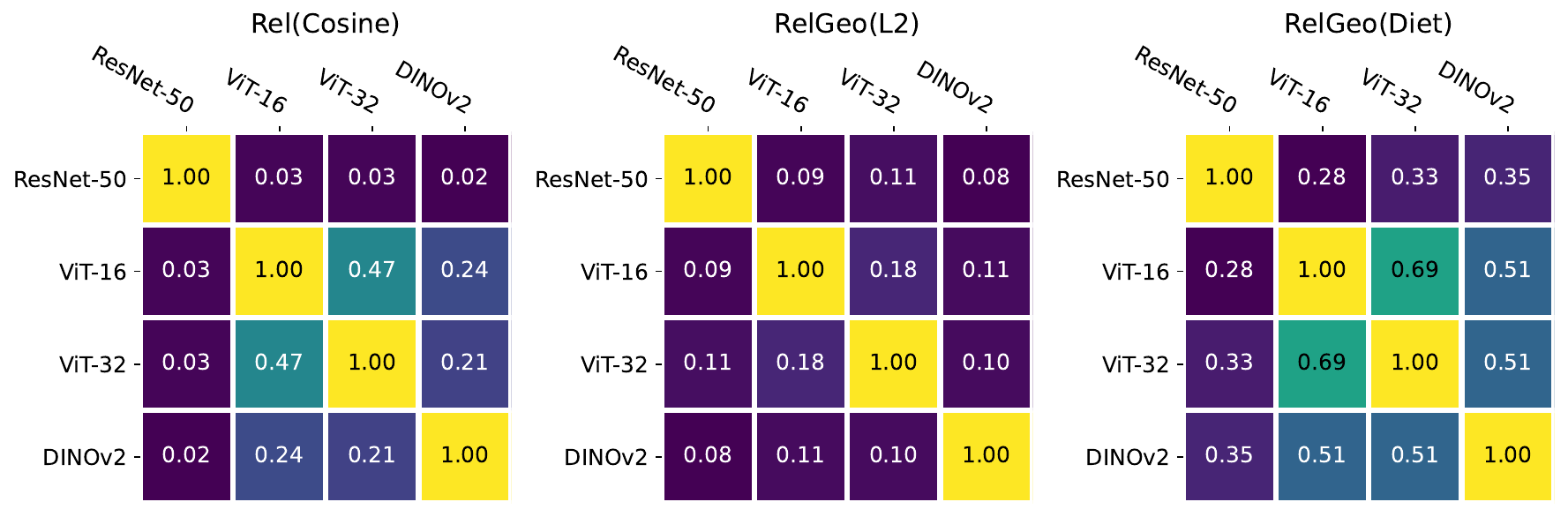}
  \end{subfigure}
  \hfill
  \begin{subfigure}[t]{\textwidth}
    \centering
    \includegraphics[width=0.8\linewidth]{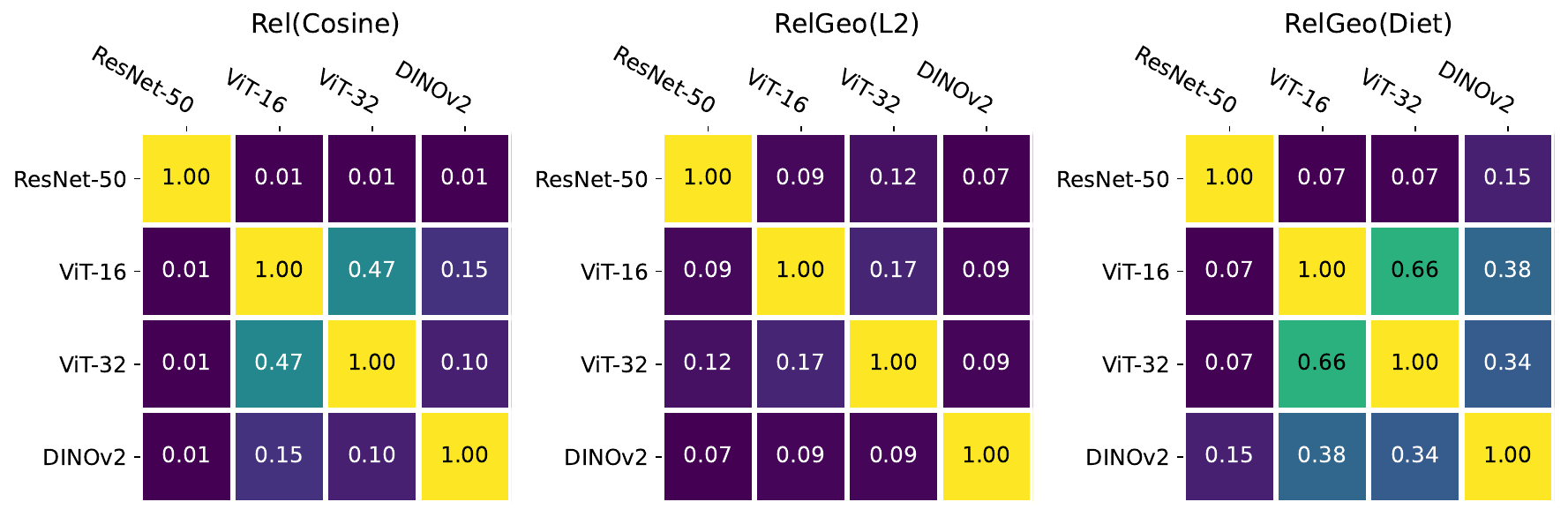}
  \end{subfigure}
  \hfill
  \begin{subfigure}[t]{\textwidth}
    \centering
    \includegraphics[width=0.8\linewidth]{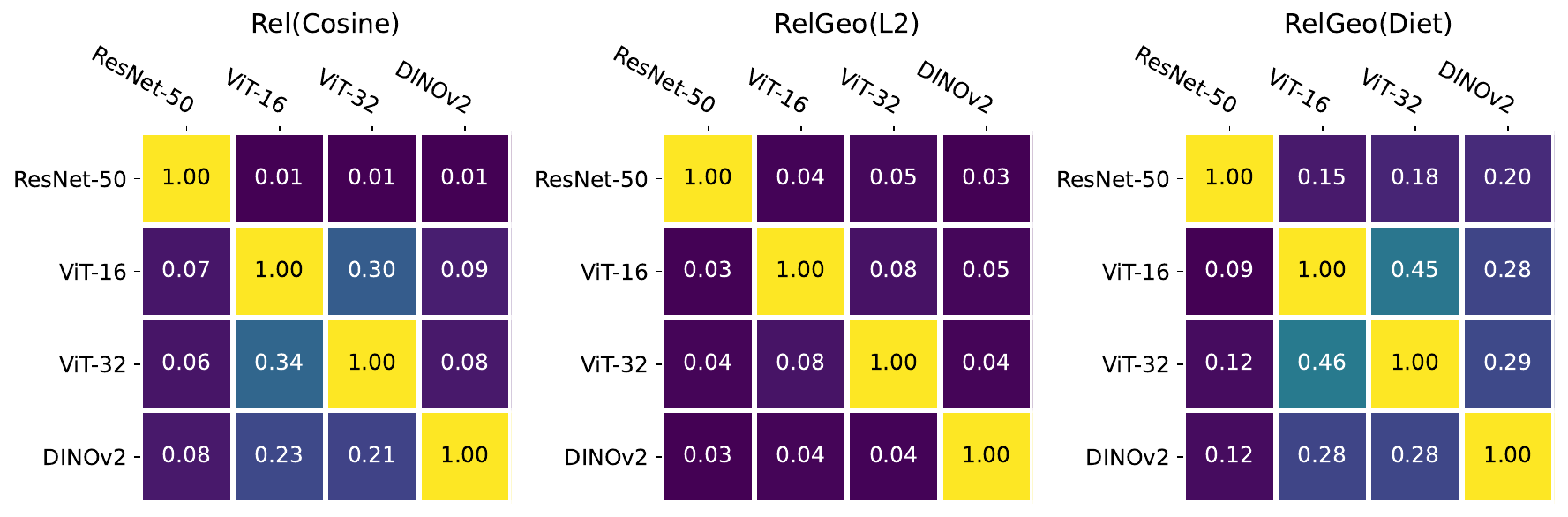}
  \end{subfigure}
  \hfill
  \begin{subfigure}[t]{\textwidth}
    \centering
    \includegraphics[width=0.8\linewidth]{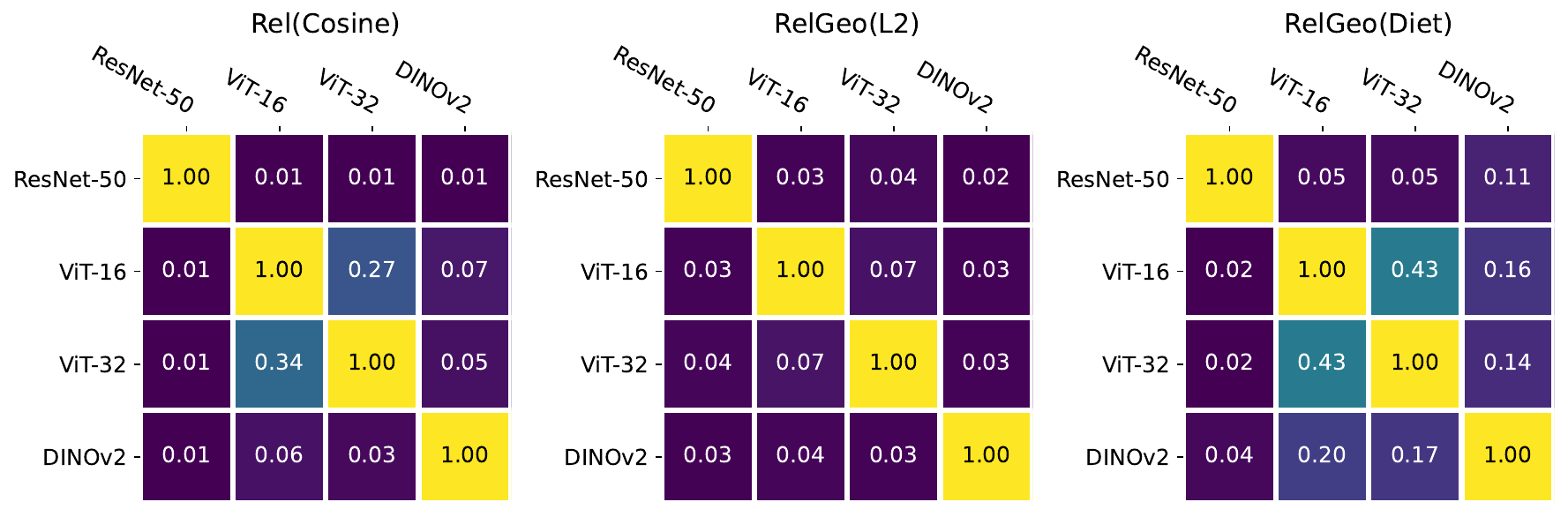}
  \end{subfigure}
  \caption{Results on CIFAR-100. From top to bottom: Accuracy, MRR Cosine Sym, MRR CDist Sym, MRR Cosine, MRR CDist}
  \label{fig:cifar100}
\end{figure}

\begin{figure}[htbp]
  \centering
  \begin{subfigure}[t]{\textwidth}
    \centering
    \includegraphics[width=0.8\linewidth]{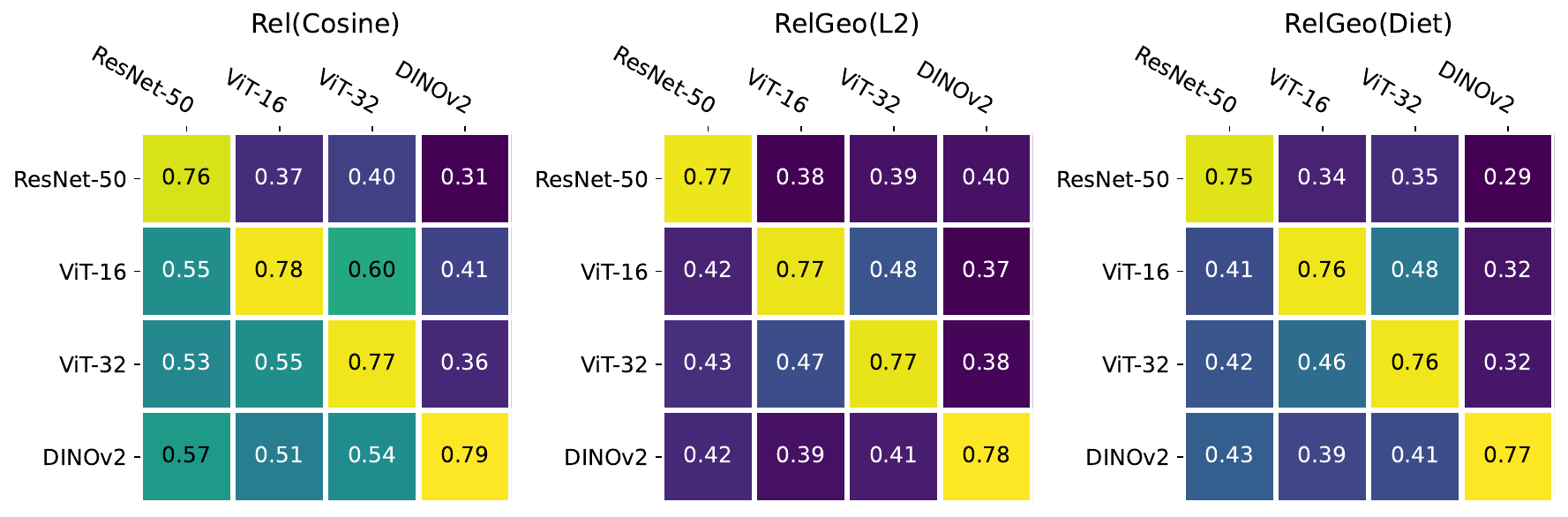}
  \end{subfigure}
  \hfill
  \begin{subfigure}[t]{\textwidth}
    \centering
    \includegraphics[width=0.8\linewidth]{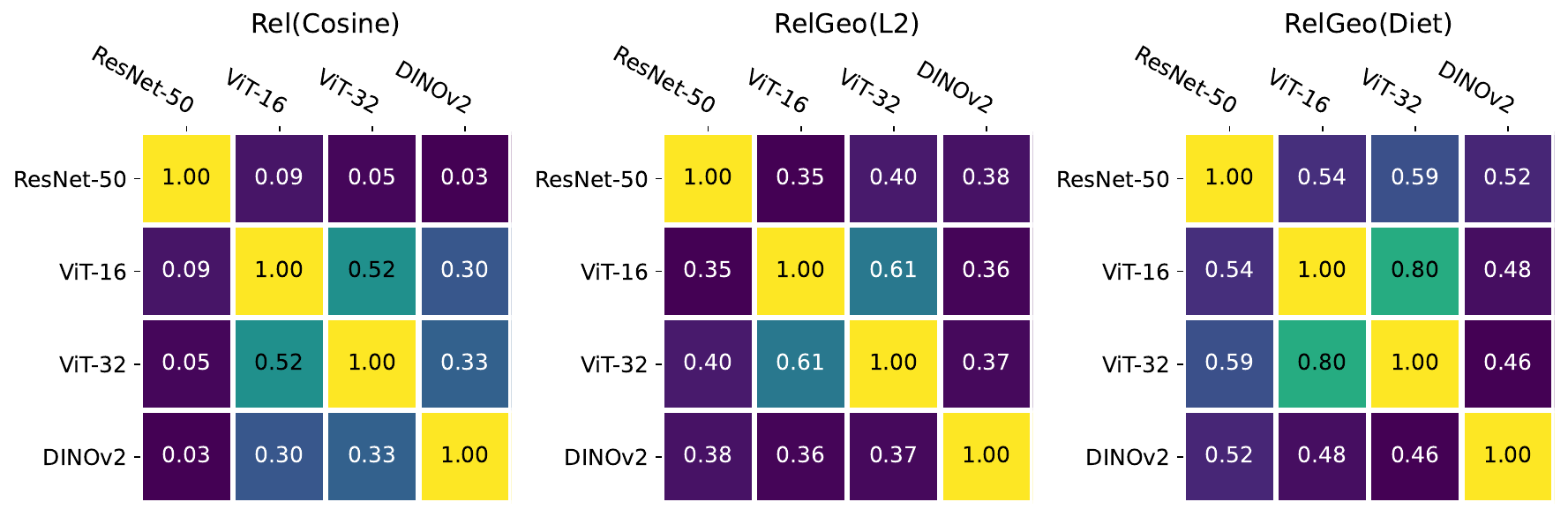}
  \end{subfigure}
  \hfill
  \begin{subfigure}[t]{\textwidth}
    \centering
    \includegraphics[width=0.8\linewidth]{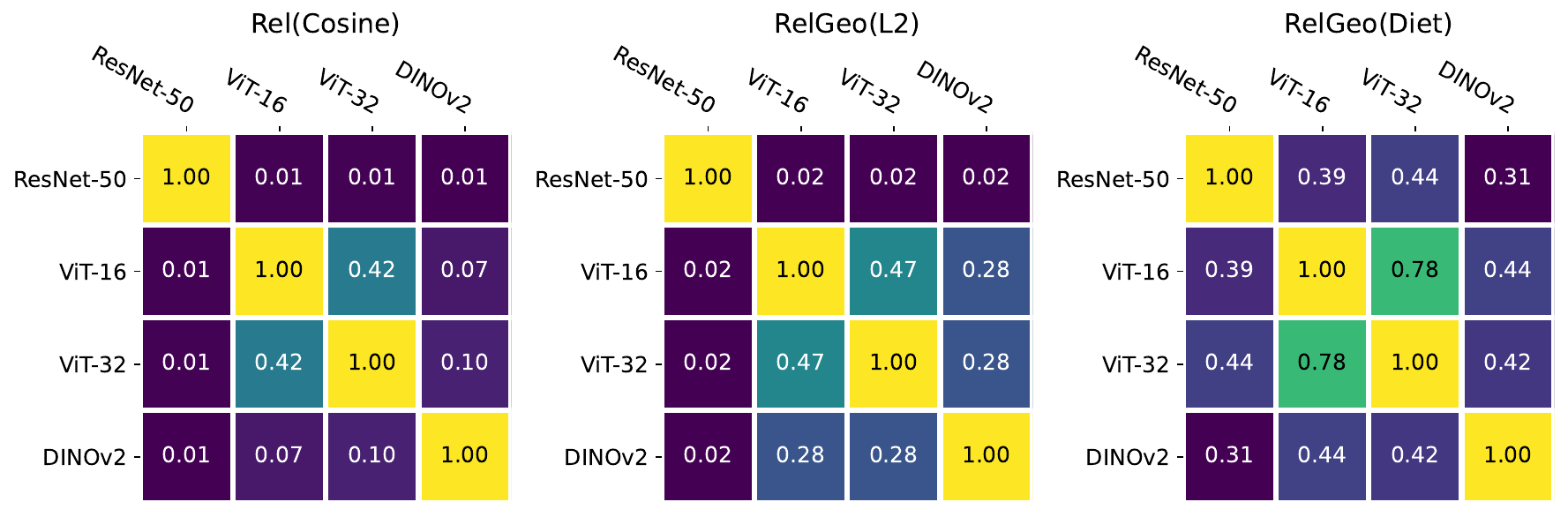}
  \end{subfigure}
  \hfill
  \begin{subfigure}[t]{\textwidth}
    \centering
    \includegraphics[width=0.8\linewidth]{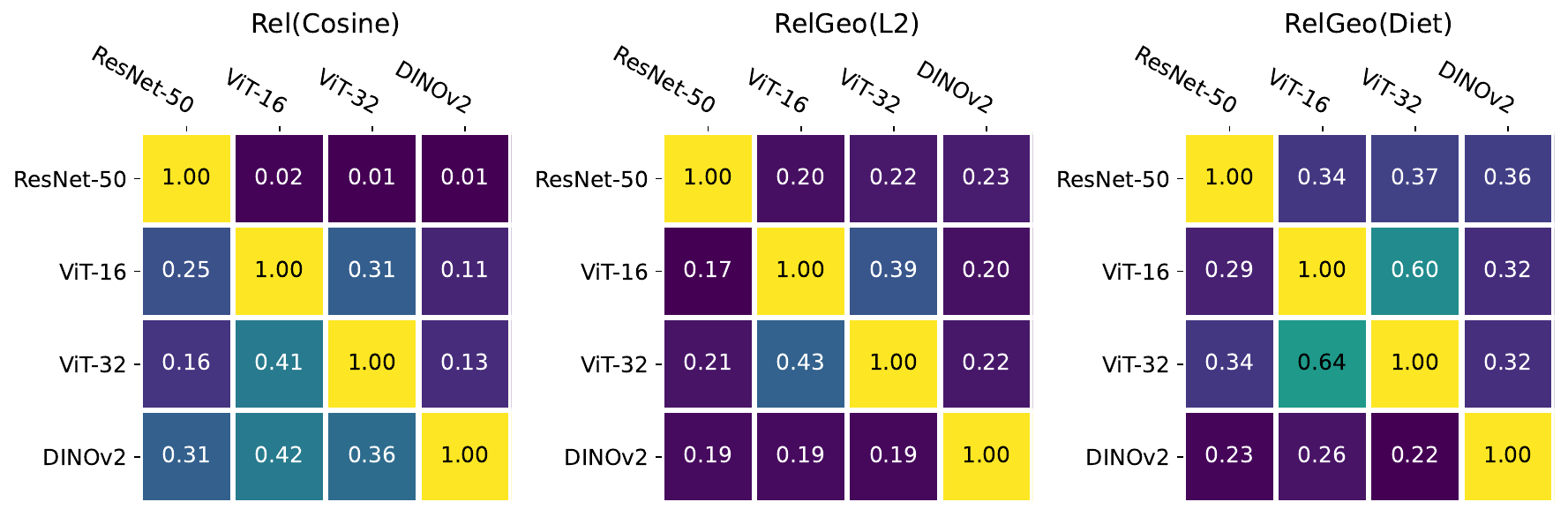}
  \end{subfigure}
  \hfill
  \begin{subfigure}[t]{\textwidth}
    \centering
    \includegraphics[width=0.8\linewidth]{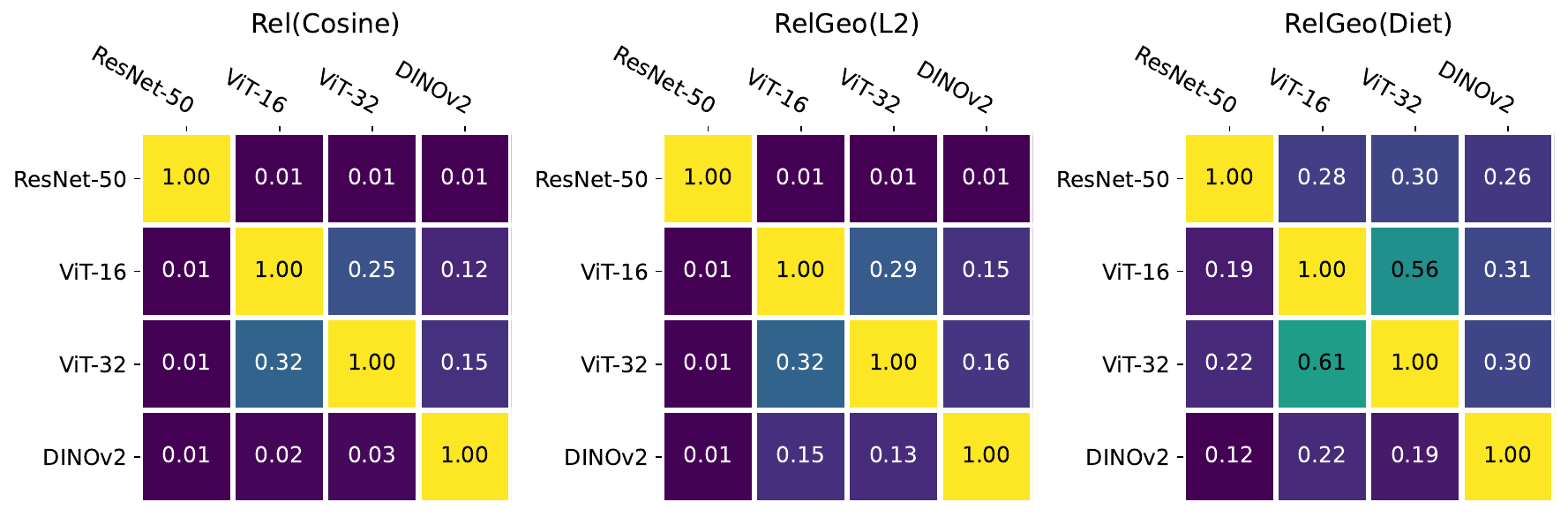}
  \end{subfigure}
  \caption{Results on ImageNet-1k. From top to bottom: Accuracy, MRR Cosine Sym, MRR CDist Sym, MRR Cosine, MRR CDist}
  \label{fig:imagenet1k}
\end{figure}

\begin{figure}[htbp]
  \centering
  \begin{subfigure}[t]{\textwidth}
    \centering
    \includegraphics[width=0.8\linewidth]{figs/cla_results/cub_accuracies.pdf}
  \end{subfigure}
  \hfill
  \begin{subfigure}[t]{\textwidth}
    \centering
    \includegraphics[width=0.8\linewidth]{figs/cla_results/cub_mrr_cosine_sym.pdf}
  \end{subfigure}
  \hfill
  \begin{subfigure}[t]{\textwidth}
    \centering
    \includegraphics[width=0.8\linewidth]{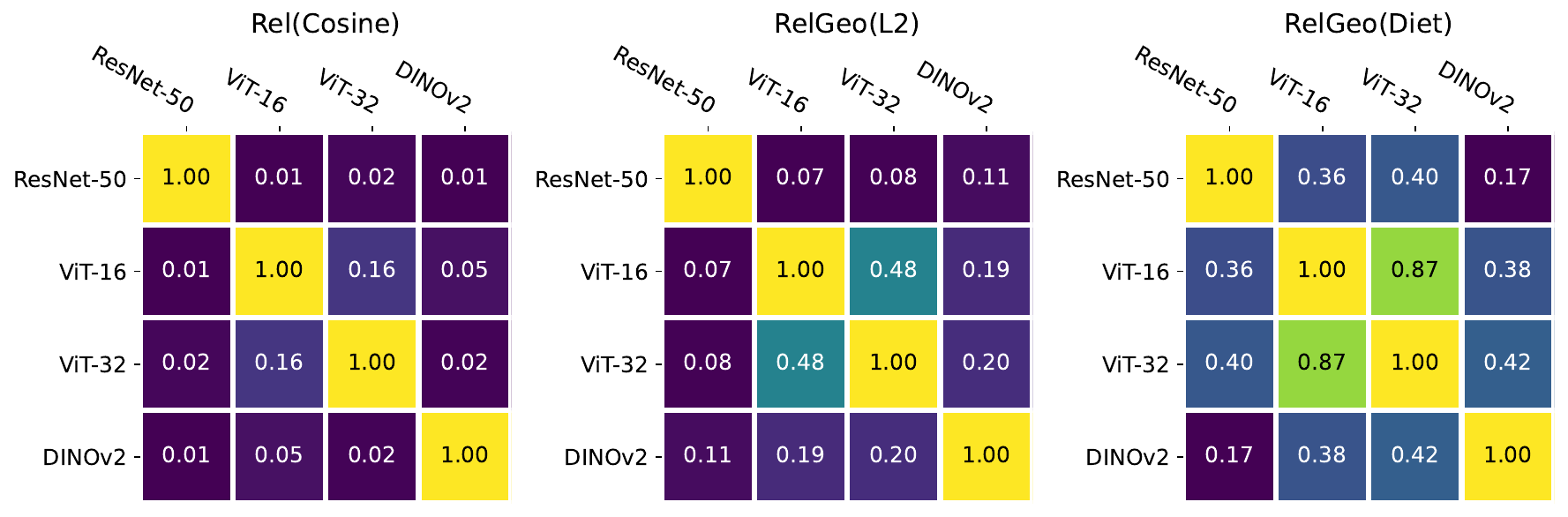}
  \end{subfigure}
  \hfill
  \begin{subfigure}[t]{\textwidth}
    \centering
    \includegraphics[width=0.8\linewidth]{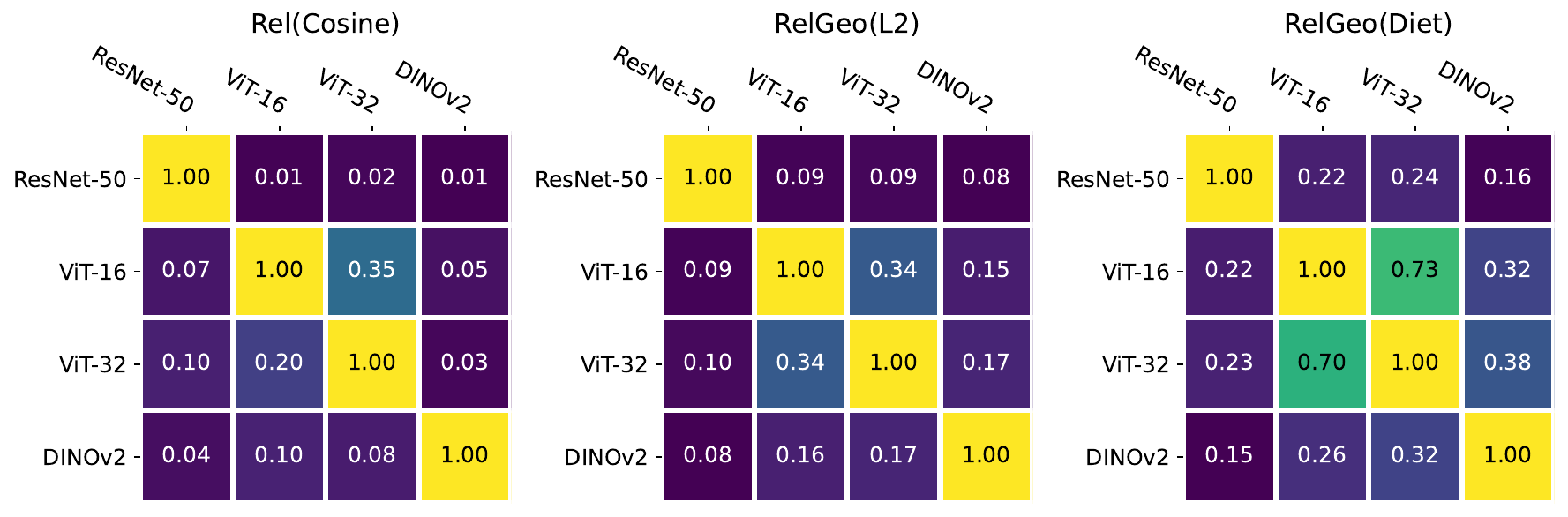}
  \end{subfigure}
  \hfill
  \begin{subfigure}[t]{\textwidth}
    \centering
    \includegraphics[width=0.8\linewidth]{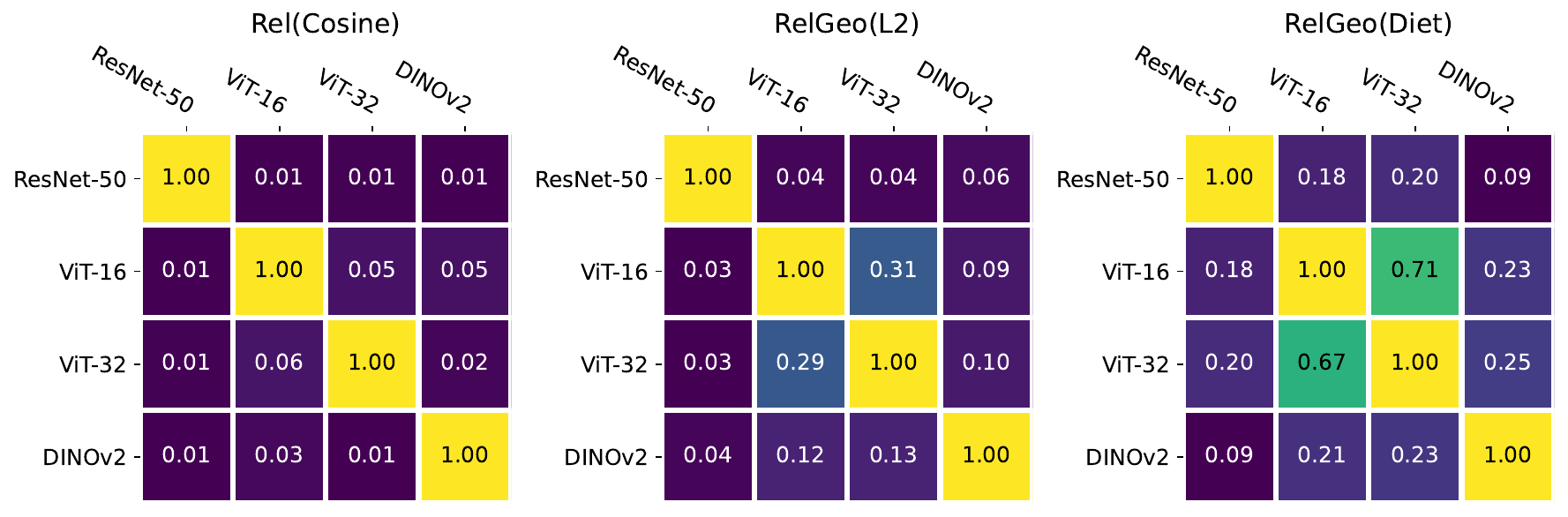}
  \end{subfigure}
  \caption{Results on CUB. From top to bottom: Accuracy, MRR Cosine Sym, MRR CDist Sym, MRR Cosine, MRR CDist}
  \label{fig:cub}
\end{figure}

\begin{figure}[htbp]
  \centering
  \begin{subfigure}[t]{\textwidth}
    \centering
    \includegraphics[width=0.8\linewidth]{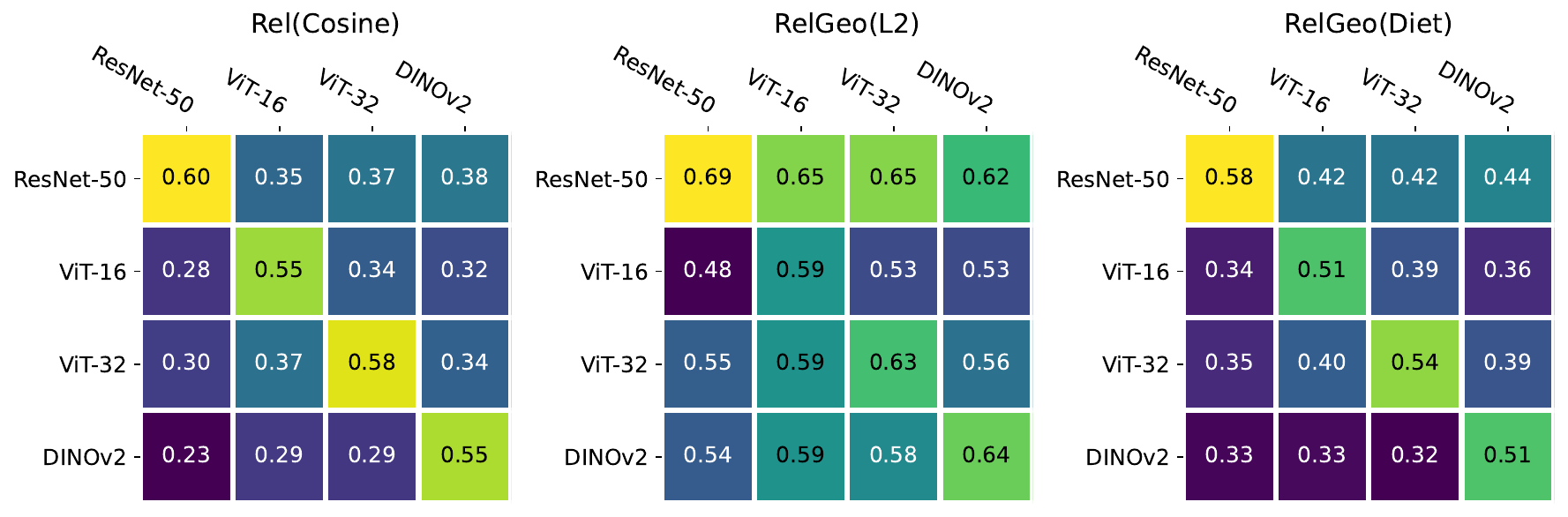}
  \end{subfigure}
  \hfill
  \begin{subfigure}[t]{\textwidth}
    \centering
    \includegraphics[width=0.8\linewidth]{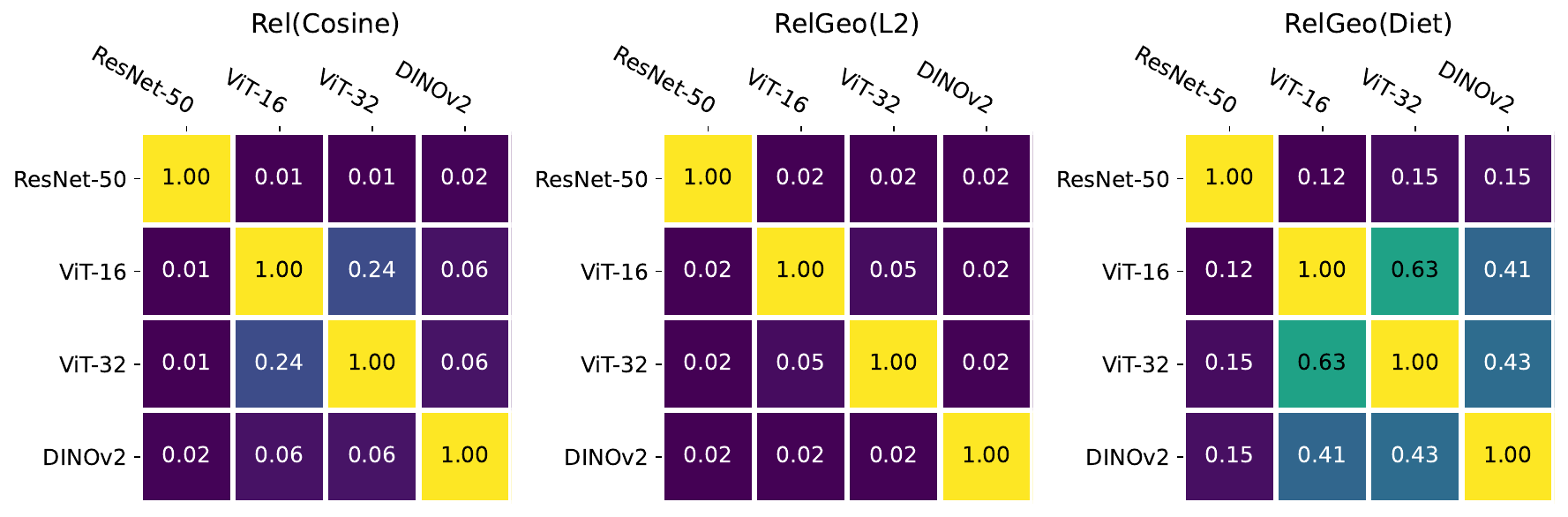}
  \end{subfigure}
  \hfill
  \begin{subfigure}[t]{\textwidth}
    \centering
    \includegraphics[width=0.8\linewidth]{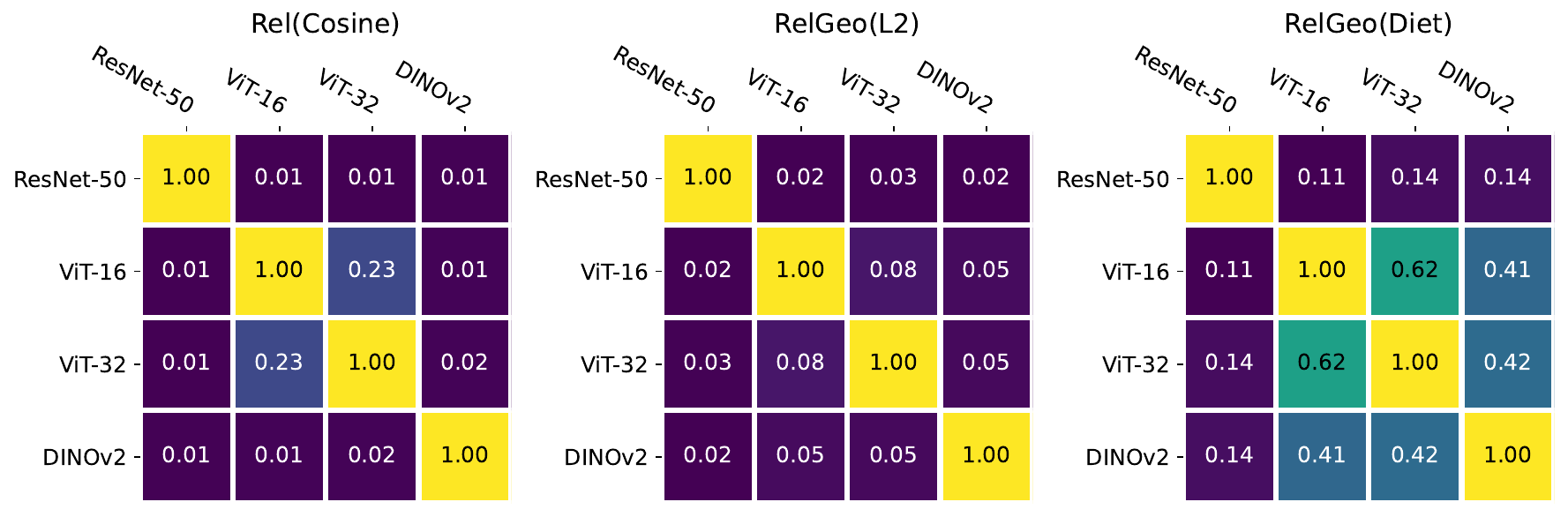}
  \end{subfigure}
  \hfill
  \begin{subfigure}[t]{\textwidth}
    \centering
    \includegraphics[width=0.8\linewidth]{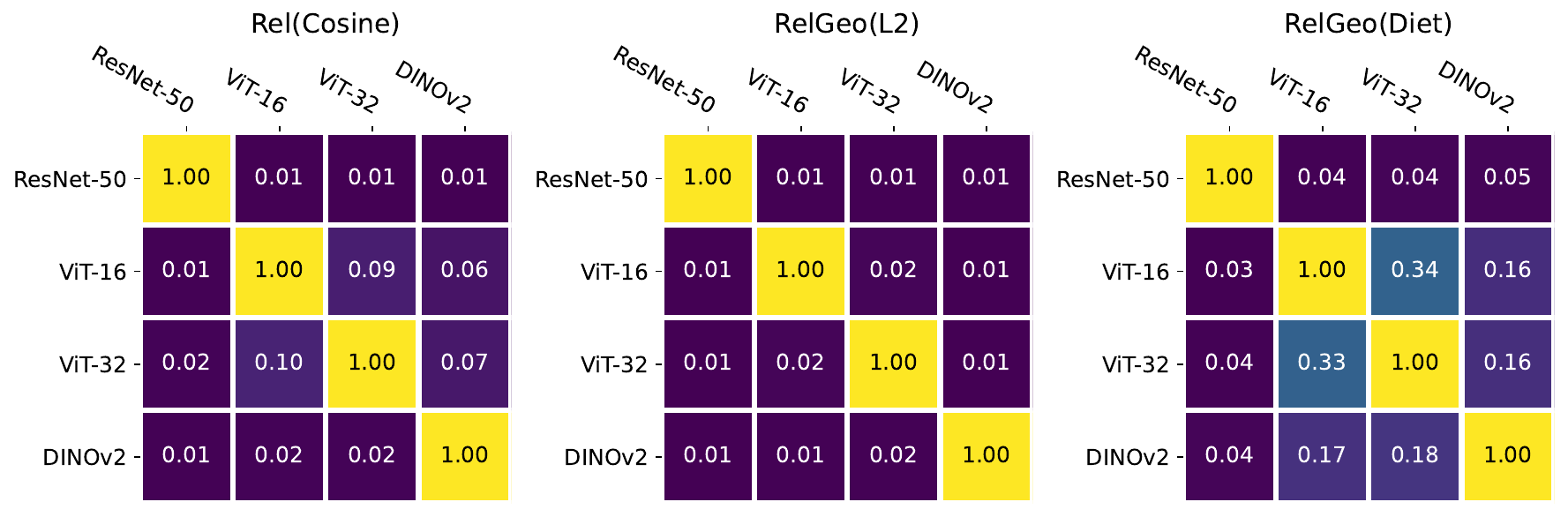}
  \end{subfigure}
  \hfill
  \begin{subfigure}[t]{\textwidth}
    \centering
    \includegraphics[width=0.8\linewidth]{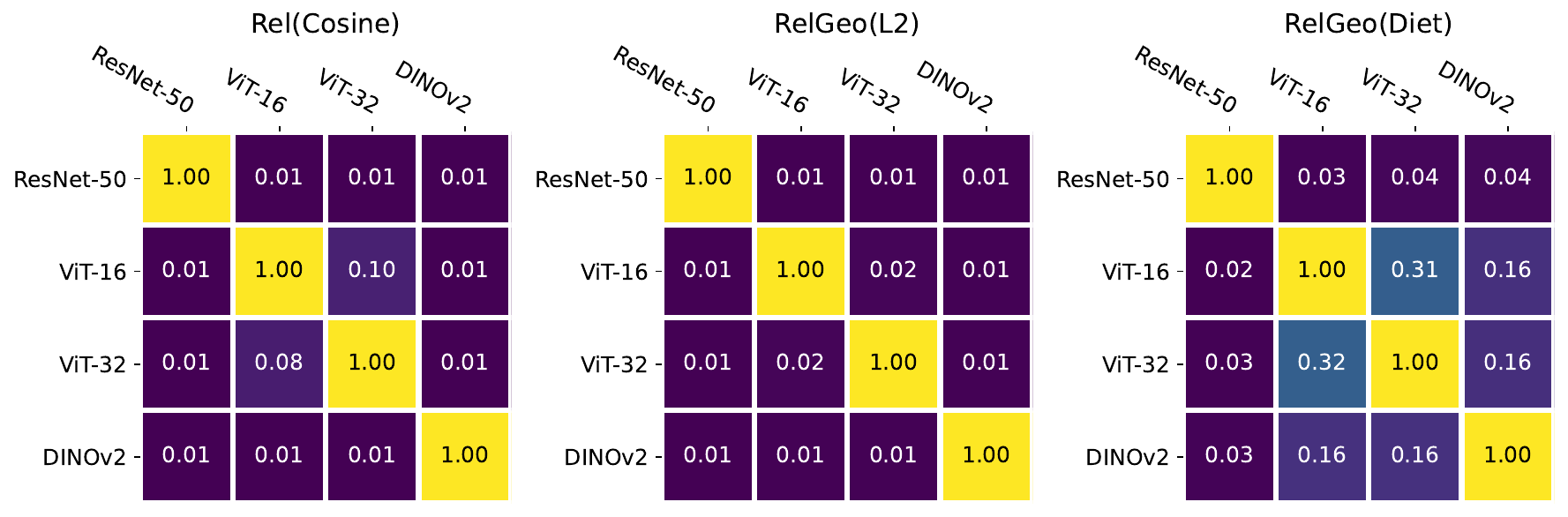}
  \end{subfigure}
  \caption{Results on SVHN. From top to bottom: Accuracy, MRR Cosine Sym, MRR CDist Sym, MRR Cosine, MRR CDist}
  \label{fig:svhn}
\end{figure}

We provide the heatmaps on the different datasets in Figure~\ref{fig:cifar10}, Figure~\ref{fig:cifar100}, Figure~\ref{fig:imagenet1k}, Figure~\ref{fig:cub} and Figure~\ref{fig:svhn}.

\subsubsection{Other evaluation metrics}

We provide the results of other evaluation metrics in Table~\ref{tbl:supp_mrr_cdist_sym}, Table~\ref{tbl:supp_mrr_cosine} and Table~\ref{tbl:supp_mrr_cdist}.

\begin{table*}
    \centering
    \caption{Aggregated results of MRR CDist Sym.}
    \resizebox{\linewidth}{!}{%
      \begin{tabular}{lccccc}
        \rowcolor{gray!20}\toprule
        \textbf{Method} & \textbf{CIFAR-10} & \textbf{CIFAR-100} & \textbf{ImageNet-1k} & \textbf{CUB} & \textbf{SVHN} \\
        \midrule
        \texttt{Rel(Cosine)} \citep{Moschella2023}       & $0.098 \pm 0.133$ & $0.122 \pm 0.164$ & $0.103 \pm 0.146$ & $0.046 \pm 0.055$ & $0.046 \pm 0.081$ \\
        \texttt{RelGeo(L2)}  & $0.046 \pm 0.013$ & $0.105 \pm 0.031$ & $0.179 \pm 0.173$ & $0.187 \pm 0.141$ & $0.04 \pm 0.021$ \\
        \texttt{RelGeo(Diet)}      & $\mathbf{0.252} \pm 0.189$ & $\mathbf{0.278} \pm 0.211$ & $\mathbf{0.462} \pm 0.148$ & $\mathbf{0.433} \pm 0.212$ & $\mathbf{0.306} \pm 0.188$ \\
        \bottomrule
      \end{tabular}
    }
    \label{tbl:supp_mrr_cdist_sym}
\end{table*}

\begin{table*}
    \centering
    \caption{Aggregated results of MRR Cosine.}
    \resizebox{\linewidth}{!}{%
      \begin{tabular}{lccccc}
        \rowcolor{gray!20}\toprule
        \textbf{Method} & \textbf{CIFAR-10} & \textbf{CIFAR-100} & \textbf{ImageNet-1k} & \textbf{CUB} & \textbf{SVHN} \\
        \midrule
        \texttt{Rel(Cosine)} \citep{Moschella2023}       & $0.08 \pm 0.077$ & $0.122 \pm 0.109$ & $0.21 \pm 0.149$ & $0.089 \pm 0.094$ & $0.035 \pm 0.034$ \\
        \texttt{RelGeo(L2)}  & $0.019 \pm 0.005$ & $0.046 \pm 0.016$ & $0.236 \pm 0.08$ & $0.156 \pm 0.089$ & $0.013 \pm 0.005$ \\
        \texttt{RelGeo(Diet)}      & $\mathbf{0.189} \pm 0.108$ & $\mathbf{0.241} \pm 0.117$ & $\mathbf{0.358} \pm 0.126$ & $\mathbf{0.327} \pm 0.184$ & $\mathbf{0.131} \pm 0.107$ \\
        \bottomrule
      \end{tabular}
    }
    \label{tbl:supp_mrr_cosine}
\end{table*}

\begin{table*}
    \centering
    \caption{Aggregated results of MRR CDist.}
    \resizebox{\linewidth}{!}{%
      \begin{tabular}{lccccc}
        \rowcolor{gray!20}\toprule
        \textbf{Method} & \textbf{CIFAR-10} & \textbf{CIFAR-100} & \textbf{ImageNet-1k} & \textbf{CUB} & \textbf{SVHN} \\
        \midrule
        \texttt{Rel(Cosine)} \citep{Moschella2023}       & $0.051 \pm 0.072$ & $0.071 \pm 0.107$ & $0.078 \pm 0.105$ & $0.023 \pm 0.02$ & $0.02 \pm 0.032$ \\
        \texttt{RelGeo(L2)}  & $0.019 \pm 0.005$ & $0.04 \pm 0.015$ & $0.106 \pm 0.11$ & $0.108 \pm 0.092$ & $0.012 \pm 0.005$ \\
        \texttt{RelGeo(Diet)}      & $\mathbf{0.127} \pm 0.118$ & $\mathbf{0.151} \pm 0.138$ & $\mathbf{0.298} \pm 0.141$ & $\mathbf{0.269} \pm 0.195$ & $\mathbf{0.123} \pm 0.103$ \\
        \bottomrule
      \end{tabular}
    }
    \label{tbl:supp_mrr_cdist}
\end{table*}

\subsubsection{Alternative aggregation}

Here we consider an alternative way to aggregate the results, i.e. grouping by the models. The results are reported in Table~\ref{tbl:alt_accuracies}, Table~\ref{tbl:alt_mrr_cosine_sym}, Table~\ref{tbl:alt_mrr_cdist_sym}, Table~\ref{tbl:alt_mrr_cosine} and Table~\ref{tbl:alt_mrr_cdist}. In general, the observation remains: RelGeo(L2) yields good accuracies and RelGeo(Diet) yields good MRRs.

\begin{table*}
    \centering
    \caption{Alternatively aggregated results of Accuracy.}
    \resizebox{\linewidth}{!}{%
        \begin{tabular}{lcccc}
          \rowcolor{gray!20}\toprule
          \textbf{Method} & \textbf{ResNet-50} & \textbf{ViT-16} & \textbf{ViT-32} & \textbf{DINOv2} \\
          \midrule
          \texttt{Rel(Cosine)} \citep{Moschella2023}& $0.507 \pm 0.2$ & $0.669 \pm 0.229$ & $0.664 \pm 0.218$ & $0.678 \pm 0.24$ \\
          \texttt{RelGeo(L2)} & $\mathbf{0.646} \pm 0.209$ & $\mathbf{0.709} \pm 0.208$ & $\mathbf{0.72} \pm 0.194$ & $\mathbf{0.737} \pm 0.209$ \\
          \texttt{RelGeo(Diet)} & $0.529 \pm 0.194$ & $0.658 \pm 0.229$ & $0.661 \pm 0.219$ & $0.668 \pm 0.237$ \\
          \bottomrule
        \end{tabular}
    }
    \label{tbl:alt_accuracies}
\end{table*}

\begin{table*}
    \centering
    \caption{Alternatively aggregated results of MRR Cosine Sym.}
    \resizebox{\linewidth}{!}{%
        \begin{tabular}{lcccc}
          \rowcolor{gray!20}\toprule
          \textbf{Method} & \textbf{ResNet-50} & \textbf{ViT-16} & \textbf{ViT-32} & \textbf{DINOv2} \\
          \midrule
          \texttt{Rel(Cosine)} \citep{Moschella2023}& $0.032 \pm 0.023$ & $0.212 \pm 0.173$ & $0.208 \pm 0.175$ & $0.124 \pm 0.104$ \\
          \texttt{RelGeo(L2)} & $0.143 \pm 0.132$ & $0.197 \pm 0.186$ & $0.205 \pm 0.189$ & $0.154 \pm 0.137$ \\
          \texttt{RelGeo(Diet)} & $\mathbf{0.336} \pm 0.143$ & $\mathbf{0.506} \pm 0.2$ & $\mathbf{0.526} \pm 0.187$ & $\mathbf{0.42} \pm 0.106$ \\
          \bottomrule
        \end{tabular}
    }
    \label{tbl:alt_mrr_cosine_sym}
\end{table*}

\begin{table*}
    \centering
    \caption{Alternatively aggregated results of MRR CDist Sym.}
    \resizebox{\linewidth}{!}{%
        \begin{tabular}{lcccc}
          \rowcolor{gray!20}\toprule
          \textbf{Method} & \textbf{ResNet-50} & \textbf{ViT-16} & \textbf{ViT-32} & \textbf{DINOv2} \\
          \midrule
          \texttt{Rel(Cosine)} \citep{Moschella2023}& $0.009 \pm 0.005$ & $0.141 \pm 0.156$ & $0.134 \pm 0.158$ & $0.049 \pm 0.047$ \\
          \texttt{RelGeo(L2)} & $0.052 \pm 0.033$ & $0.144 \pm 0.146$ & $0.147 \pm 0.146$ & $0.103 \pm 0.085$ \\
          \texttt{RelGeo(Diet)} & $\mathbf{0.204} \pm 0.13$ & $\mathbf{0.432} \pm 0.235$ & $\mathbf{0.437} \pm 0.232$ & $\mathbf{0.313} \pm 0.108$ \\
          \bottomrule
        \end{tabular}
    }
    \label{tbl:alt_mrr_cdist_sym}
\end{table*}

\begin{table*}
    \centering
    \caption{Alternatively aggregated results of MRR Cosine.}
    \resizebox{\linewidth}{!}{%
        \begin{tabular}{lcccc}
          \rowcolor{gray!20}\toprule
          \textbf{Method} & \textbf{ResNet-50} & \textbf{ViT-16} & \textbf{ViT-32} & \textbf{DINOv2} \\
          \midrule
          \texttt{Rel(Cosine)} \citep{Moschella2023}& $0.011 \pm 0.005$ & $0.138 \pm 0.11$ & $0.133 \pm 0.112$ & $0.147 \pm 0.128$ \\
          \texttt{RelGeo(L2)} & $0.074 \pm 0.077$ & $0.107 \pm 0.118$ & $0.116 \pm 0.126$ & $0.079 \pm 0.074$ \\
          \texttt{RelGeo(Diet)} & $\mathbf{0.182} \pm 0.107$ & $\mathbf{0.299} \pm 0.184$ & $\mathbf{0.316} \pm 0.182$ & $\mathbf{0.201} \pm 0.076$ \\
          \bottomrule
        \end{tabular}
    }
    \label{tbl:alt_mrr_cosine}
\end{table*}

\begin{table*}
    \centering
    \caption{Alternatively aggregated results of MRR CDist.}
    \resizebox{\linewidth}{!}{%
        \begin{tabular}{lcccc}
          \rowcolor{gray!20}\toprule
          \textbf{Method} & \textbf{ResNet-50} & \textbf{ViT-16} & \textbf{ViT-32} & \textbf{DINOv2} \\
          \midrule
          \texttt{Rel(Cosine)} \citep{Moschella2023}& $0.007 \pm 0.001$ & $0.079 \pm 0.086$ & $0.089 \pm 0.112$ & $0.019 \pm 0.015$ \\
          \texttt{RelGeo(L2)} & $0.023 \pm 0.016$ & $0.075 \pm 0.096$ & $0.078 \pm 0.099$ & $0.051 \pm 0.05$ \\
          \texttt{RelGeo(Diet)} & $\mathbf{0.121} \pm 0.094$ & $\mathbf{0.253} \pm 0.193$ & $\mathbf{0.258} \pm 0.194$ & $\mathbf{0.143} \pm 0.065$ \\
          \bottomrule
        \end{tabular}
    }
    \label{tbl:alt_mrr_cdist}
\end{table*}

\subsubsection{Number of anchors}

\begin{figure*}
    \centering
    \begin{subfigure}[t]{\textwidth}
    \centering
    \includegraphics[width=0.8\linewidth]{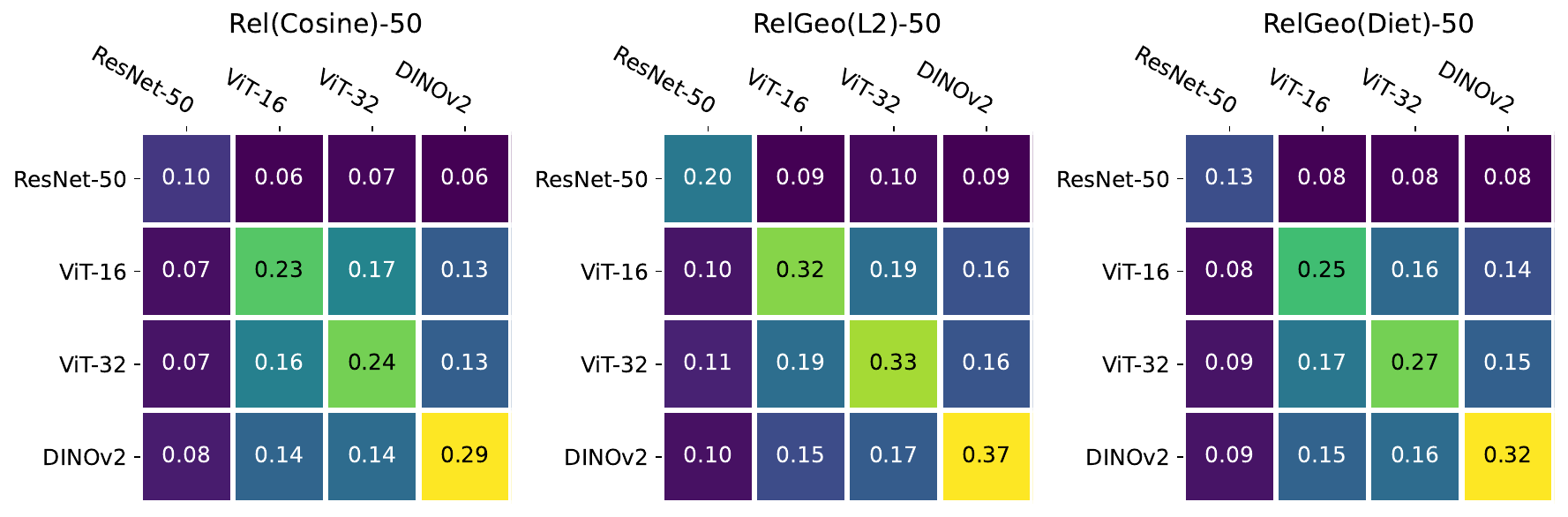}
    \end{subfigure}
    \hfill
    \begin{subfigure}[t]{\textwidth}
    \centering
    \includegraphics[width=0.8\linewidth]{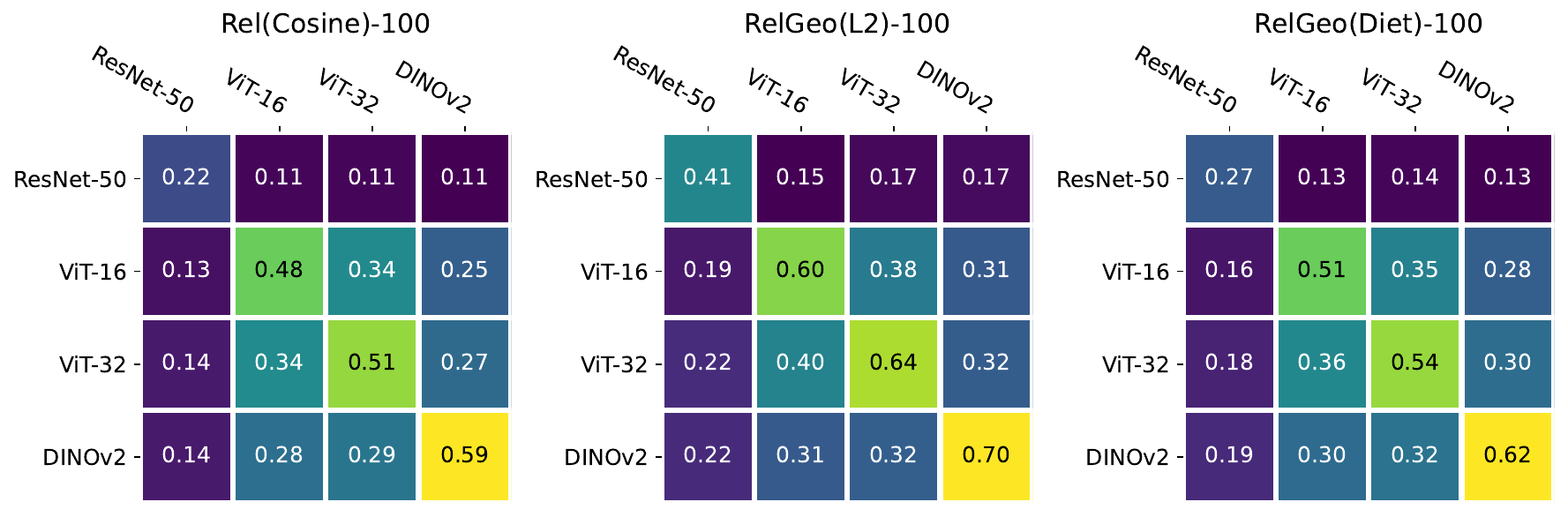}
    \end{subfigure}
    \hfill
    \begin{subfigure}[t]{\textwidth}
    \centering
    \includegraphics[width=0.8\linewidth]{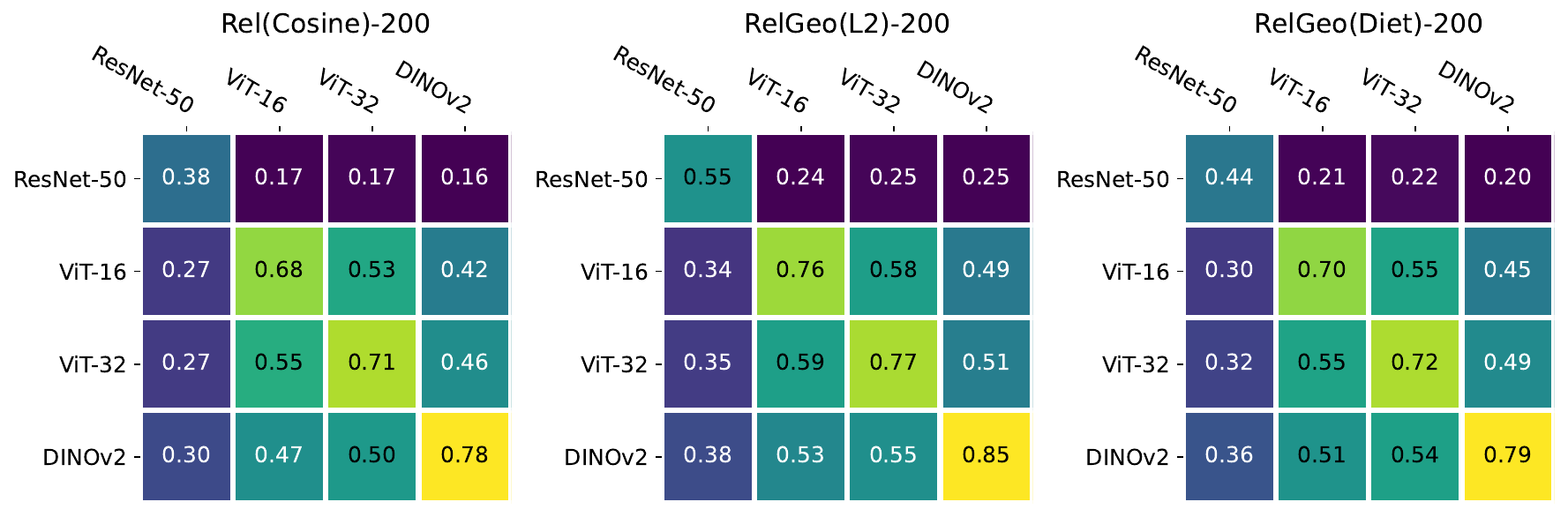}
    \end{subfigure}
    \hfill
    \begin{subfigure}[t]{\textwidth}
    \centering
    \includegraphics[width=0.8\linewidth]{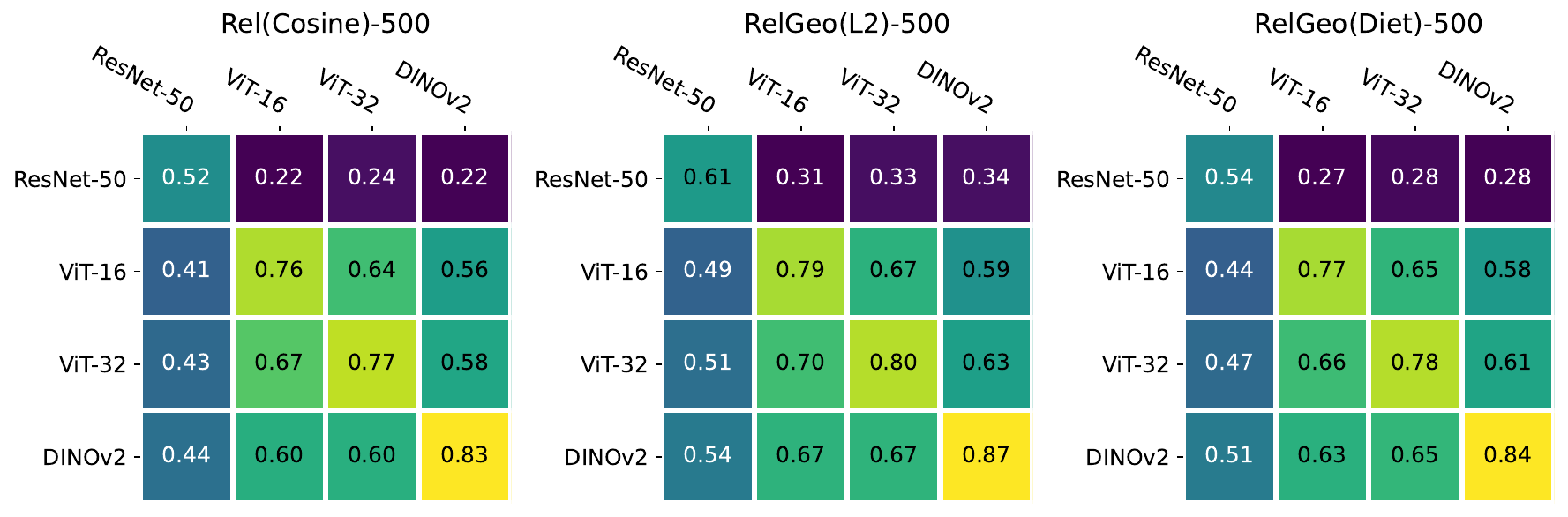}
    \end{subfigure}
    \hfill
    \begin{subfigure}[t]{\textwidth}
    \centering
    \includegraphics[width=0.8\linewidth]{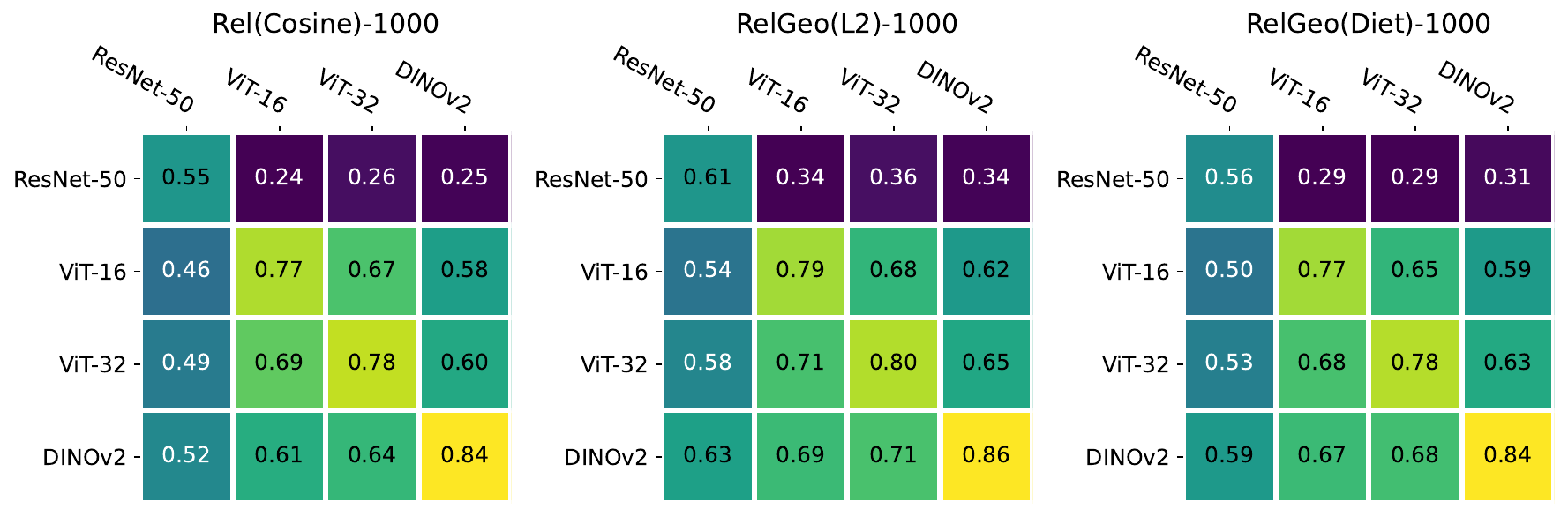}
    \end{subfigure}
  \caption{Accuracies on CUB with varying number of anchors. From top to bottom: 50, 100, 200, 500, 1000 anchors.}
  \label{fig:num_anchors_accuracies}
\end{figure*}

\begin{figure*}
    \centering
    \begin{subfigure}[t]{\textwidth}
    \centering
    \includegraphics[width=0.8\linewidth]{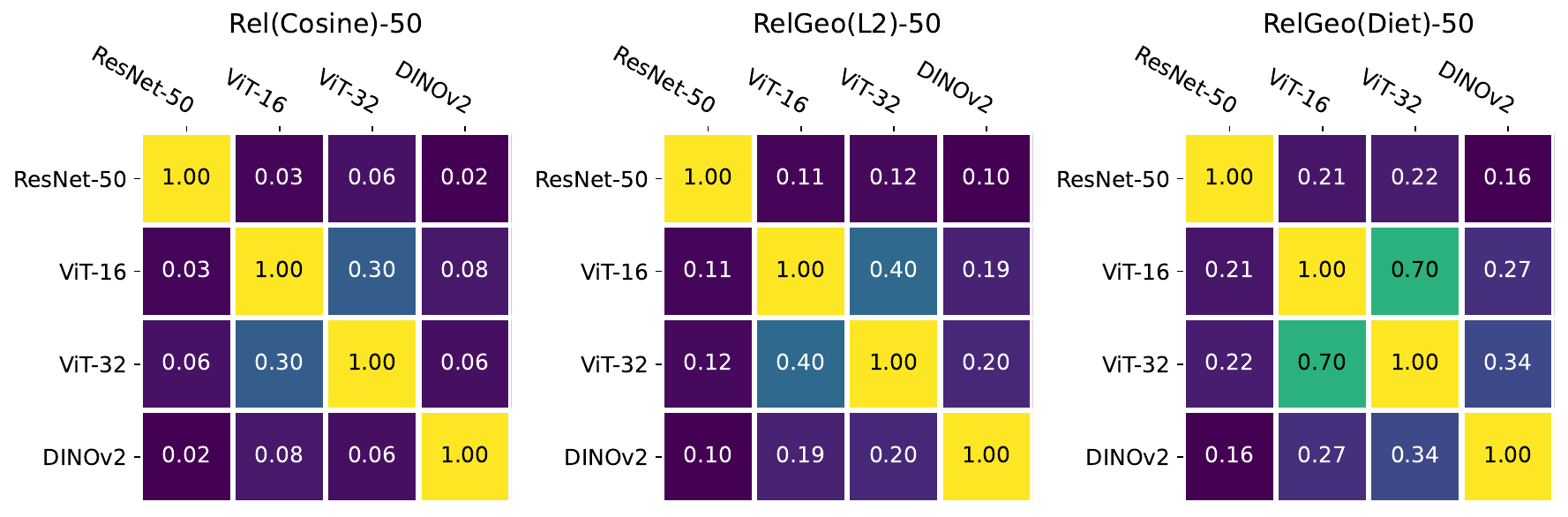}
    \end{subfigure}
    \hfill
    \begin{subfigure}[t]{\textwidth}
    \centering
    \includegraphics[width=0.8\linewidth]{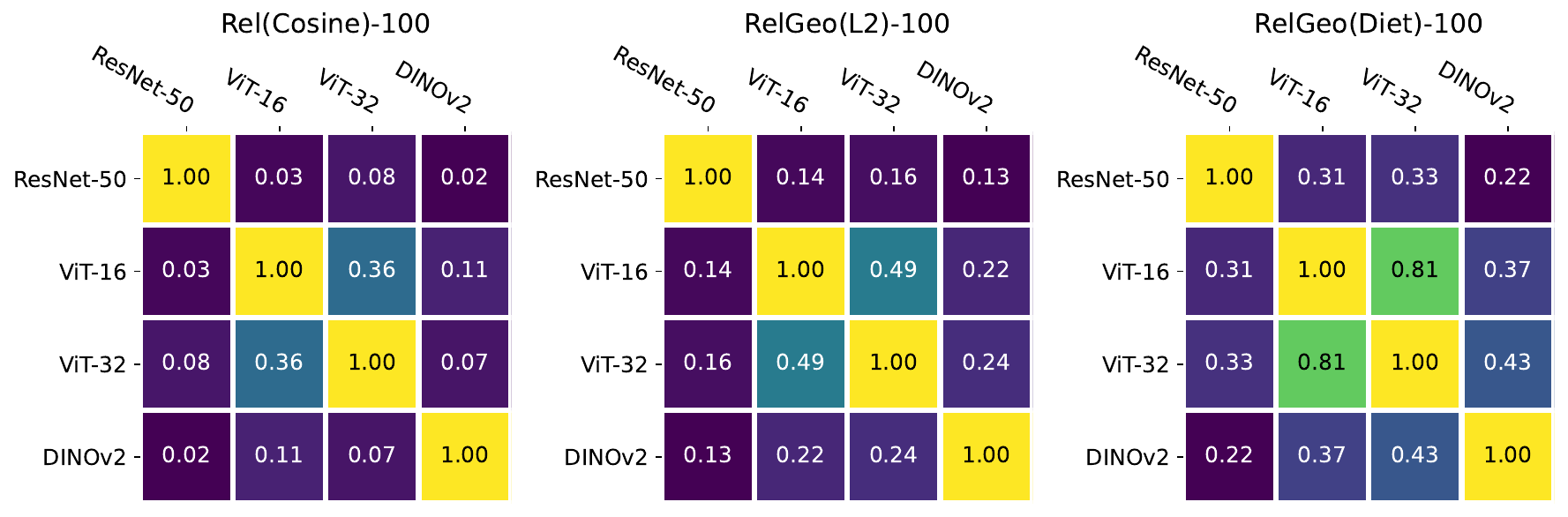}
    \end{subfigure}
    \hfill
    \begin{subfigure}[t]{\textwidth}
    \centering
    \includegraphics[width=0.8\linewidth]{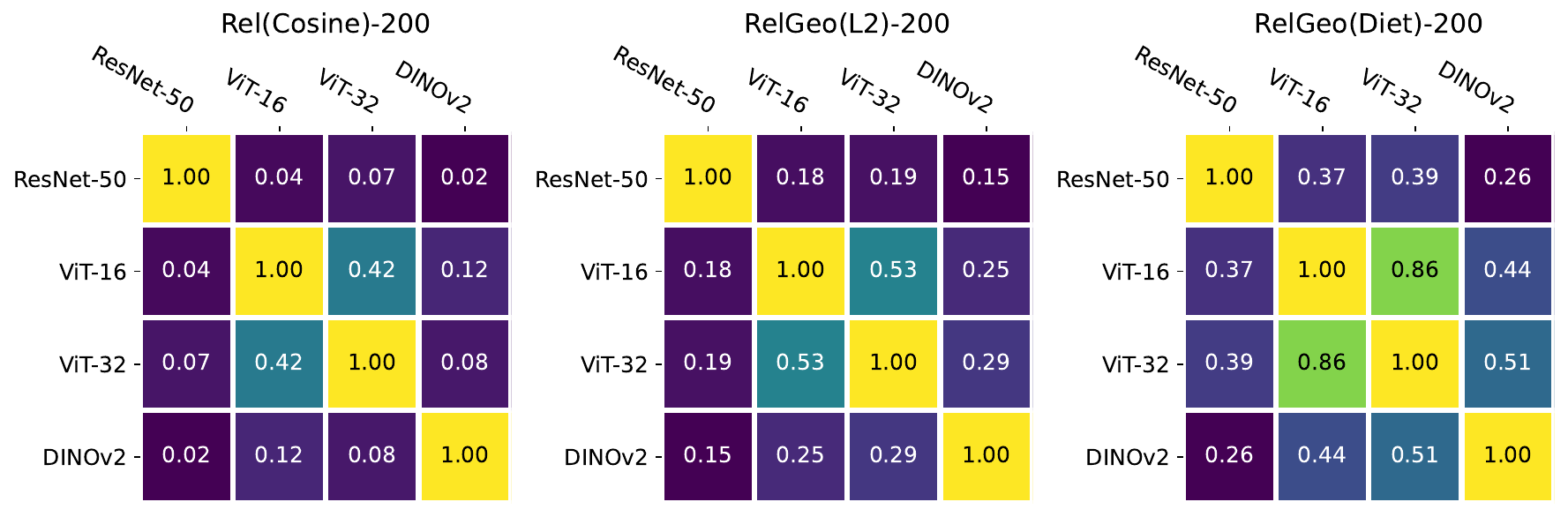}
    \end{subfigure}
    \hfill
    \begin{subfigure}[t]{\textwidth}
    \centering
    \includegraphics[width=0.8\linewidth]{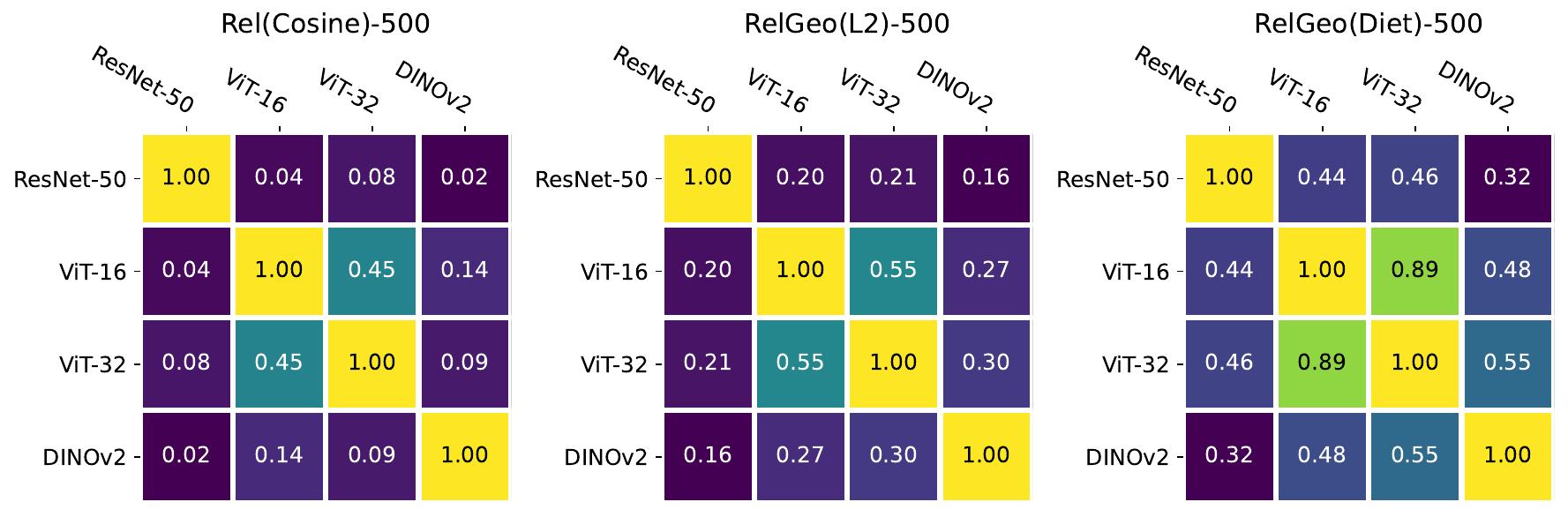}
    \end{subfigure}
    \hfill
    \begin{subfigure}[t]{\textwidth}
    \centering
    \includegraphics[width=0.8\linewidth]{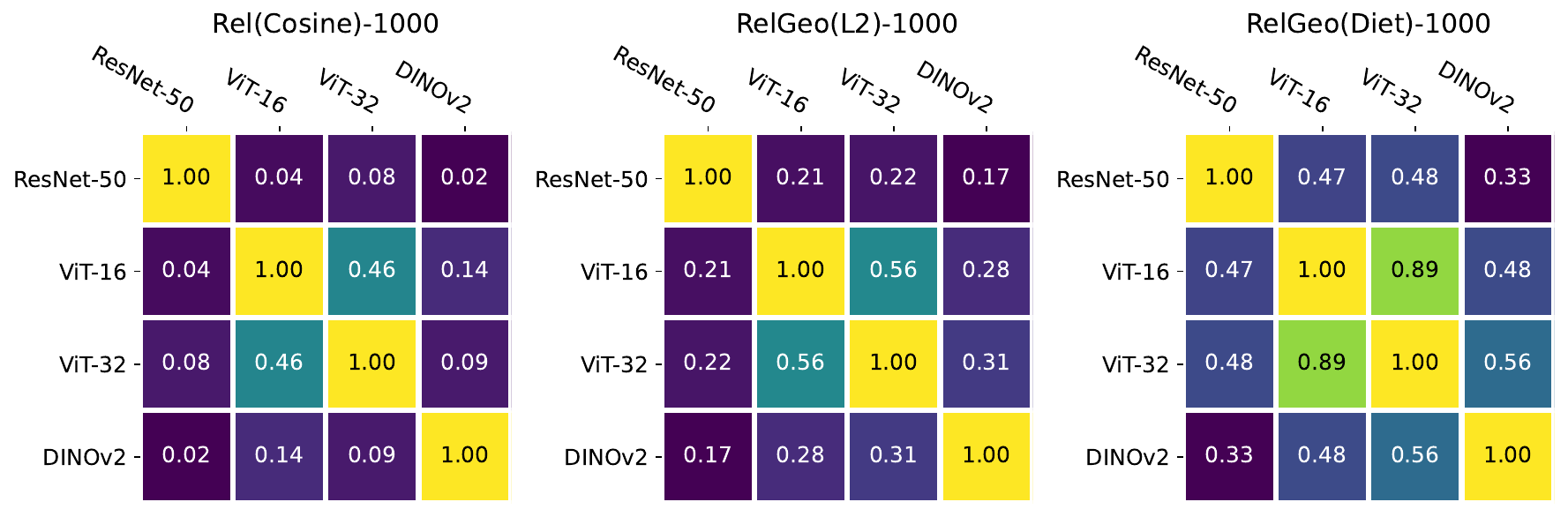}
    \end{subfigure}
  \caption{MRR Cosine Sym on CUB with varying number of anchors. From top to bottom: 50, 100, 200, 500, 1000 anchors.}
  \label{fig:num_anchors_mrr_cosine_sym}
\end{figure*}

We investigate the impact of the number of anchors. The results are shown in Figure~\ref{fig:num_anchors_accuracies} and Figure~\ref{fig:num_anchors_mrr_cosine_sym}. The general conclusion that RelGeo(L2) is good in terms of accuracies, RelGeo(Diet) is good in terms of MRRs persist with varying number of anchors.

\subsubsection{Number of Diet points}

\begin{figure*}
    \centering
    \begin{subfigure}[t]{\textwidth}
    \centering
    \includegraphics[width=0.8\linewidth]{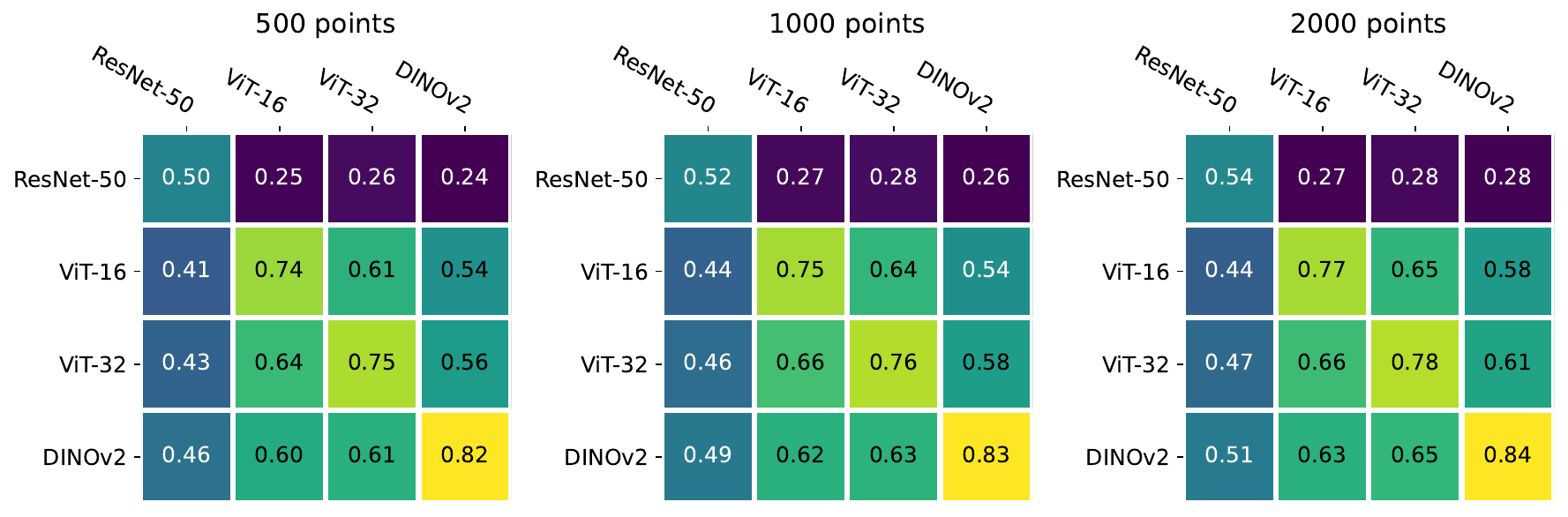}
    \end{subfigure}
    \hfill
    \begin{subfigure}[t]{\textwidth}
    \centering
    \includegraphics[width=0.8\linewidth]{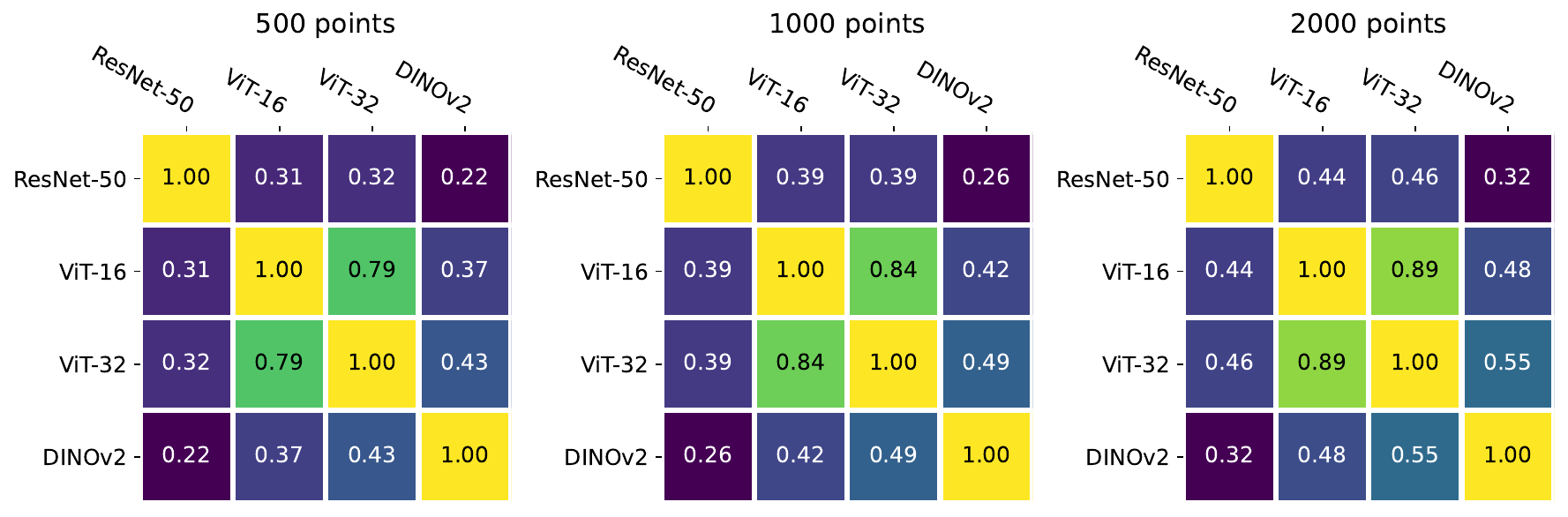}
    \end{subfigure}
  \caption{Results of RelGeo(Diet) on CUB with varying number of diet points. Top: accuracies; bottom: MRR Cosine Sym.}
  \label{fig:num_diets}
\end{figure*}

We analyze the impact of the number of Diet points. The results are shown in Figure~\ref{fig:num_diets}. The performances of RelGeo(Diet) improve as the number of diet points become larger.

\subsubsection{Number of discretization steps}

\begin{figure*}
    \centering
    \begin{subfigure}[t]{\textwidth}
    \centering
    \includegraphics[width=0.8\linewidth]{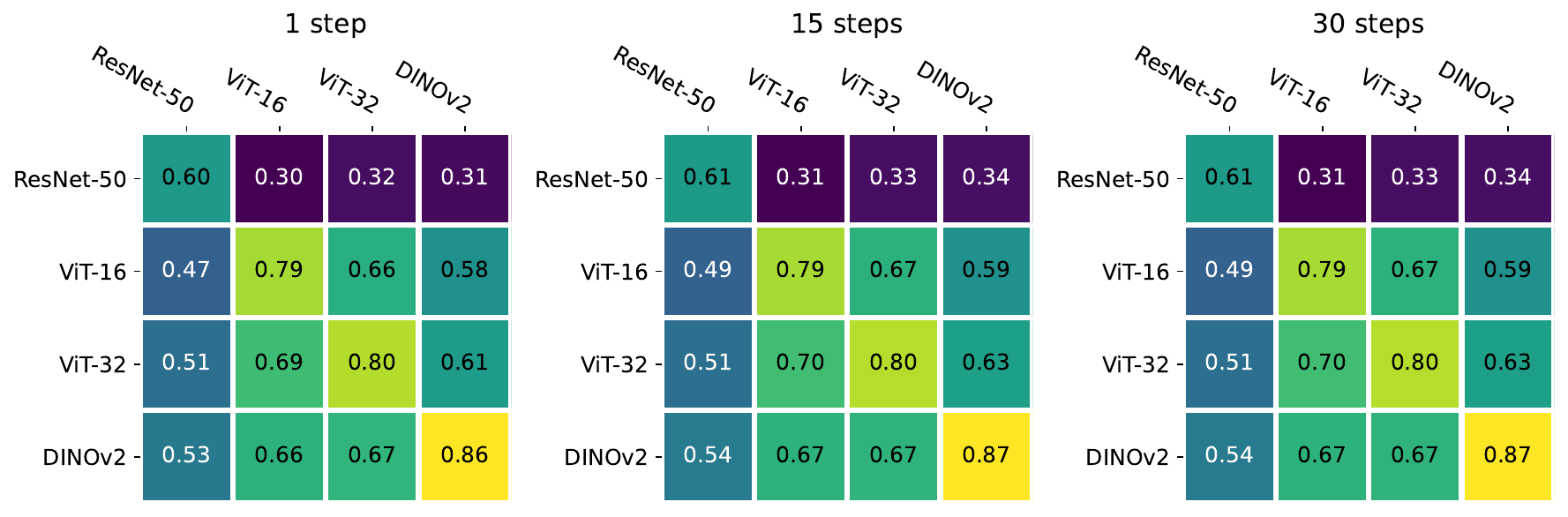}
    \end{subfigure}
    \hfill
    \begin{subfigure}[t]{\textwidth}
    \centering
    \includegraphics[width=0.8\linewidth]{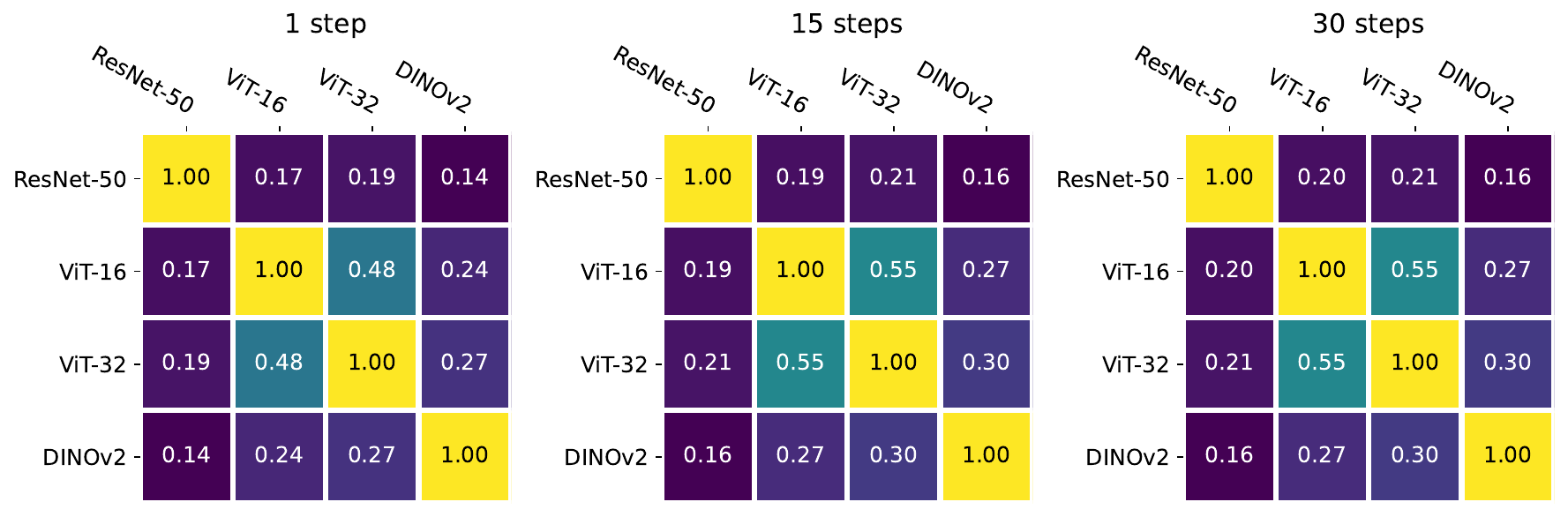}
    \end{subfigure}
  \caption{Results of RelGeo(L2) on CUB with varying number of discretization steps. Top: accuracies; bottom: MRR Cosine Sym.}
  \label{fig:num_steps_pullback}
\end{figure*}

\begin{figure*}
    \centering
    \begin{subfigure}[t]{\textwidth}
    \centering
    \includegraphics[width=0.8\linewidth]{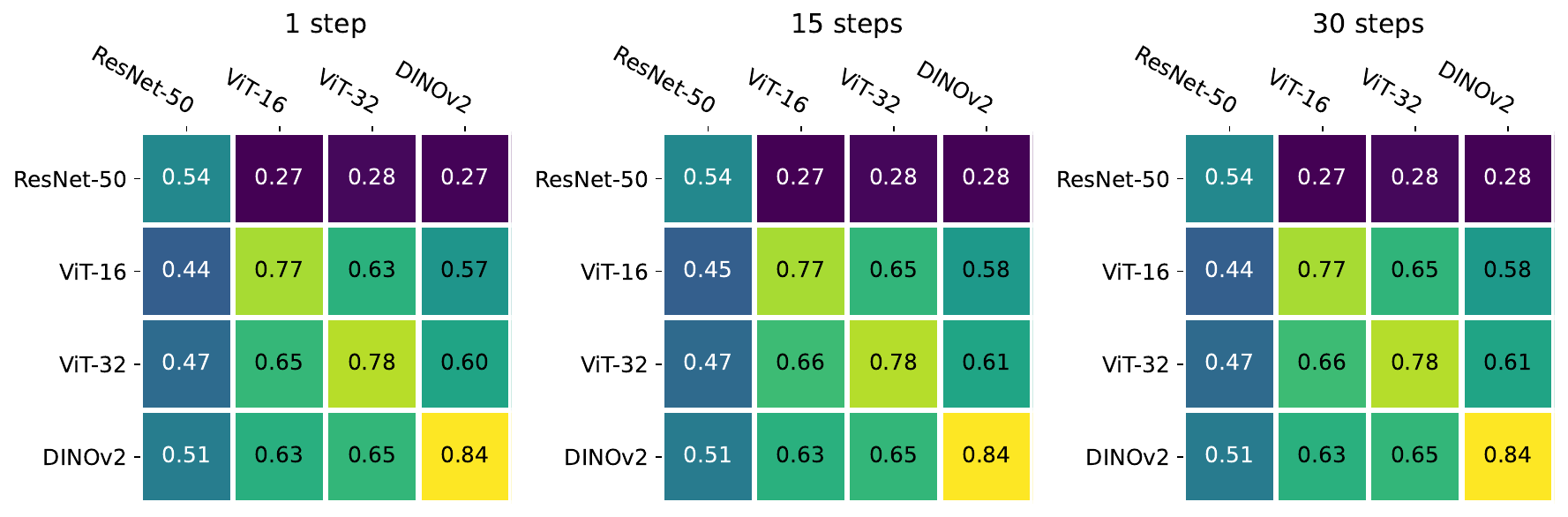}
    \end{subfigure}
    \hfill
    \begin{subfigure}[t]{\textwidth}
    \centering
    \includegraphics[width=0.8\linewidth]{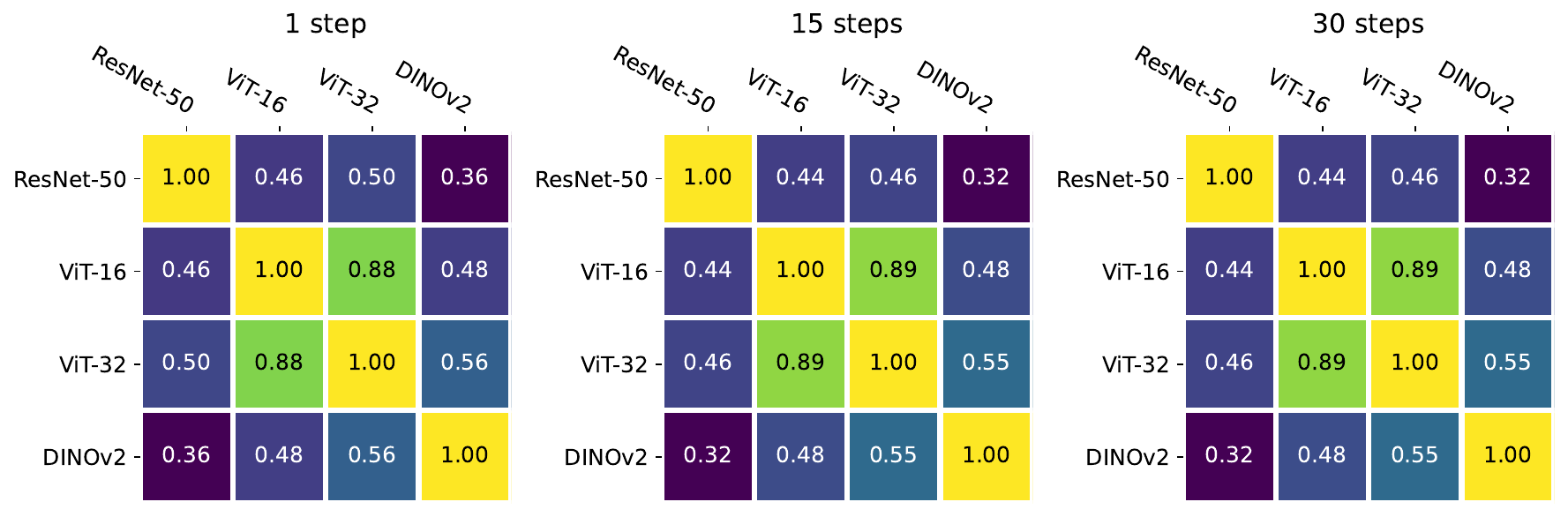}
    \end{subfigure}
  \caption{Results of RelGeo(Diet) on CUB with varying number of discretization steps. Top: accuracies; bottom: MRR Cosine Sym.}
  \label{fig:num_steps_diet}
\end{figure*}

We analyze the impact of the number of discretization steps on RelGeo(L2) and RelGeo(Diet) and provide the results in Figure~\ref{fig:num_steps_pullback} and Figure~\ref{fig:num_steps_diet}. The performances do not vary much depending on the discretization steps, though using multiple steps seems to help.

\subsubsection{Diet augmentation strengths}

\begin{figure*}
    \centering
    \begin{subfigure}[t]{\textwidth}
    \centering
    \includegraphics[width=0.8\linewidth]{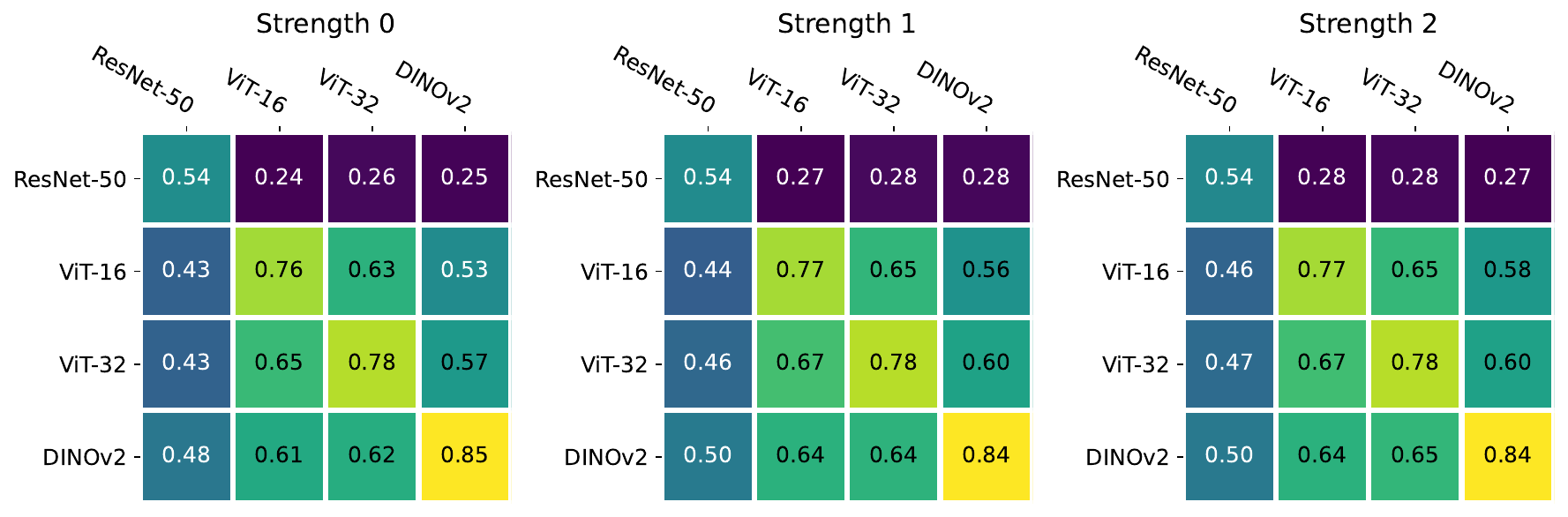}
    \end{subfigure}
    \hfill
    \begin{subfigure}[t]{\textwidth}
    \centering
    \includegraphics[width=0.8\linewidth]{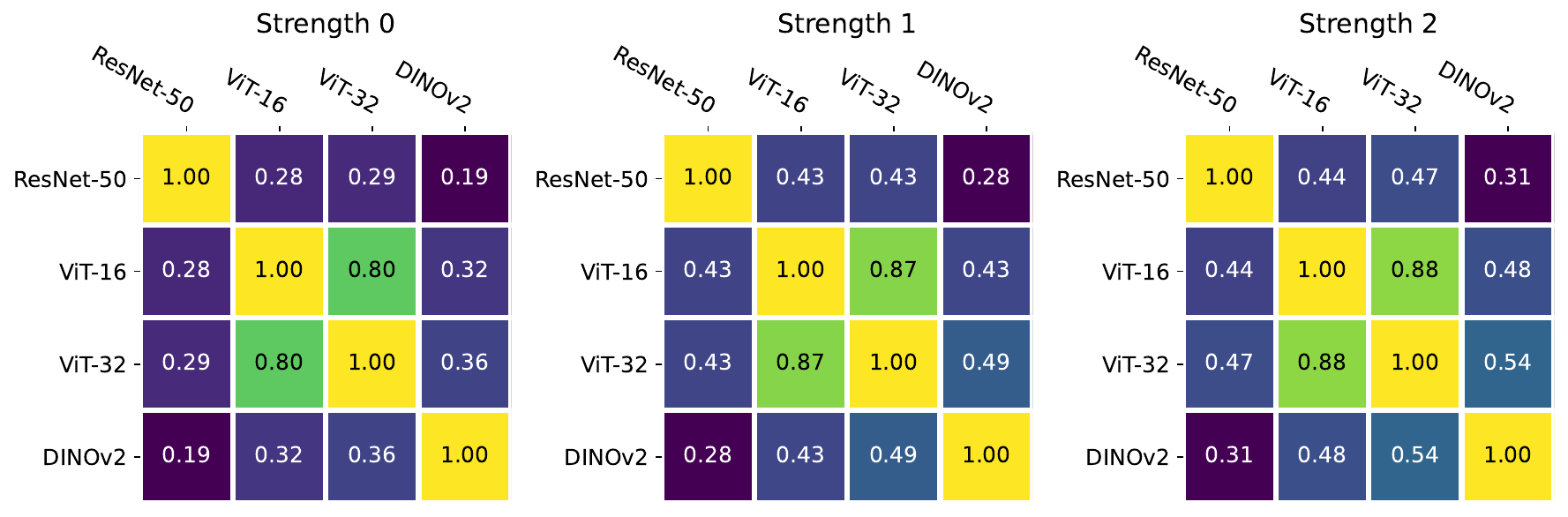}
    \end{subfigure}
  \caption{Results of RelGeo(Diet) on CUB with varying diet augmentation strengths. Results with strength 3 can be seen above. Top: accuracies; bottom: MRR Cosine Sym.}
  \label{fig:strength_diet}
\end{figure*}

We analyze the impact of different data augmentation strengths on RelGeo(Diet). The results are shown in Figure~\ref{fig:strength_diet}. Similar to the observations in terms of self-supervised learning \citep{ibrahim2024occams}, RelGeo(Diet) benefits from stronger data augmentations.

\subsubsection{Anchor selection scheme}

As discussed in \citet{Moschella2023}, there are different ways of choosing the anchors. In the main paper, we consider the case where the anchors are selected uniformly at random, referred to as \textit{uniform}. There are other choices as well, e.g. using farthest point sampling, referred to as \textit{fps}, and using as anchors the data point close to the centroids of K-means clustering, referred to as \textit{kmeans}.

Here we additionally report the results for \textit{fps} and  Since we need to align multiple models, in practice we use the selection mechanism to select a fixed number of anchors based on the representations of each model, and combine them while employing random subsampling to obtain the final anchors of a given number.

\begin{figure*}
    \centering
    \begin{subfigure}[t]{\textwidth}
    \centering
    \includegraphics[width=0.8\linewidth]{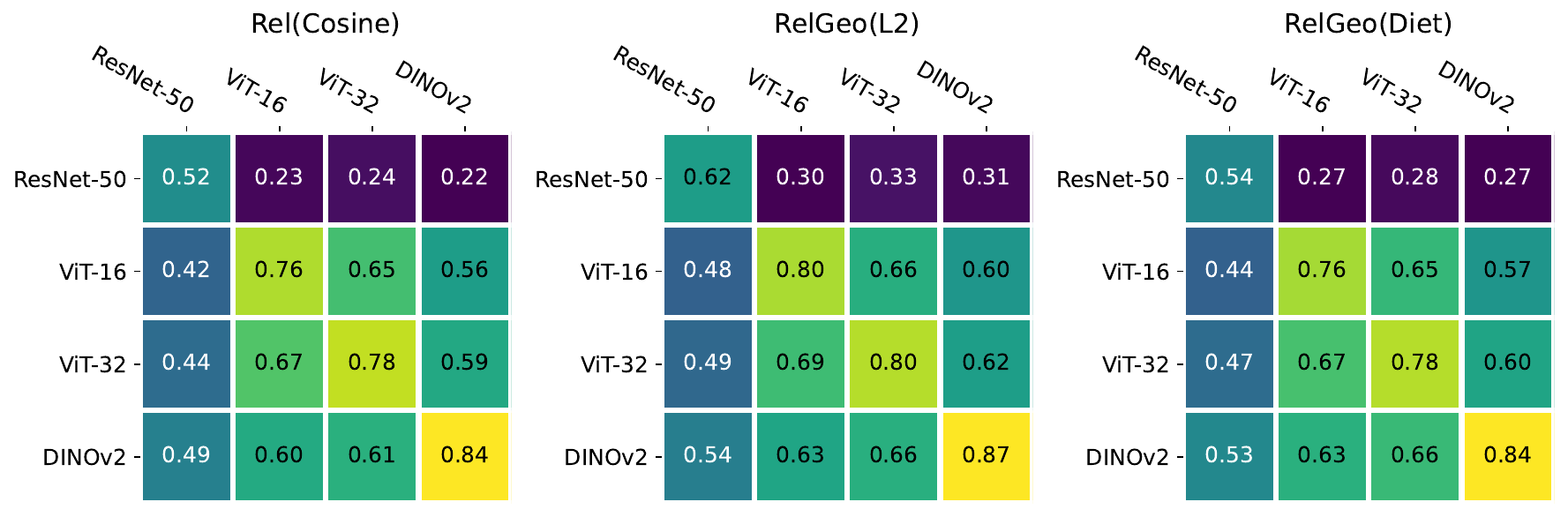}
    \end{subfigure}
    \hfill
    \begin{subfigure}[t]{\textwidth}
    \centering
    \includegraphics[width=0.8\linewidth]{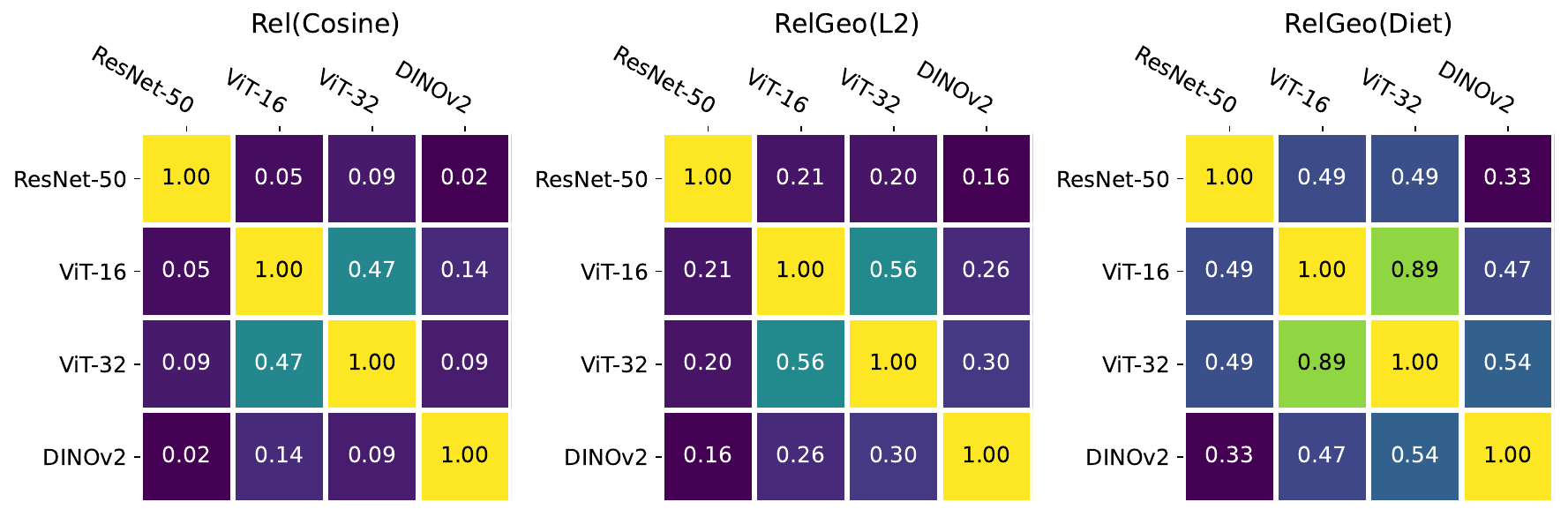}
    \end{subfigure}
    \hfill
    \begin{subfigure}[t]{\textwidth}
    \centering
    \includegraphics[width=0.8\linewidth]{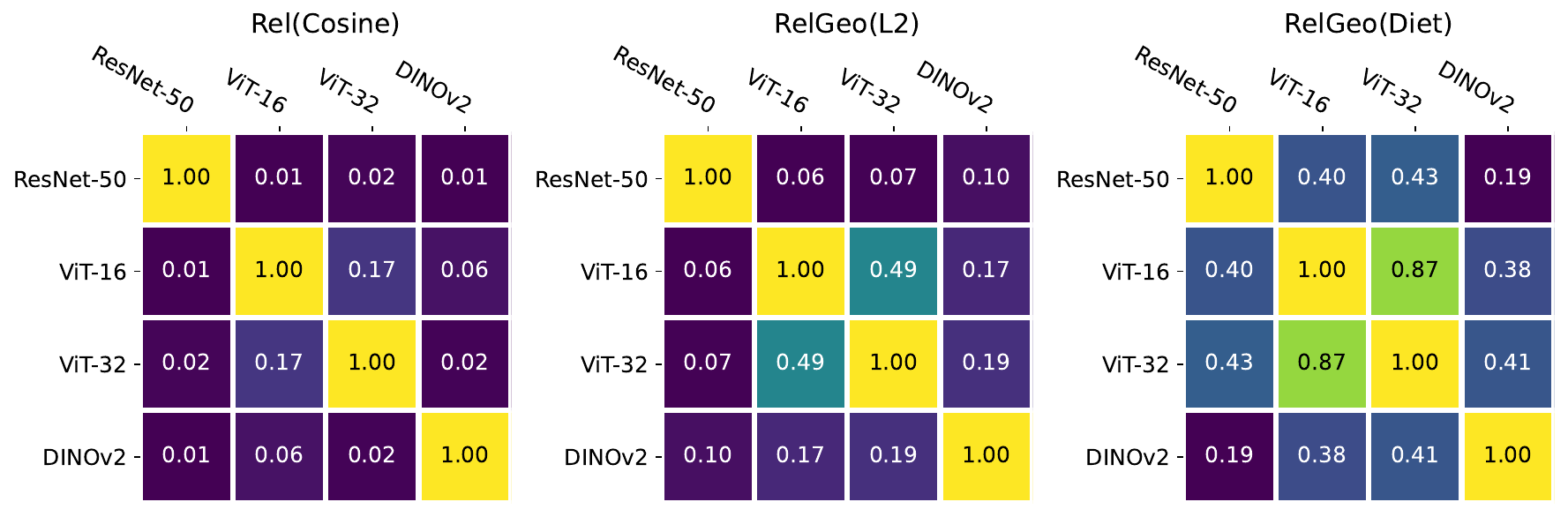}
    \end{subfigure}
    \hfill
    \begin{subfigure}[t]{\textwidth}
    \centering
    \includegraphics[width=0.8\linewidth]{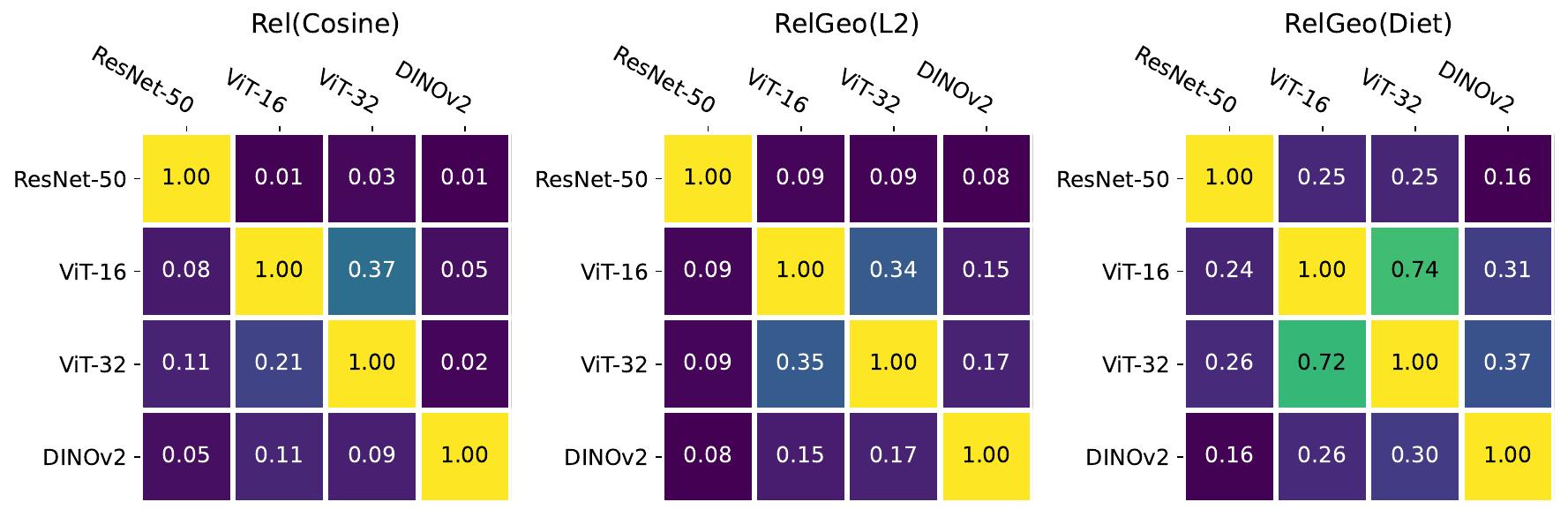}
    \end{subfigure}
    \hfill
    \begin{subfigure}[t]{\textwidth}
    \centering
    \includegraphics[width=0.8\linewidth]{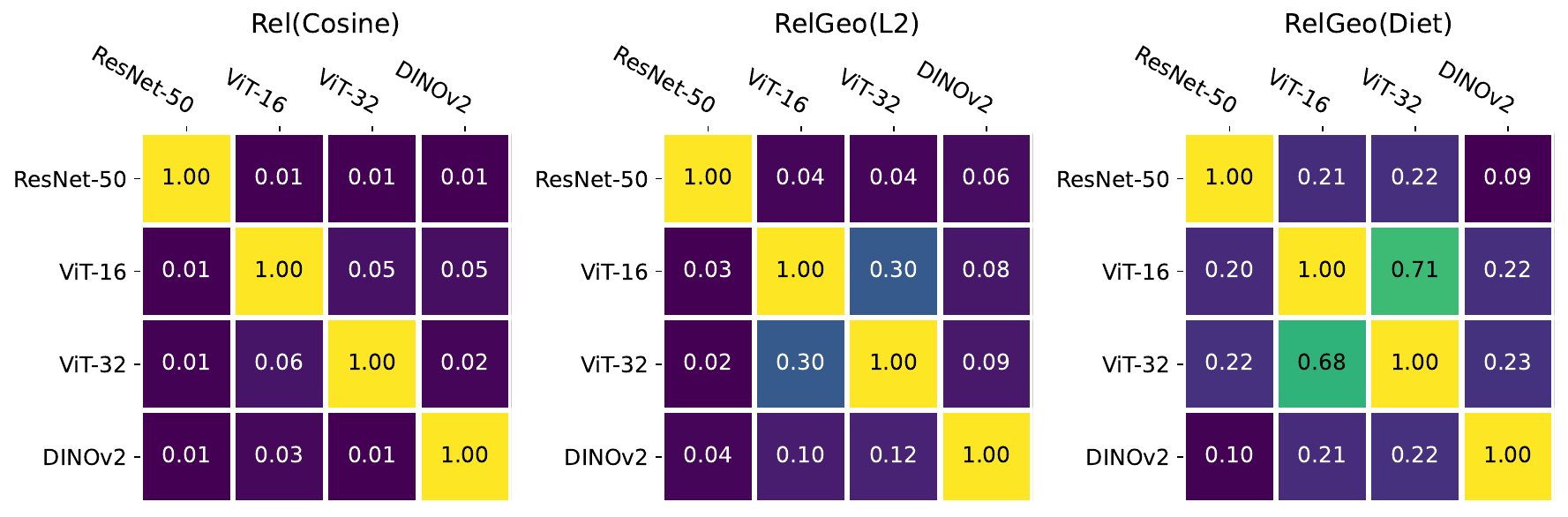}
    \end{subfigure}
  \caption{Results using the \textit{fps} scheme to select the anchors. From top to bottom: accuracies, MRR Cosine Sym, MRR CDist Sym, MRR Cosine, MRR CDist.}
  \label{fig:anchors_fps}
\end{figure*}

\begin{figure*}
    \centering
    \begin{subfigure}[t]{\textwidth}
    \centering
    \includegraphics[width=0.8\linewidth]{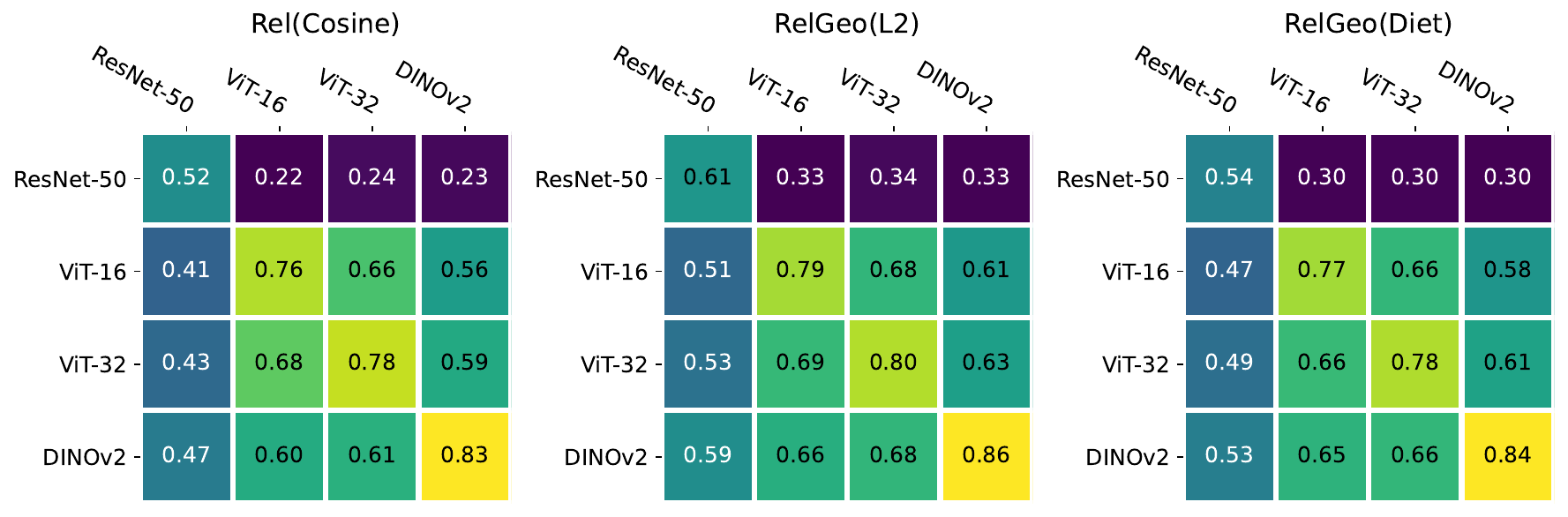}
    \end{subfigure}
    \hfill
    \begin{subfigure}[t]{\textwidth}
    \centering
    \includegraphics[width=0.8\linewidth]{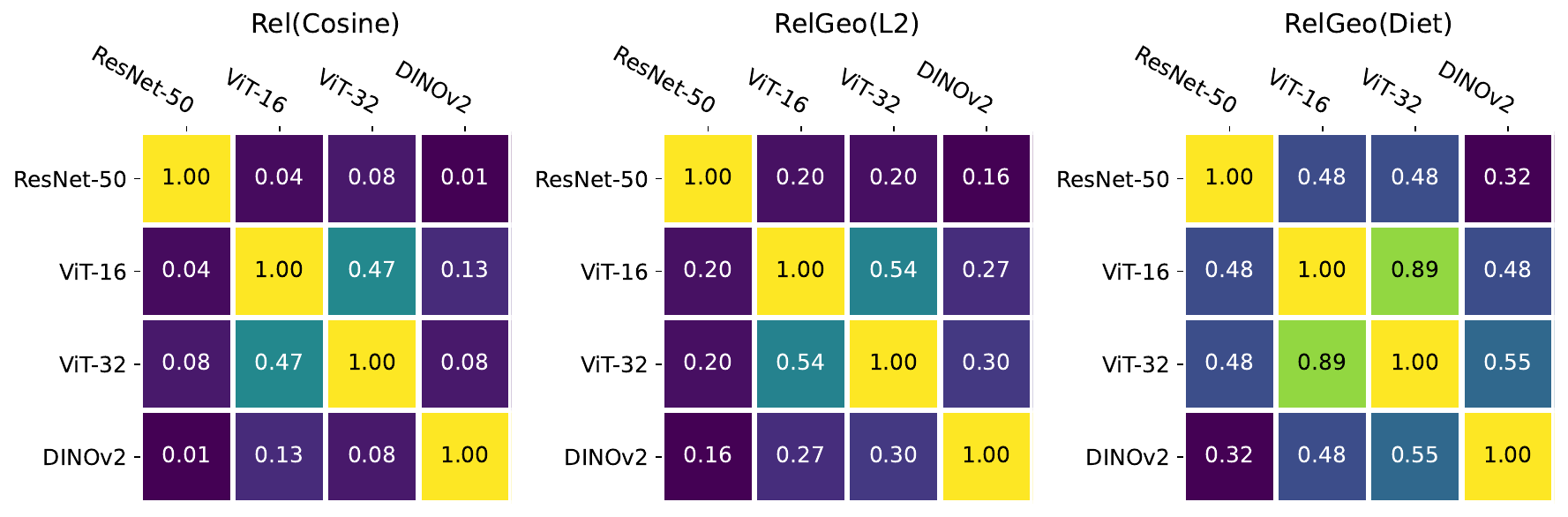}
    \end{subfigure}
    \hfill
    \begin{subfigure}[t]{\textwidth}
    \centering
    \includegraphics[width=0.8\linewidth]{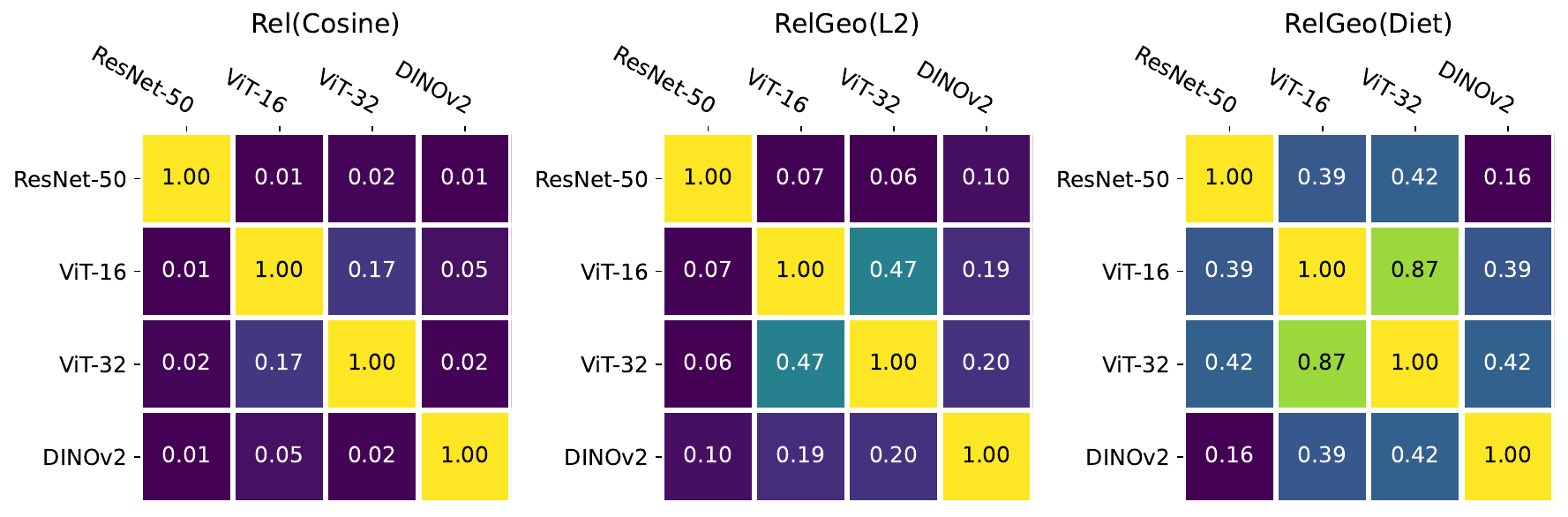}
    \end{subfigure}
    \hfill
    \begin{subfigure}[t]{\textwidth}
    \centering
    \includegraphics[width=0.8\linewidth]{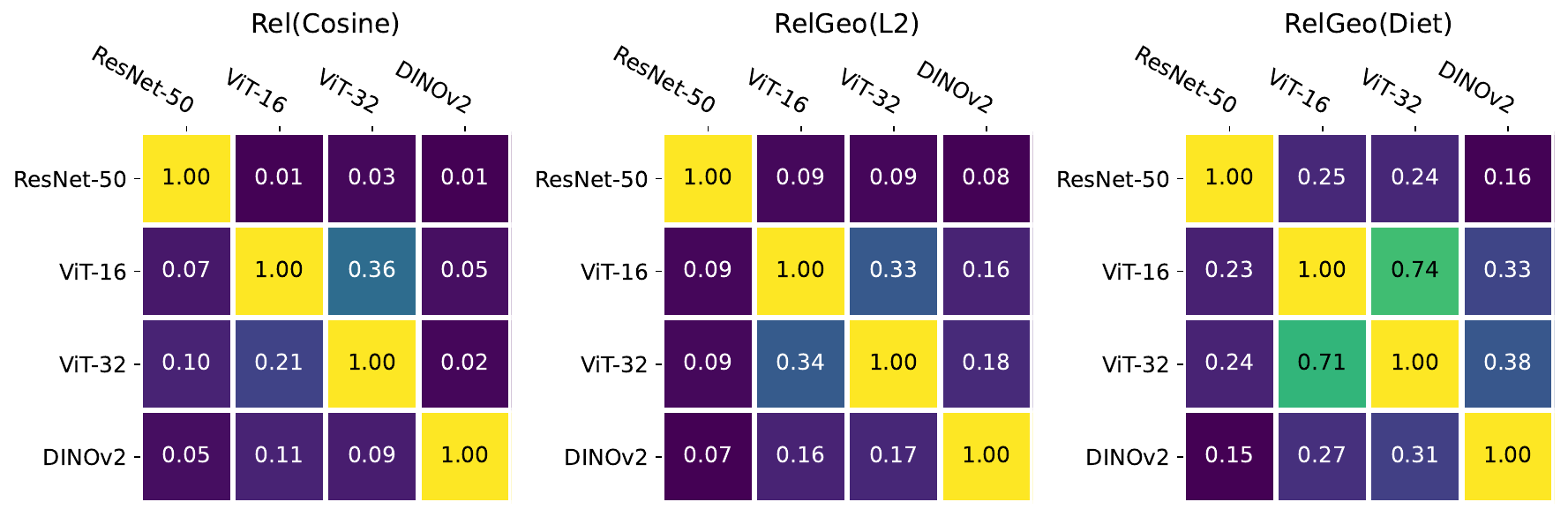}
    \end{subfigure}
    \hfill
    \begin{subfigure}[t]{\textwidth}
    \centering
    \includegraphics[width=0.8\linewidth]{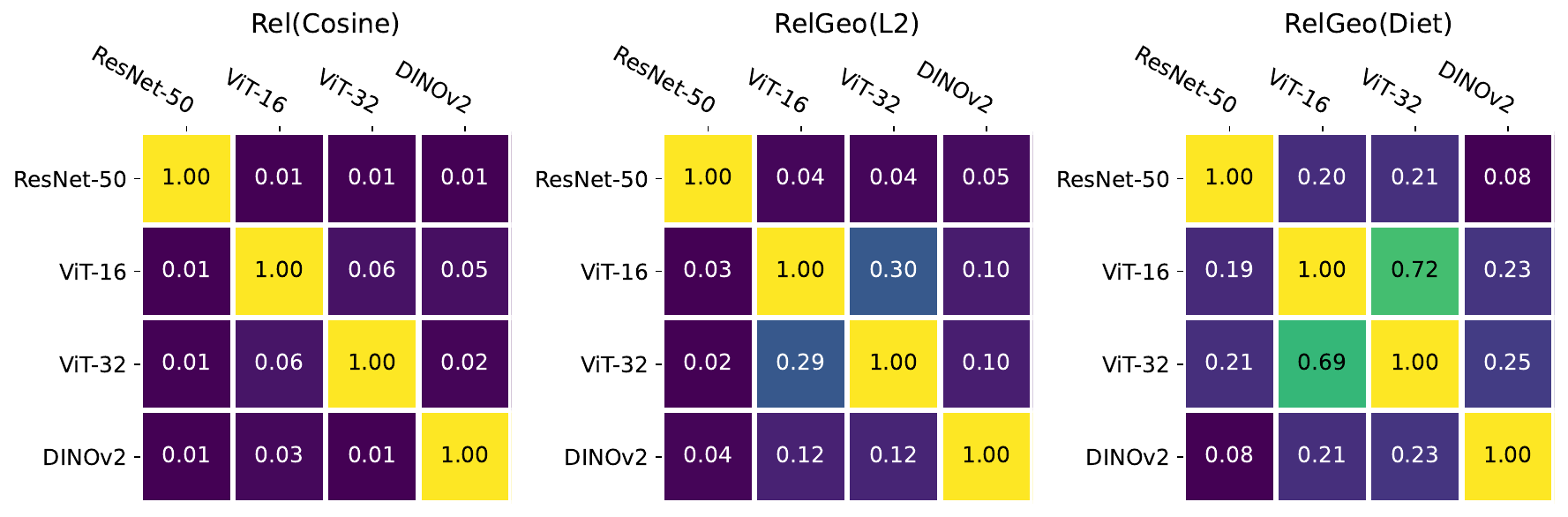}
    \end{subfigure}
  \caption{Results using the \textit{kmeans} scheme to select the anchors. From top to bottom: accuracies, MRR Cosine Sym, MRR CDist Sym, MRR Cosine, MRR CDist.}
  \label{fig:anchors_kmeans}
\end{figure*}

The experimental results for \textit{fps} and \textit{kmeans} are shown in Figure~\ref{fig:anchors_fps} and Figure~\ref{fig:anchors_kmeans}, respectively.

\subsubsection{Multimodal}
\label{app:full_multimodal_results}

\begin{figure*}
    \centering
    \begin{subfigure}[t]{\textwidth}
    \centering
    \includegraphics[width=0.65\linewidth]{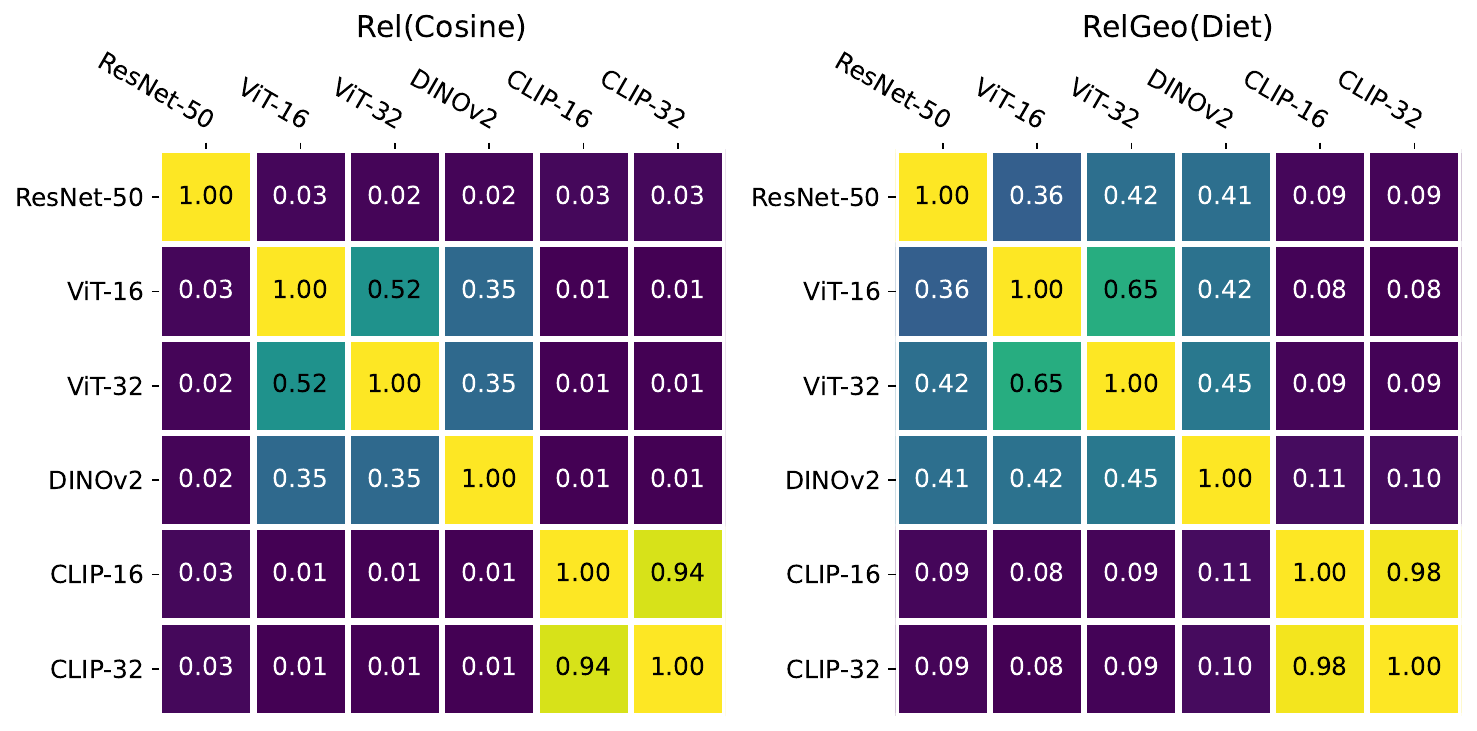}
    \end{subfigure}
    \hfill
    \begin{subfigure}[t]{\textwidth}
    \centering
    \includegraphics[width=0.65\linewidth]{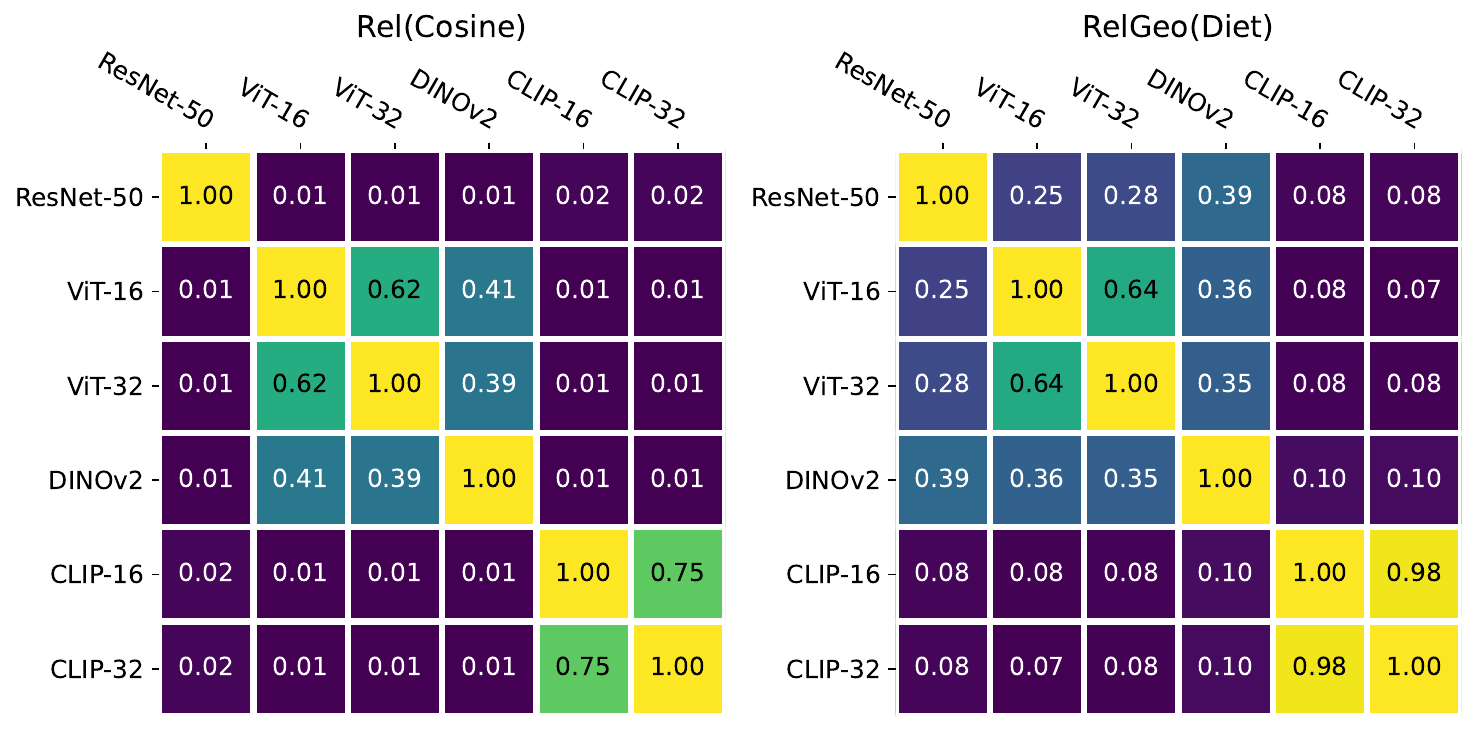}
    \end{subfigure}
    \hfill
    \begin{subfigure}[t]{\textwidth}
    \centering
    \includegraphics[width=0.65\linewidth]{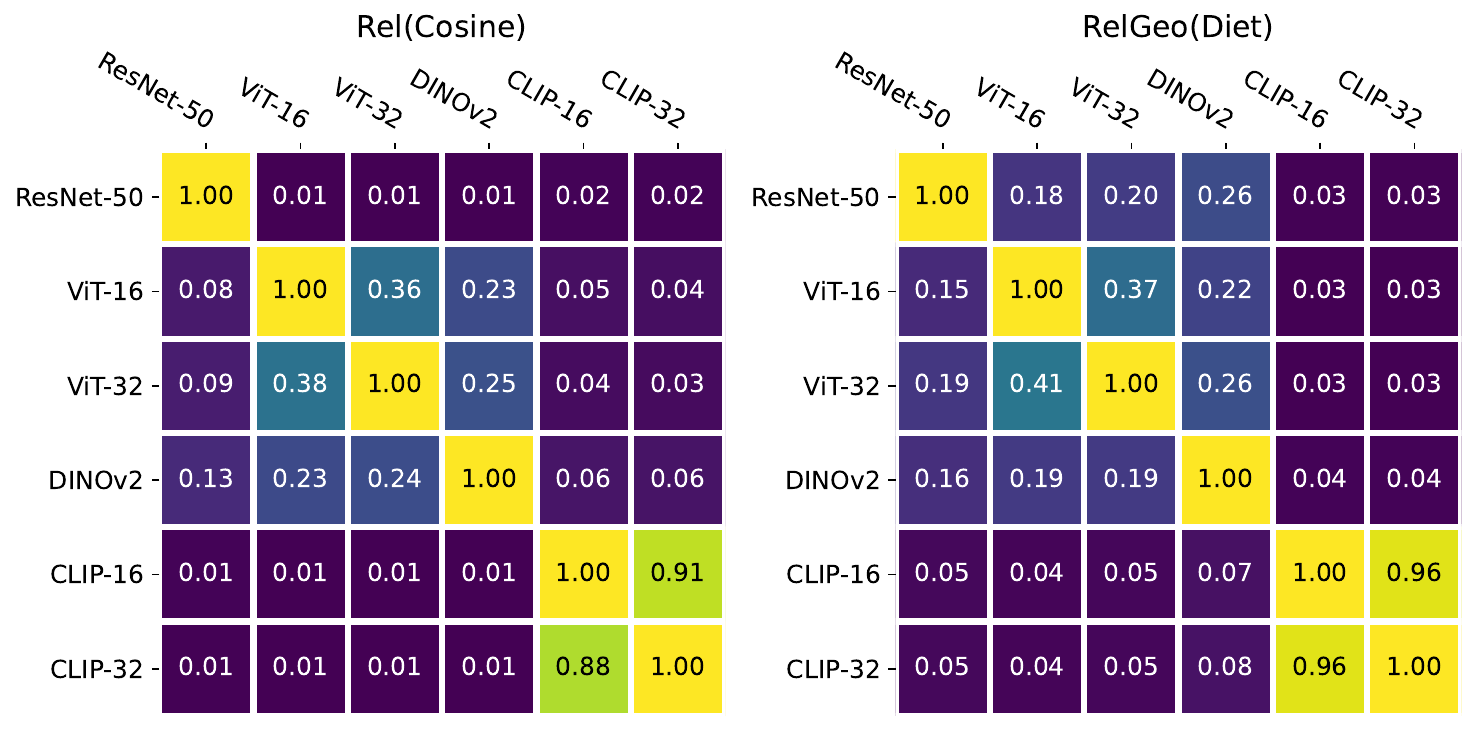}
    \end{subfigure}
    \hfill
    \begin{subfigure}[t]{\textwidth}
    \centering
    \includegraphics[width=0.65\linewidth]{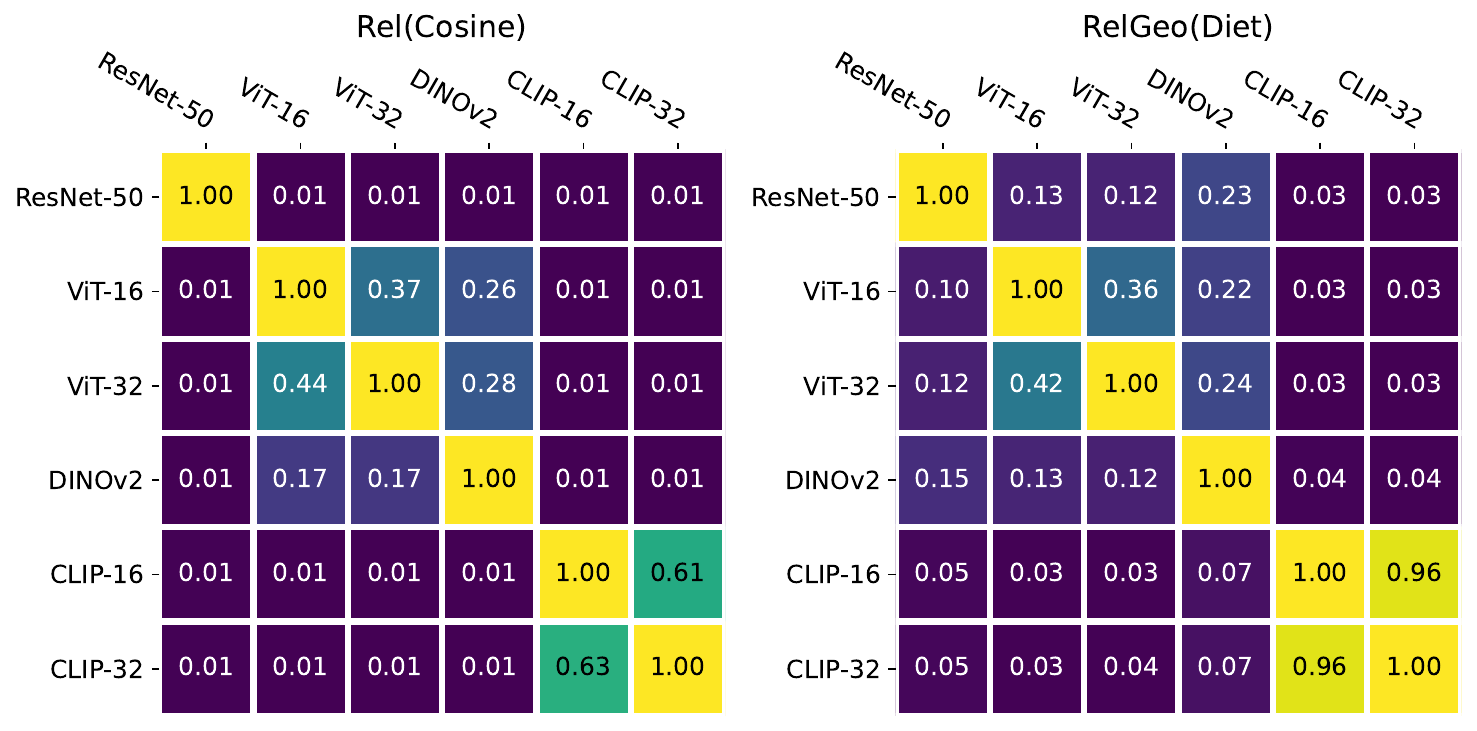}
    \end{subfigure}
  \caption{Stitching results for multimodal scenario. From top to bottom: MRR Cosine Sym, MRR CDist Sym, MRR Cosine, MRR CDist.}
  \label{fig:multimodal}
\end{figure*}

In the main paper we reported results based on MRR Cosine Sym. Below we show the full experimental results in Figure~\ref{fig:multimodal}, and report aggregated results in Table~\ref{tbl:multimodal}.

\begin{table}[h!]
  \centering
    \caption{Average MRR results for different methods across different datasets. Relative representations pulling back from diet decoder (\texttt{RelGeo(Diet)}) consistently provides better retrievals.}
  \resizebox{\linewidth}{!}{%
    \begin{tabular}{lccccc}
      \rowcolor{gray!20}\toprule
      \textbf{Method} & \textbf{MRR Cosine Sym} & \textbf{MRR CDist Sym} & \textbf{MRR Cosine} & \textbf{MRR CDist} \\
      \midrule
      \texttt{Rel(Cosine)} \citep{Moschella2023}& $0.298 \pm 0.395$ & $0.293 \pm 0.389$ & $0.283 \pm 0.382$ & $0.252 \pm 0.373$ \\
      \texttt{RelGeo(Diet)} & $\mathbf{0.413} \pm 0.353$ & $\mathbf{0.384} \pm 0.359$ & $\mathbf{0.317} \pm 0.372$ & $\mathbf{0.302} \pm 0.378$ \\
      \bottomrule
    \end{tabular}
  }
\label{tbl:multimodal}
\end{table}

\subsubsection{Running times}
\label{sec:running_times}

We report running times for autoencoder experiments with an NVIDIA 3080 Ti GPU and vision foundational model experiments with an NVIDIA A100 GPU.

\paragraph{Times to train the models}

For autoencoder experiments, training the autoencoders is not expensive, as it merely takes around $20$ minutes on an RTX3080Ti for $50$ epochs.

For vision foundation models, the employed pretrained backbones are typically computationally costly to train. For instance, \citet{Oquab2024} reported that training DINOv2 ViT-L/14 on ImagetNet-22k using 96 A100-80GB GPUs takes approximately 3.3 days. In contrast, training the decoders is much faster. We report the running times to train the decoders in Table~\ref{tbl:dec_time}, where for each setting we average over the different models to obtain uncertainty estimates. We remark that the exact running times depend heavily on implementation details, while currently we focus on correctness instead of the speed, and the running times could possibly be improved with better implementations.

\begin{table*}[h!]
  \centering
  \caption{Times in seconds for training the decoders.}
   \resizebox{\linewidth}{!}{%
  \begin{tabular}{lccccc}
    \rowcolor{gray!20}\toprule
    \textbf{Decoder} & \textbf{CIFAR-10} & \textbf{CIFAR-100} & \textbf{ImageNet-1k} & \textbf{CUB} & \textbf{SVHN} \\
    \midrule
    \texttt{Abs}       & $10.467 \pm 0.765$ & $10.71 \pm 0.655$ & $8.804 \pm 0.581$ & $1.593 \pm 0.615$ & $15.57 \pm 0.843$ \\
    \texttt{Diet(0)}  & $3.462 \pm 0.03$ & $3.501 \pm 0.045$ & $3.556 \pm 0.018$ & $3.509 \pm 0.028$ & $3.506 \pm 0.044$ \\
    \texttt{Diet(1)}      & $718.818 \pm 216.179$ & $730.446 \pm 226.91$ & $1651.49 \pm 211.037$ & $1937.656 \pm 192.894$ & $705.929 \pm 202.844$ \\
    \texttt{Diet(2)}      & $735.144 \pm 219.211$ & $739.671 \pm 222.36$ & $2034.677 \pm 276.309$ & $2428.942 \pm 196.388$ & $727.477 \pm 206.929$ \\
    \texttt{Diet(3)}      & $771.299 \pm 213.022$ & $767.453 \pm 222.62$ & $2645.914 \pm 260.069$ & $3201.588 \pm 242.339$ & $754.965 \pm 202.021$ \\
    \bottomrule
  \end{tabular}
  }
  \label{tbl:dec_time}
\end{table*}

\paragraph{Times to obtain the representations}

We first investigate the running times - accuracy tradeoff of RelGeo representations on CUB dataset, where we vary the number of anchors and monitor the times to obtain the representations and the qualities of the resulting representations. We report the results on autoencoders in Table~\ref{tbl:time_ae} and the results on vision foundation models in Table~\ref{tbl:time_cosine}, Table~\ref{tbl:time_pullback} and Table~\ref{tbl:time_diet}, respectively.

\begin{table*}
    \centering
    \caption{Running time / accuracy tradeoff of RelGeo(L2).}
    \resizebox{0.5\linewidth}{!}{%
      \begin{tabular}{lcc}
        \rowcolor{gray!20}\toprule
        \textbf{Num Anchors} & \textbf{Time (s) $\pm$ Std} & \textbf{MRR} \\
        \midrule
        \texttt{2}       & $0.8480 \pm 0.0407$ & $0.0168$ \\
        \texttt{3}       & $0.8348 \pm 0.0367$ & $0.0807$ \\
        \texttt{5}       & $0.8377 \pm 0.0435$ & $0.3503$ \\
        \texttt{8}       & $0.9450 \pm 0.0285$ & $0.7004$ \\
        \texttt{10}       & $1.0969 \pm 0.0297$ & $0.8384$ \\
        \texttt{15}       & $1.2853 \pm 0.0264$ & $0.9296$ \\
        \texttt{20}       & $1.6160 \pm 0.0188$ & $0.9616$ \\
        \texttt{25}       & $1.8548 \pm 0.0218$ & $0.9868$ \\
        \texttt{50}       & $3.3311 \pm 0.0286$ & $0.9981$ \\
        \texttt{100}       & $6.2122 \pm 0.0318$ & $0.9986$ \\
        \texttt{300}       & $17.6494 \pm 0.0653$ & $0.9982$ \\
        \texttt{500}       & $29.1925 \pm 0.0806$ & $0.9986$ \\
        \bottomrule
      \end{tabular}
    }
    \label{tbl:time_ae}
\end{table*}

\begin{table*}
    \centering
    \caption{Running time / accuracy tradeoff of Rel(Cosine).}
    \resizebox{0.4\linewidth}{!}{%
      \begin{tabular}{lcc}
        \rowcolor{gray!20}\toprule
        \textbf{Num Anchors} & \textbf{Time (s)} & \textbf{Accuracy} \\
        \midrule
        \texttt{200}       & $0.088 \pm 0.14$ & $0.425 \pm 0.189$ \\
        \texttt{500}  & $0.175 \pm 0.053$ & $0.531 \pm 0.188$ \\
        \texttt{1000}      & $0.062 \pm 0.091$ & $0.559 \pm 0.179$ \\
        \bottomrule
      \end{tabular}
    }
    \label{tbl:time_cosine}
\end{table*}

\begin{table*}
    \centering
    \caption{Running time / accuracy tradeoff of RelGeo(L2).}
    \resizebox{0.4\linewidth}{!}{%
      \begin{tabular}{lcc}
        \rowcolor{gray!20}\toprule
        \textbf{Num Anchors} & \textbf{Time (s)} & \textbf{Accuracy} \\
        \midrule
        \texttt{200}       & $8.004 \pm 3.304$ & $0.5 \pm 0.184$ \\
        \texttt{500}  & $21.297 \pm 8.774$ & $0.595 \pm 0.163$ \\
        \texttt{1000}      & $47.218 \pm 19.486$ & $0.619 \pm 0.154$ \\
        \bottomrule
      \end{tabular}
    }
    \label{tbl:time_pullback}
\end{table*}

\begin{table*}
    \centering
    \caption{Running time / accuracy tradeoff of RelGeo(Diet).}
    \resizebox{0.4\linewidth}{!}{%
      \begin{tabular}{lcc}
        \rowcolor{gray!20}\toprule
        \textbf{Num Anchors} & \textbf{Time (s)} & \textbf{Accuracy} \\
        \midrule
        \texttt{200}       & $7.274 \pm 3.28$ & $0.459 \pm 0.177$ \\
        \texttt{500}  & $19.427 \pm 8.771$ & $0.559 \pm 0.171$ \\
        \texttt{1000}      & $43.108 \pm 19.502$ & $0.585 \pm 0.163$ \\
        \bottomrule
      \end{tabular}
    }
    \label{tbl:time_diet}
\end{table*}

We then investigate the times to evaluate the representations across different datasets and report the results in Table~\ref{tbl:time}, under the same experimental settings as reported in the main paper.

\begin{table*}[h!]
  \centering
  \caption{Times in seconds of \texttt{Rel(Cosine)}, \texttt{RelGeo(L2)} and \texttt{RelGeo(Diet)} for generating the representations.}
   \resizebox{\linewidth}{!}{%
  \begin{tabular}{lccccc}
    \rowcolor{gray!20}\toprule
    \textbf{Method} & \textbf{CIFAR-10} & \textbf{CIFAR-100} & \textbf{ImageNet-1k} & \textbf{CUB} & \textbf{SVHN} \\
    \midrule
    \texttt{Rel(Cosine)} \citep{Moschella2023}       & $0.071 \pm 0.113$ & $0.184 \pm 0.065$ & $0.085 \pm 0.124$ & $0.05 \pm 0.071$ \\
    \texttt{RelGeo(L2)}  & $85.414 \pm 40.61$ & $86.079 \pm 40.652$ & $258.788 \pm 70.064$ & $139.998 \pm 66.509$ \\
    \texttt{RelGeo(Diet)}      & $89.899 \pm 40.599$ & $89.902 \pm 40.588$ & $155.991 \pm 70.136$ & $147.336 \pm 66.505$ \\
    \bottomrule
  \end{tabular}
  }
  \label{tbl:time}
\end{table*}

\subsubsection{RelGeo(Fisher)}
\label{app:relgeo_fisher}

\begin{figure*}
    \centering
    \begin{subfigure}[t]{\textwidth}
    \centering
    \includegraphics[width=0.8\linewidth]{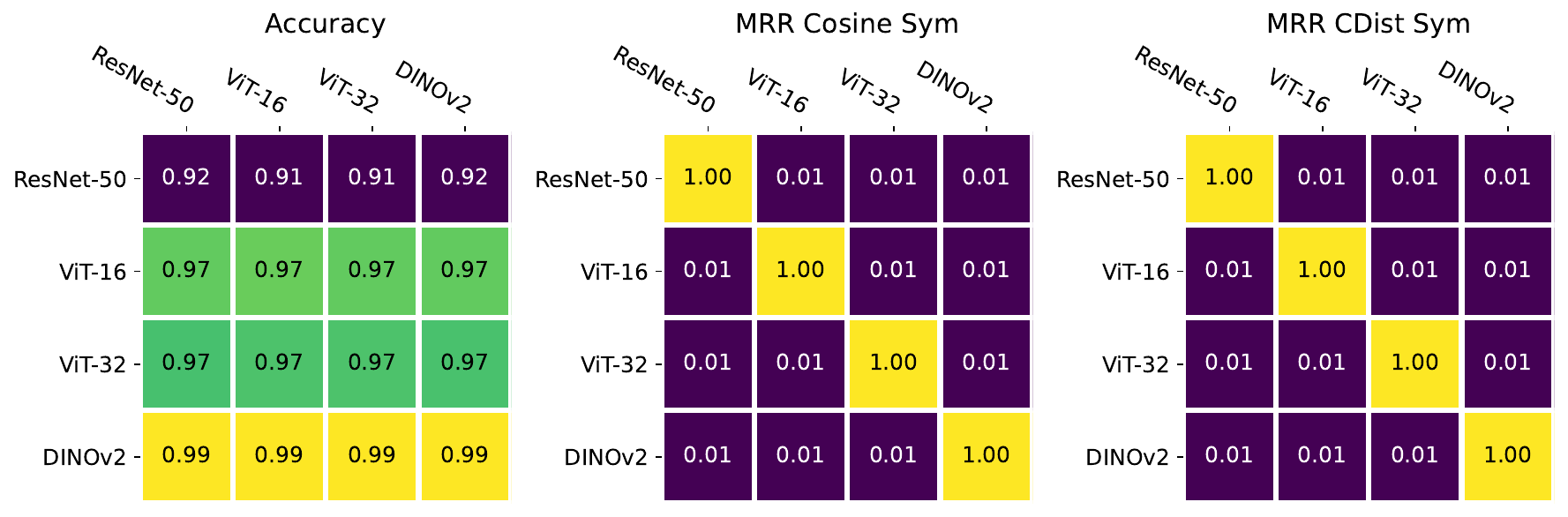}
    \end{subfigure}
    \hfill
    \begin{subfigure}[t]{\textwidth}
    \centering
    \includegraphics[width=0.53\linewidth]{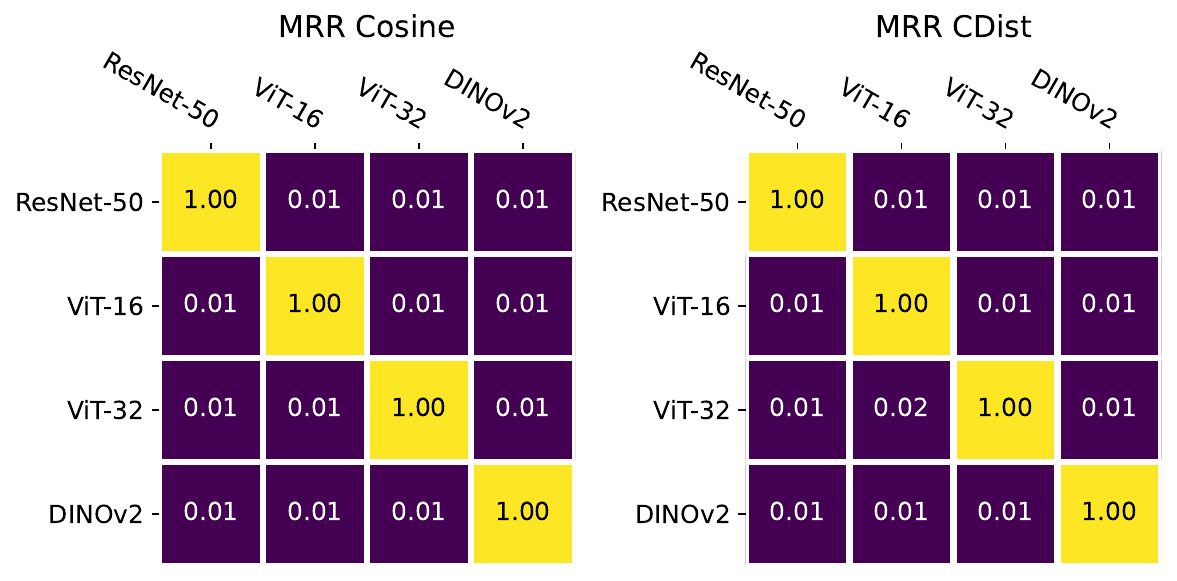}
    \end{subfigure}
    \caption{Results of RelGeo(Fisher) on CIFAR-10.}
  \label{fig:ig_cifar10}
\end{figure*}

\begin{figure*}
    \centering
    \begin{subfigure}[t]{\textwidth}
    \centering
    \includegraphics[width=0.8\linewidth]{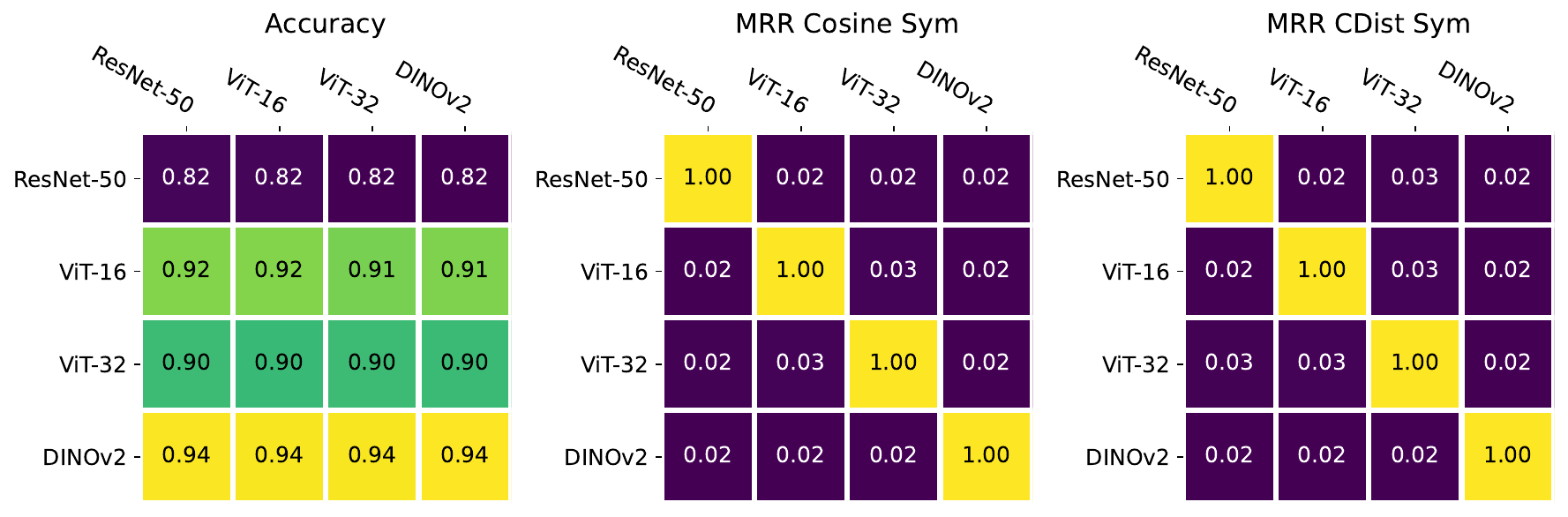}
    \end{subfigure}
    \hfill
    \begin{subfigure}[t]{\textwidth}
    \centering
    \includegraphics[width=0.53\linewidth]{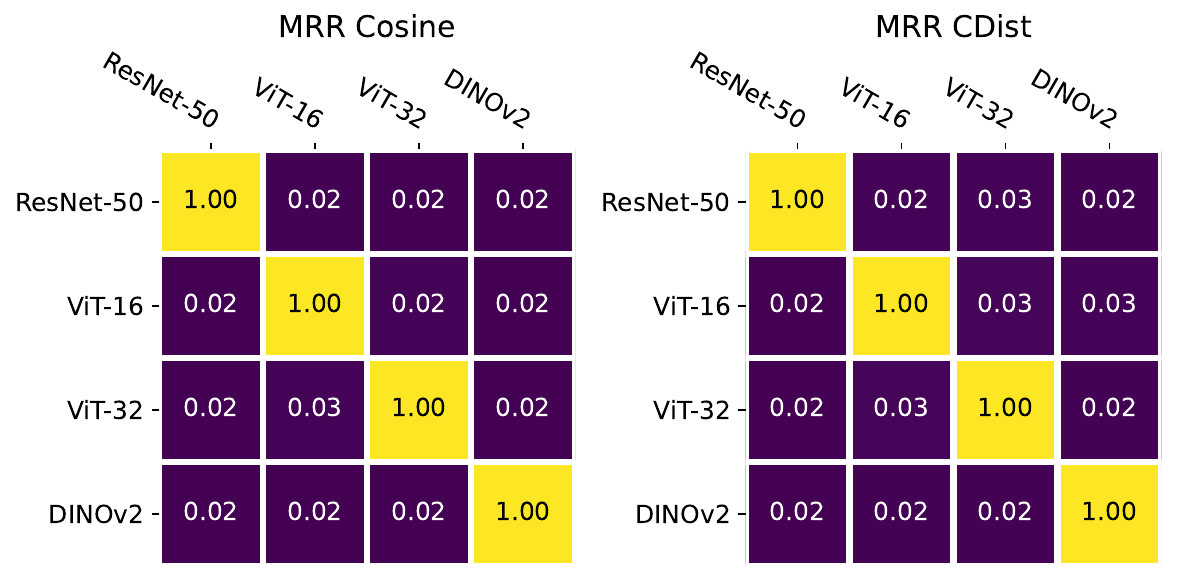}
    \end{subfigure}
    \caption{Results of RelGeo(Fisher) on CIFAR-100.}
  \label{fig:ig_cifar100}
\end{figure*}

\begin{figure*}
    \centering
    \begin{subfigure}[t]{\textwidth}
    \centering
    \includegraphics[width=0.8\linewidth]{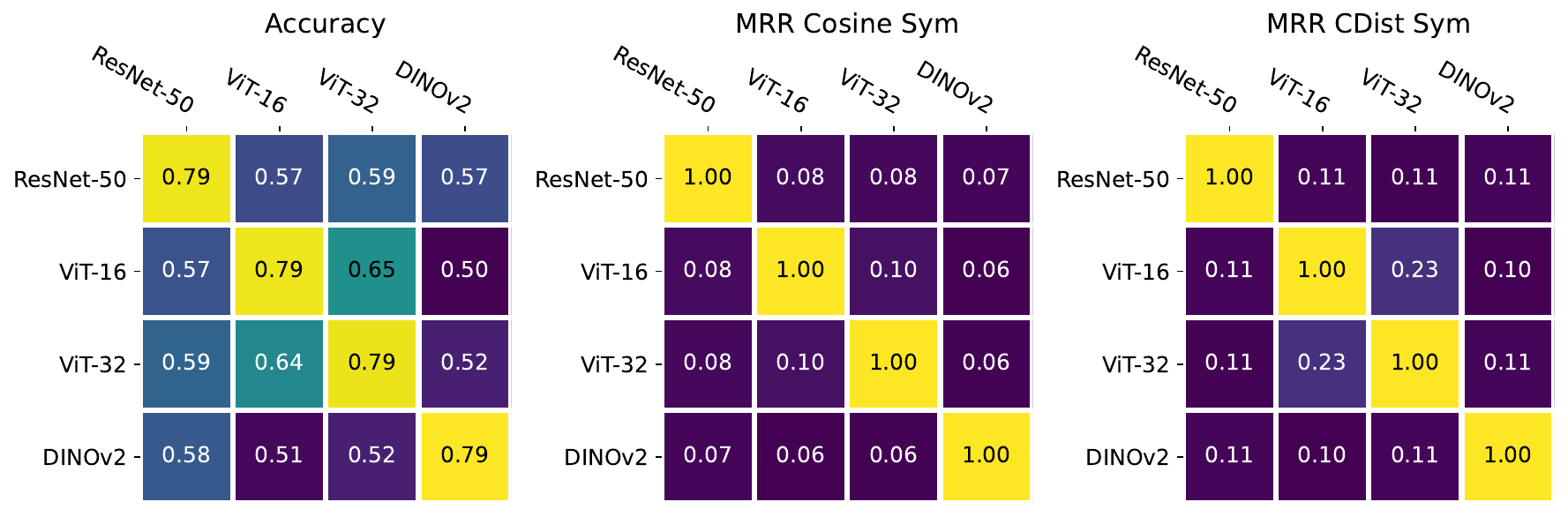}
    \end{subfigure}
    \hfill
    \begin{subfigure}[t]{\textwidth}
    \centering
    \includegraphics[width=0.53\linewidth]{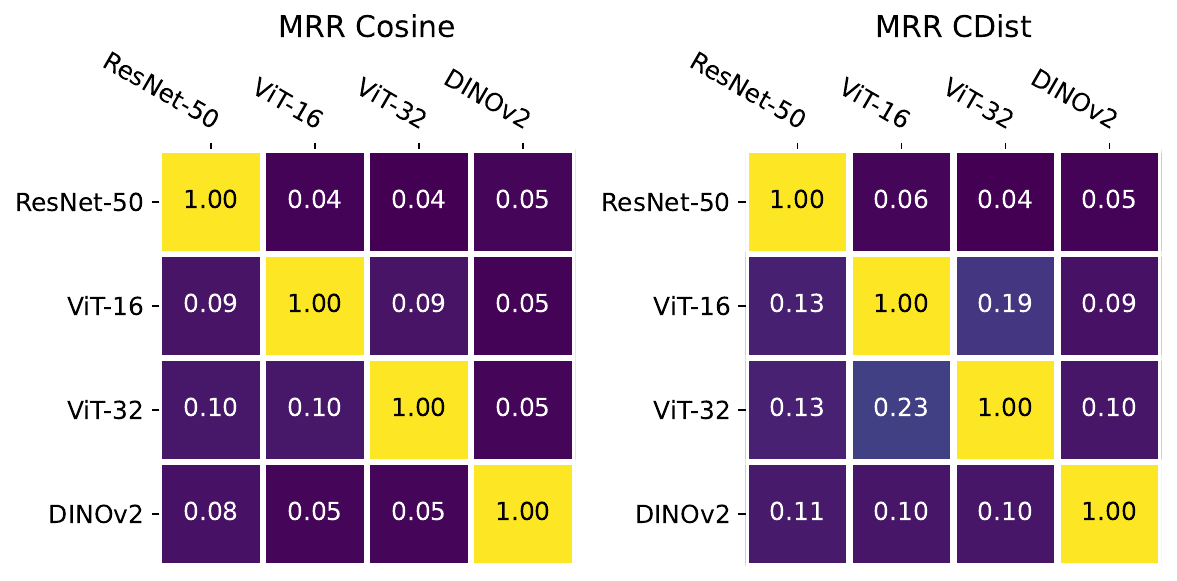}
    \end{subfigure}
\caption{Results of RelGeo(Fisher) on ImageNet-1k.}
  \label{fig:ig_imagenet1k}
\end{figure*}

\begin{figure*}
    \centering
    \begin{subfigure}[t]{\textwidth}
    \centering
    \includegraphics[width=0.8\linewidth]{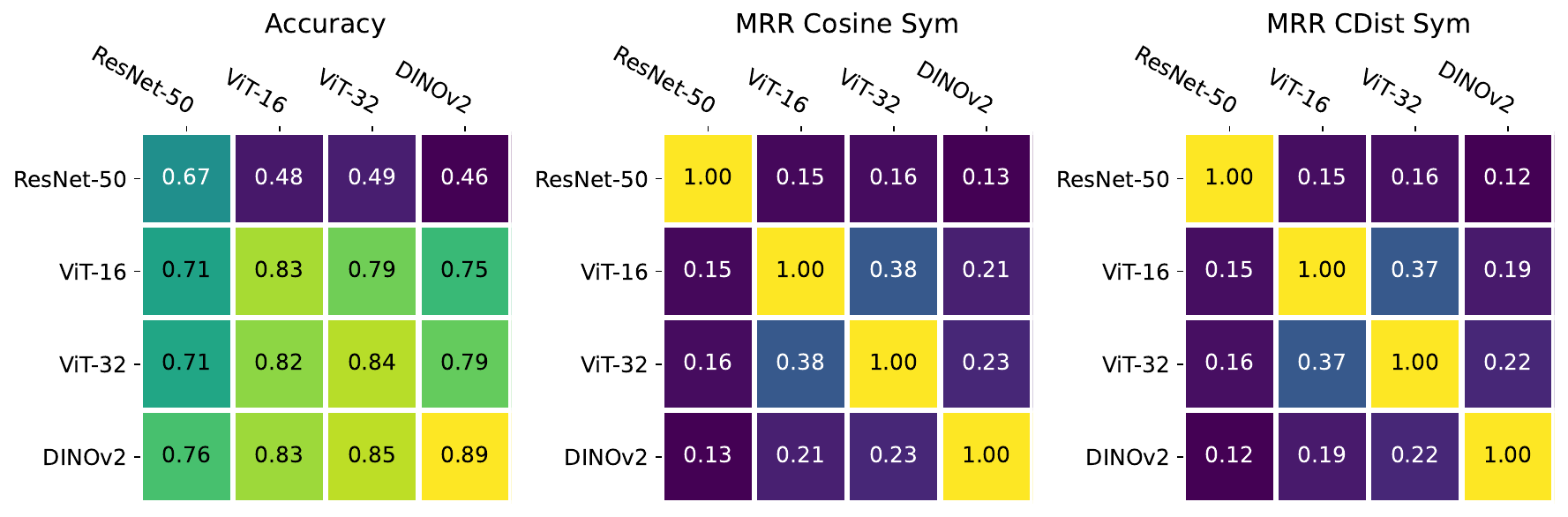}
    \end{subfigure}
    \hfill
    \begin{subfigure}[t]{\textwidth}
    \centering
    \includegraphics[width=0.53\linewidth]{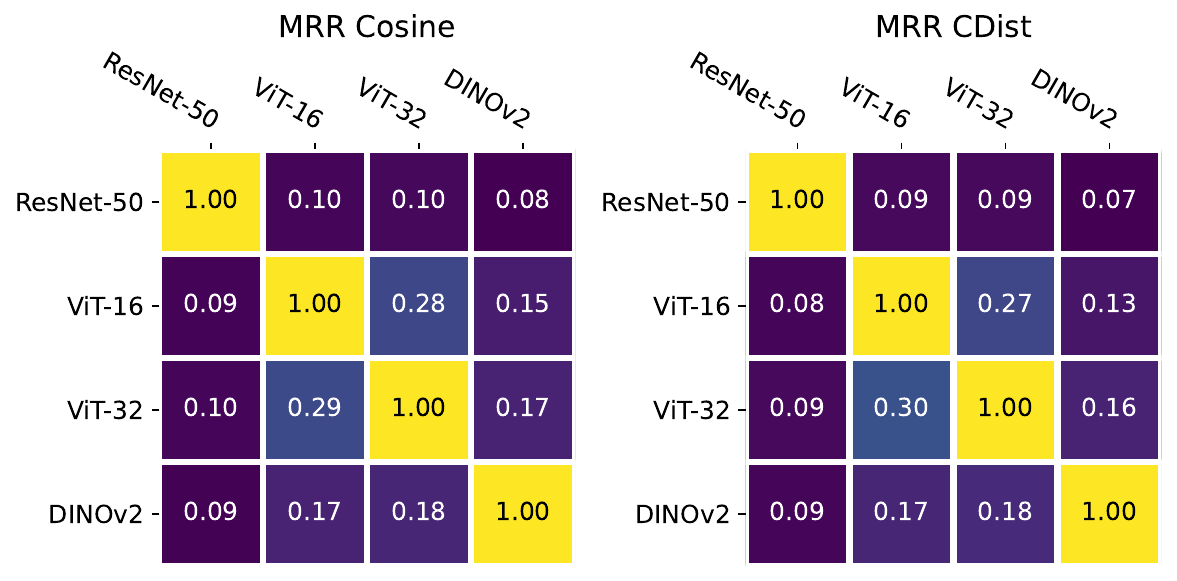}
    \end{subfigure}
\caption{Results of RelGeo(Fisher) on CUB.}
  \label{fig:ig_cub}
\end{figure*}

\begin{figure*}
    \centering
    \begin{subfigure}[t]{\textwidth}
    \centering
    \includegraphics[width=0.8\linewidth]{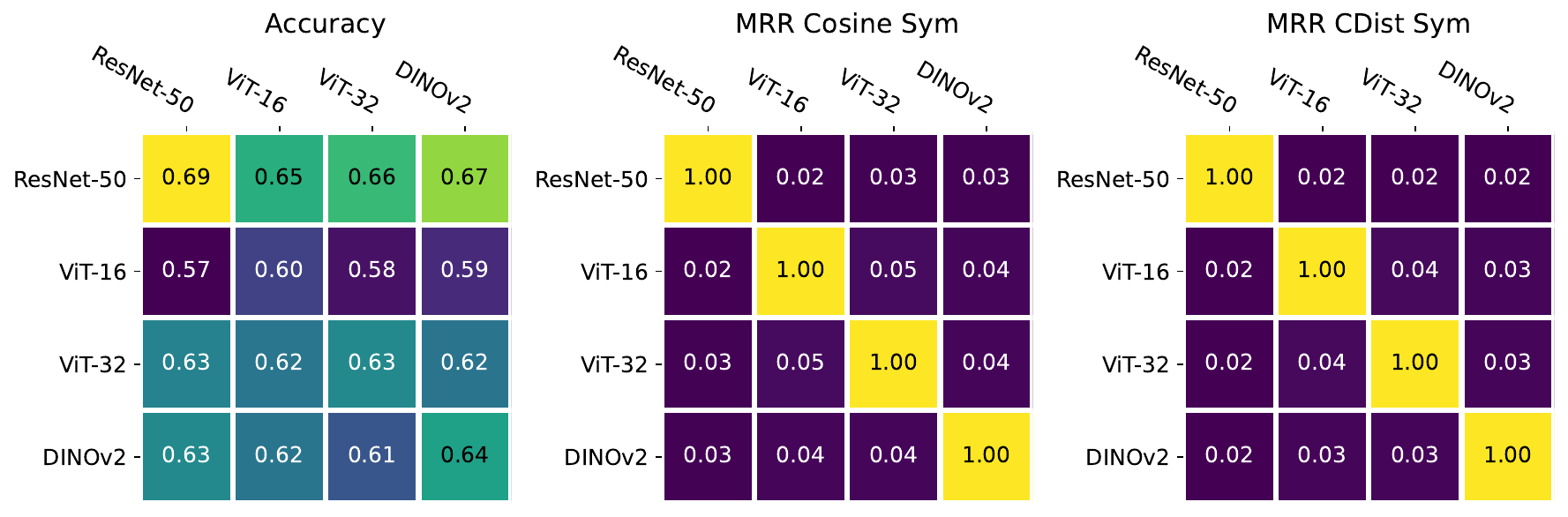}
    \end{subfigure}
    \hfill
    \begin{subfigure}[t]{\textwidth}
    \centering
    \includegraphics[width=0.53\linewidth]{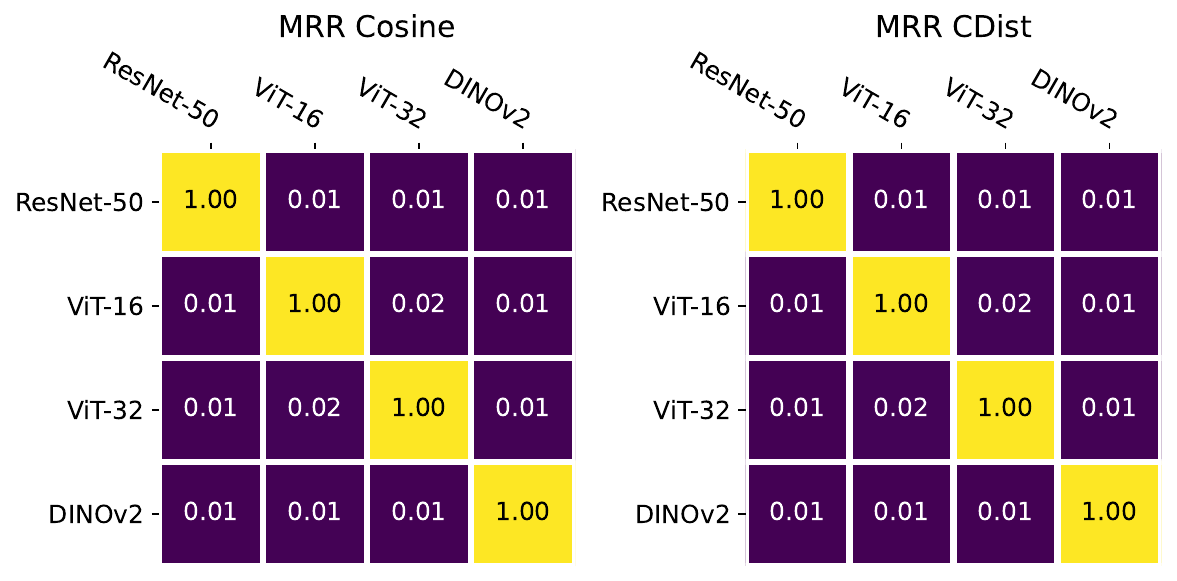}
    \end{subfigure}
  \caption{Results of RelGeo(Fisher) on SVHN.}
  \label{fig:ig_svhn}
\end{figure*}

We additionally report the results of relative geodesic representations based on another choice of Rimennian metric, RelGeo(Fisher), which pulls back the Fisher-Rao metric from the classification heads' output probabilities on the different datasets. We report the aggregated results in Table~\ref{tbl:ig}, and the results on the individual datasets in Figure~\ref{fig:ig_cifar10}, Figure~\ref{fig:ig_cifar100}, Figure~\ref{fig:ig_imagenet1k}, Figure~\ref{fig:ig_cub} and Figure~\ref{fig:ig_svhn}. RelGeo(Fisher) often results in higher accuracies and lower MRRs; we hypothesize that this is due to the Neural Collapse phenomenon \citep{kothapalli2023neural} observed in well-trained neural networks.

\begin{table*}[h!]
  \centering
  \caption{Results of RelGeo(Fisher).}
   \resizebox{\linewidth}{!}{%
  \begin{tabular}{lccccc}
    \rowcolor{gray!20}\toprule
    \textbf{Metric} & \textbf{CIFAR-10} & \textbf{CIFAR-100} & \textbf{ImageNet-1k} & \textbf{CUB} & \textbf{SVHN} \\
    \midrule
    \texttt{Accuracy} & $0.959 \pm 0.026$ & $0.894 \pm 0.043$ & $0.623 \pm 0.103$ & $0.729 \pm 0.133$ & $0.625 \pm 0.033$ \\
    \texttt{MRR Cosine Sym}  & $0.012 \pm 0.0$ & $0.021 \pm 0.003$ & $0.075 \pm 0.014$ & $0.211 \pm 0.082$ & $0.034 \pm 0.01$ \\
    \texttt{MRR CDist Sym}      & $0.012 \pm 0.001$ & $0.025 \pm 0.004$ & $0.13 \pm 0.046$ & $0.201 \pm 0.08$ & $0.028 \pm 0.008$ \\
    \texttt{MRR Cosine}      & $0.012 \pm 0.001$ & $0.019 \pm 0.003$ & $0.066 \pm 0.022$ & $0.152 \pm 0.069$ & $0.011 \pm 0.002$ \\
    \texttt{MRR CDist}      & $0.012 \pm 0.001$ & $0.023 \pm 0.005$ & $0.111 \pm 0.052$ & $0.144 \pm 0.073$ & $0.011 \pm 0.003$ \\
    \bottomrule
  \end{tabular}
  }
  \label{tbl:ig}
\end{table*}

\clearpage
\newpage

\section*{NeurIPS Paper Checklist}

\begin{enumerate}

\item {\bf Claims}
    \item[] Question: Do the main claims made in the abstract and introduction accurately reflect the paper's contributions and scope?
    \item[] Answer: \answerYes{} 
    \item[] Justification: Main claims have been listed in the introduction and they have been explained in theoretical aspects (see Sec. \ref{sec:method}) and have been supported by experimental results (Sec. \ref{sec:experiments}).

\item {\bf Limitations}
    \item[] Question: Does the paper discuss the limitations of the work performed by the authors?
    \item[] Answer: \answerYes{}{} %
    \item[] Justification: The limitations in our introduced approach has been discussed in Sec. \ref{sec:conclusion_and_discussion}, including the possible limitations with scalibility and computational efficiency. 

\item {\bf Theory assumptions and proofs}
    \item[] Question: For each theoretical result, does the paper provide the full set of assumptions and a complete (and correct) proof?
    \item[] Answer: \answerYes{}{} %
    \item[] Justification: Theoretical background and notation are clearly explained in detail in Sec. \ref{sec:notation_and_background}. 

    \item {\bf Experimental result reproducibility}
    \item[] Question: Does the paper fully disclose all the information needed to reproduce the main experimental results of the paper to the extent that it affects the main claims and/or conclusions of the paper (regardless of whether the code and data are provided or not)?
    \item[] Answer: \answerYes{} %
    \item[] Justification: We provide the pseudocode for our approach in Sec. \ref{sec:method}, as well as additional implementation details (such as architectural details and hyperparameters) in Appendix~\ref{sec:additional_details}. All the models and the datasets used in the experiments have been cited. Experimental settings have also been explained in Sec. \ref{sec:experiments}, \ref{sec:experiments_on_vision_foundation_models}, including details such as number of anchors used in each experiment. 

\item {\bf Open access to data and code}
    \item[] Question: Does the paper provide open access to the data and code, with sufficient instructions to faithfully reproduce the main experimental results, as described in supplemental material?
    \item[] Answer: \answerYes{} 
   \item[] Justification: All the datasets used in the experiments have open access and they been cited accordingly in the paper. Codebase is not included in the main paper (it is available in the supplementary material), and the architectural details of the implemented models, such as classification head and decoders, are explained in Appendix~\ref{app:arch}. Other pretrained models are explained clearly (including the hyperparamer choices) and cited, partly in the main paper, and mostly in Appendix~\ref{sec:additional_details}. 

\item {\bf Experimental setting/details}
    \item[] Question: Does the paper specify all the training and test details (e.g., data splits, hyperparameters, how they were chosen, type of optimizer, etc.) necessary to understand the results?
    \item[] Answer: \answerYes{} %
    \item[] Justification: Some experimental results are provided in the main paper, while more are provided in Appendix~\ref{sec:additional_details}.

\item {\bf Experiment statistical significance}
    \item[] Question: Does the paper report error bars suitably and correctly defined or other appropriate information about the statistical significance of the experiments?
    \item[] Answer: \answerYes{} %
    \item[] Justification: Our result tables report both the means and the standard deviations. Further details on the statistical properties of the performances of the methods are provided in Appendix~\ref{app:additional_results}, where we provide the full results and results under alternative aggregation.

\item {\bf Experiments compute resources}
    \item[] Question: For each experiment, does the paper provide sufficient information on the computer resources (type of compute workers, memory, time of execution) needed to reproduce the experiments?
    \item[] Answer: \answerYes{} %
    \item[] Justification: Details on the compute resources are provided in Appendix~\ref{app:compute_resources}.
    
\item {\bf Code of ethics}
    \item[] Question: Does the research conducted in the paper conform, in every respect, with the NeurIPS Code of Ethics \url{https://neurips.cc/public/EthicsGuidelines}?
    \item[] Answer: \answerYes{} %
    \item[] Justification: We closely follow the NeurIPS Code of Ethics.

\item {\bf Broader impacts}
    \item[] Question: Does the paper discuss both potential positive societal impacts and negative societal impacts of the work performed?
    \item[] Answer: \answerYes{} %
    \item[] Justification: Our work studies geometric concepts in distinct neural network models, thus has the potential to advance our understandings of deep learning in general. We remark that our work thus also suffers from the potential negative societal impacts of deep learning, as the trained models might be used for dubious purposes.
    
\item {\bf Safeguards}
    \item[] Question: Does the paper describe safeguards that have been put in place for responsible release of data or models that have a high risk for misuse (e.g., pretrained language models, image generators, or scraped datasets)?
    \item[] Answer: \answerNA{} %
    \item[] Justification: We do not release such data or models. We largely focus on using open source data and open source models.

\item {\bf Licenses for existing assets}
    \item[] Question: Are the creators or original owners of assets (e.g., code, data, models), used in the paper, properly credited and are the license and terms of use explicitly mentioned and properly respected?
    \item[] Answer: \answerYes{} %
    \item[] Justification: We provide some information in the main paper, and more in the Appendix~\ref{app:arch}.

\item {\bf New assets}
    \item[] Question: Are new assets introduced in the paper well documented and is the documentation provided alongside the assets?
    \item[] Answer: \answerYes{} %
    \item[] Justification: Concerning code, our code is provided in the Supplement and will be openly available upon acceptance. Concerning data, we focus on open source datasets.

\item {\bf Crowdsourcing and research with human subjects}
    \item[] Question: For crowdsourcing experiments and research with human subjects, does the paper include the full text of instructions given to participants and screenshots, if applicable, as well as details about compensation (if any)? 
    \item[] Answer: \answerNA{}{} %
    \item[] Justification: The paper does not involve research with human subjects, therefore it is not relevant. 

\item {\bf Institutional review board (IRB) approvals or equivalent for research with human subjects}
    \item[] Question: Does the paper describe potential risks incurred by study participants, whether such risks were disclosed to the subjects, and whether Institutional Review Board (IRB) approvals (or an equivalent approval/review based on the requirements of your country or institution) were obtained?
    \item[] Answer: \answerNA{}{} %
    \item[] Justification: The paper does not involve research with human subjects, therefore the question is not relevant. 

\item {\bf Declaration of LLM usage}
    \item[] Question: Does the paper describe the usage of LLMs if it is an important, original, or non-standard component of the core methods in this research? Note that if the LLM is used only for writing, editing, or formatting purposes and does not impact the core methodology, scientific rigorousness, or originality of the research, declaration is not required.
    \item[] Answer: \answerNA{} %
    \item[] Justification: The core method of the paper does not involve LLMs.

\end{enumerate}

\end{document}